%% file: main.tex
\pdfoutput=1

\documentclass[11pt]{article}

\usepackage[preprint]{acl}
\usepackage{times}
\usepackage{latexsym}
 \usepackage{tablefootnote}
\usepackage{wrapfig}
\usepackage{amssymb, amsmath}
\usepackage{pifont}
\usepackage{times}
\usepackage{latexsym}
\usepackage[most]{tcolorbox}
\usepackage{inconsolata}

\usepackage{booktabs}
\usepackage{multirow}
\usepackage{newtxmath}
\usepackage[inkscapelatex=false]{svg}
\usepackage[T1]{fontenc}

\usepackage[utf8]{inputenc}

\usepackage{microtype}

\usepackage{inconsolata}

\usepackage{graphicx}
\newcommand{\bspace}{\hspace{1em}}
\title{Math Neurosurgery: Isolating Language Models' Math Reasoning Abilities Using Only Forward Passes}
\author{
    Bryan R. Christ\bspace
    Zack Gottesman\bspace 
    Jonathan Kropko\bspace
    Thomas Hartvigsen\\
    University of Virginia\\
    {\small \texttt{brc4cb@virginia.edu,qdw5jf@virginia.edu,jk8sd@virginia.edu,hartvigsen@virginia.edu}}
}

\begin{document}
\maketitle
\begin{abstract} \input{sections/abstract}
\end{abstract}

\section{Introduction} 
\input{sections/intro}
\vspace{-\topsep}
\section{Related Work} \label{related_work}
\vspace{-\topsep}
\input{sections/related_work}

\section{Methods} \label{methods}
\input{sections/methods}

\section{Experiments} \label{experiments}
\input{sections/experiments}

\section{Conclusion} \label{conclusion}
\input{sections/conclusion}

\section*{Limitations} \label{limitations}
\input{sections/limitations}

\section*{Ethics Statement} \label{ethics}
\input{sections/ethics}

\bibliography{custom}

\appendix
\section{GSM8K Pruning Results} 
\label{sec:pruning_figures}
\input{sections/pruning_figures}

\section{GSM8K Scaling Results} 
\label{sec:scaling_figures}
\input{sections/scaling_figures}

\section{One Sample GSM8K Pruning Results} 
\label{sec:one_sample_pruning_figures}
\input{sections/one_sample_pruning_figures}

\section{One Sample GSM8K Scaling Results} 
\label{sec:one_sample_scaling_figures}
\input{sections/one_sample_scaling_figures}

\section{Impact of MathNeuro on Unseen Downstream Tasks} 
\label{sec:unseen_tasks}
\input{sections/unseen_tasks}

\section{MathNeuro Using the MATH Dataset} 
\label{sec:math_results}
\input{sections/math_results}

\section{Sample Outputs} \label{sec:sample_outputs}
\input{sections/sample_outputs}

\section{Scaling Factor Grid Search} 
\label{sec:hyperparameters}
\input{sections/hyperparameter_search}

\section{Number and Location of Math-specific Parameters Using MMLU as $\mathcal{D}_{\text{non-math}}$ } 
\label{sec:mmlu_param_location_number}
\input{sections/mmlu_param_location_number_figures}

\end{document}

%% file: sections/abstract.tex
Math reasoning is an active area of Large Language Model (LLM) research because it is a hallmark of artificial intelligence and has implications in several domains, including math education.
However, few works have explored how math reasoning is encoded within LLM parameters and if it is a skill that can be isolated within models.
Doing so could allow targeted intervention to improve math performance without altering non-math behavior and foster understanding of how models encode math reasoning. We introduce Math Neurosurgery (\textit{MathNeuro}), a computationally efficient method we use to isolate math-specific parameters in LLMs using only forward passes. MathNeuro builds on existing work by using weights and activations to calculate parameter importance, but isolates math-specific parameters by filtering out those important for general language tasks. Through pruning parameters MathNeuro identifies, we delete a LLM's math reasoning ability without significantly impacting its general language ability. Scaling the identified parameters by a small constant improves a pretrained or instruction-tuned LLM's performance by 4-17\% on GSM8K and 5-35\% on MATH while leaving non-math behavior unaltered. MathNeuro is also data efficient: most of its effectiveness holds when identifying math-specific parameters using a single sample. MathNeuro highlights the potential for future work to intervene on math-specific parameters.\footnote{Our experimental code can be found at \href{https://github.com/bryanchrist/MathNeuro}{https://github.com/bryanchrist/MathNeuro}}

%% file: sections/intro.tex
Math reasoning, or solving math problems with logic, is an active area of LLM research because it represents artificial intelligence (e.g., \citealt{ahn_large_2024, li_gsm-plus_2024}) and has implications in many domains, including math education (e.g, \citealt{christ_mathwell_2024, wang_tutor_2024}) and automated theorem proving (e.g., \citealt{song_towards_2024, xin_deepseek-prover_2024}). Yet, few works have explored how LLMs encode math reasoning abilities in their parametric knowledge. Identifying math-specific parameters could be beneficial for many reasons, including a) targeting the right parameters to intervene on to improve a model's math reasoning ability as others have done in other domains (e.g., \citealt{tang_language-specific_2024, suau_whispering_2024}), b) doing so without altering behavior on other tasks like these works have done in their domains, and c) fostering knowledge of how LLMs encode math reasoning. While some works explore how different math concepts or terms are stored or processed in model layers or neurons (e.g., \citealt{hanna_how_2023,rai_investigation_2024, stolfo_mechanistic_2023}), none have developed a method for isolating parameters for math reasoning. 

Outside of math reasoning, several works have explored how to identify neurons or parameters associated with particular knowledge or skills in LLMs \cite{chang_localization_2024, dai_knowledge_2022,panigrahi_task-specific_2023, tang_language-specific_2024, wang_finding_2022}. While some methods are computationally expensive because they use gradient information, which may not be feasible for large models (e.g., \citealt{panigrahi_task-specific_2023}), others are easier to compute because they rely on information obtained through forward passes, particularly as captured by activations (e.g., \citealt{tang_language-specific_2024}). However, it is unknown if these domain-specific methods for single skill identification can effectively isolate a broad concept like math reasoning, which may be entangled with many other abilities within a LLM (e.g., reading comprehension, general knowledge).


We conduct the first study of parameter importance in LLMs for math reasoning. We apply two state-of-the-art (SOTA) gradient-free parameter importance methods to math reasoning. We find one of these methods, LAPE \cite{tang_language-specific_2024}, performs inconsistently at identifying math-specific neurons across models, while the other, Wanda \cite{sun_simple_2023}, identifies parameters important for math, but is unable to isolate math-specific parameters because the parameters it identifies overlap significantly with those important for other tasks. To address these limitations of existing methods, we develop a new method called Math Neurosurgery (MathNeuro) we use to isolate math-specific parameters. Building on Wanda, MathNeuro uses weights and activations to calculate parameter importance and achieve a context-aware representation of importance. MathNeuro extends beyond Wanda by adding task-specific parameter identification. Specifically, to isolate parameters important for math and not other abilities, MathNeuro filters out identified parameters that are found to be important for other general language understanding tasks.

We provide evidence that MathNeuro effectively isolates math-specific parameters by evaluating it with five LLMs from 1-8B parameters. Pruning parameters identified by MathNeuro effectively deletes a model's math reasoning ability. Despite destroying math reasoning, pruning these parameters results in a performance drop on other, non-math tasks similar to the impact of random parameter pruning. We also find that scaling up MathNeuro-identified parameters by a small universal factor can boost both instruction-tuned and pre-trained LLMs' GSM8K \cite{cobbe_training_2021} or MATH \cite{hendrycks_measuring_2021-1} performance by 4-17\% or 5-35\% across models, respectively. We further show that our method is data efficient: MathNeuro is almost as effective using only a \textit{single} sample to calculate parameter importance. In addition, we show MathNeuro consistently identifies a similar subset of parameters as math-specific across different sets of samples and that these parameters generalize across math reasoning tasks. We find math-specific parameters are located roughly evenly throughout a model's decoder blocks, suggesting math reasoning is likely encoded throughout a model's parameters rather than being concentrated in a specific layer or layers. 

Our key contributions are as follows: 
\vspace{-\topsep}
\begin{itemize}
  \setlength{\parskip}{0pt}
  \setlength{\itemsep}{0pt plus 1pt}
    \item We design MathNeuro, a simple yet effective way to isolate LLM math reasoning by filtering out parameters important for other tasks.
    \item We demonstrate the effectiveness of this method by showing that deleting parameters identified by MathNeuro destroys a model's math performance and scaling them by a universal factor can increase it by 4-35\%.
    \item We verify MathNeuro isolates math-specific parameters by showing pruning or scaling them does not significantly impact non-math performance more than random perturbation.
\end{itemize}
\vspace{-\topsep}

%% file: sections/related_work.tex
\vspace{-\topsep}
\paragraph{Skill and Knowledge Localization in LLMs}
Several works have explored skill and knowledge localization in language models, although none focus on math specifically \cite{bau_identifying_2018,chang_localization_2024, dalvi_what_2018, dai_knowledge_2022, dalvi_analyzing_2020, gurnee_finding_2023,kojima_multilingual_2024, leng_towards_2024, panigrahi_task-specific_2023, radford_learning_2017, suau_whispering_2024, sun_simple_2023, tang_language-specific_2024, wang_finding_2022, xin_what_2019, zhao_how_2024}. Many methods use gradient information to calculate parameter importance, which is computationally infeasible for large models \cite{dai_knowledge_2022, leng_towards_2024, panigrahi_task-specific_2023, wang_finding_2022}. However, others are more lightweight and calculate parameter importance using only forward passes, predominately through using information obtained through activation values \cite{kojima_multilingual_2024, suau_whispering_2024, sun_simple_2023, tang_language-specific_2024, zhao_how_2024}. While these methods may find parameters important for the domains they study, it is unclear if they could identify parameters important for math reasoning, which could be distributed throughout a model or interwoven with other important natural language abilities given the task's complexity. To identify important parameters, MathNeuro builds upon Wanda \cite{sun_simple_2023}, a SOTA LLM pruning method that prunes parameters \textit{unimportant} for a model's output as measured by the smallest absolute value of weights times activations. MathNeuro inverts Wanda by identifying the \textit{most important} parameters for a task and isolates math-specific parameters by filtering out parameters important for non-math, general language tasks through introducing task-specific data into the importance calculation.
\paragraph{Math Skill Localization in LLMs}
Some studies have explored how math knowledge is encoded within LLMs \cite{hanna_how_2023, nikankin_arithmetic_2024, rai_investigation_2024, stolfo_mechanistic_2023, zhang_interpreting_2024, zhu_language_2025}. These works focus on how and where particular math concepts and key phrases such as addition and subtraction are processed by LLMs. While these findings are insightful, they do not identify parameters critical for a model's overall math performance but rather ones relating to processing different math concepts.

%% file: sections/methods.tex
\begin{figure*}[t]
    \centering
    \includegraphics[width=1\linewidth]{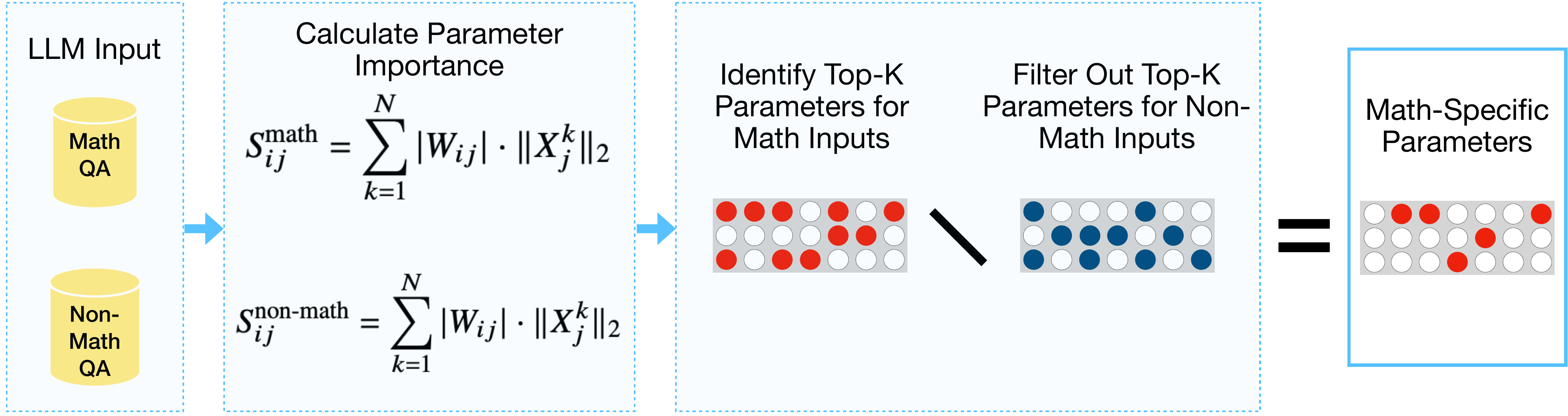}
    \caption{Overview of MathNeuro. First, we sum weights times activations over N samples for math and non-math inputs, finding the top-K parameters for each input type. Next, we find math-specific parameters by filtering out parameters important for non-math inputs.}
    \label{fig:MathNeuro}
\end{figure*}
We propose MathNeuro, a parameter identification method that calculates importance using only forward passes. First, we separately identify LLM parameters important for a math task and a non-math, general language task using samples for each task. Next, MathNeuro isolates math-specific parameters by taking the subset that are important for the math task but not for the non-math task. While MathNeuro may work for other, non-math tasks, we study math specifically. We describe the problem setup and our method in more detail below.

\subsection{Preliminaries}
Identifying parameters important for math reasoning in LLMs is beneficial because it is a critical AI capability, is understudied in interpretability work, has implications in several domains, and thus is an interesting test domain. However, this is nontrivial given that math reasoning not only involves direct computation, but also natural language reasoning. Thus, it may be difficult to distinguish parameters important only for math reasoning from those important for general language. Indeed, other work has found significant overlap between parameters important for different tasks \cite{tang_language-specific_2024}.

\subsection{Identifying Top Parameters}
MathNeuro adapts Wanda \cite{sun_simple_2023}, a SOTA LLM pruning method that prunes, or removes, parameters unimportant for a model's output. Wanda identifies parameters to prune using the absolute value of weights times activations for an input, providing a context-aware representation of parameter importance. Wanda produces a score $S_{ij}$ for weight ${j}$ in neuron ${i}$ within a weight matrix:
\begin{equation*}
    S_{ij} = |W_{ij}| \cdot \|X_j\|_2
\end{equation*}
where $W_{ij}$ represents the weight, $|\cdot|$ is the absolute value operator, and $\| X_j \|_2$ is the $\ell_2$ norm of the \(j\)-th feature aggregated across input tokens to normalize the input $X$, or activation values. Wanda then prunes the parameters with the smallest scores. Wanda considers both weights and activations as elements of parameter importance because small but highly activated weights can be highly influential, while large but lightly activated weights may be less influential. MathNeuro inverts Wanda by identifying the parameters with the largest weights times activations as being most important to a given task. 

\subsection{Isolating Math-specific Parameters}
While naively identifying the parameters with the highest absolute value of weights times activations may find parameters important for a given task, it may not isolate the parameters important for that task \textit{only}, as discussed above. Thus, we calculate parameter importance for other unrelated tasks and use the disjoint set between these sets of parameters as the ones that are math-specific, which is the critical innovation of MathNeuro. To do this, we separately sum\footnote{This summation is akin to gradient-based identification methods summing gradients over inputs (e.g., \citealt{das_unified_2023}).} the absolute value of weights times activations for each parameter in attention and MLP layers for each data point $k$ across $N$ samples from a math dataset and an unrelated natural language task dataset. We focus on attention and MLP layers because recent work has found that knowledge and skills are often distributed in these two model components \cite{wei_does_2024, yin_lofit_2024}. Using this formulation, we compute scores for each parameter over math and non-math inputs: 
\begin{equation*}
    S_{ij}^{\text{math}} = \sum_{k=1}^{N} |W_{ij}| \cdot \|X_{j}^{k}\|_2\ \text{for}\ X \in \mathcal{D}_{\text{math}}
\end{equation*}
\begin{equation*}
    S_{ij}^{\text{non-math}} = \sum_{k=1}^{N} |W_{ij}| \cdot \|X_{j}^{k}\|_2\ \text{for}\ X \in \mathcal{D}_{\text{non-math}}
\end{equation*}
Then, we separately identify the top K\% of parameters with the highest score for each task in each layer. Lastly, we take the subset of parameters most important for the math task that are not in the set of parameters most important for the unrelated task, or $T_{math} = \text{TopK}_{math} \setminus \mathrm{TopK}_{non-math}$. 


%% file: sections/experiments.tex
We next validate if MathNeuro successfully identifies math-specific parameters. We compare against SOTA alternatives and a simple baseline in two settings: 1) pruning parameters identified as important for math and 2) scaling these parameters. Pruning or scaling task-specific parameters is equivalent to the approach recent work has taken to deactivate or more highly activate neurons identified as language or knowledge specific \cite{kojima_multilingual_2024, suau_whispering_2024, tang_language-specific_2024, zhao_how_2024}, respectively, but intervenes on the weight rather than activation level. We show the impact of each intervention on both math and non-math performance across five LLMs ranging from 1-8B parameters. We perform parameter identification experiments using 500 samples and a single sample.

\subsection{Experimental Setup} 
\paragraph{Models} We evaluate five LLMs of varying sizes: Phi 1.5 (1B) \cite{li_textbooks_2023}, Llama 3.2 1B Instruction Tuned (IT) \cite{metaai_llama_nodate}, Gemma 2 2B IT \cite{gemma_team_gemma_2024}, Llama 3.2 3B IT \cite{metaai_llama_nodate}, and Llama 3.1 8B IT \cite{metaai_introducing_nodate}. We display results for Llama 3.2 1B IT below and report results for the other models in Appendices \ref{sec:pruning_figures}, \ref{sec:scaling_figures}, \ref{sec:one_sample_pruning_figures}, and \ref{sec:one_sample_scaling_figures}, which follow similar trends to those discussed below. We focus on instruction tuned models to evaluate if MathNeuro can successfully identify math-specific parameters in models that a) perform well at math given their size and b) are trained for a range of tasks, which means it may be more difficult to identify math-specific parameters. Phi 1.5 serves as a baseline for if MathNeuro works for a pretrained, non-IT model. 
\paragraph{Datasets} For identifying math-specific parameters, we use the popular and high-quality GSM8K \cite{cobbe_training_2021} and MATH \cite{hendrycks_measuring_2021-1} datasets. We calculate parameter importance using the GSM8K or MATH training split and evaluate the impact of each method on the GSM8K or MATH test split. We report GSM8K results below and MATH results in Appendix \ref{sec:math_results} given the GSM8K results replicate for the MATH dataset. Following prior work \cite{agarwal_many-shot_2024, brown_large_2024, lee_llm2llm_2024, li_common_2024}, we subset the GSM8K test split to the same 200 random samples for every model for experimental efficiency. For identifying parameters important for non-math tasks and measuring performance drops after eliminating math-specific parameters, we follow recent work that assesses catastrophic forgetting in LLMs \cite{luo_empirical_2024} by using RACE \cite{lai_race_2017} for measuring reading comprehension and MMLU \cite{hendrycks_measuring_2021} for measuring general knowledge. These datasets are general language understanding tasks that are different from math reasoning. While MMLU contains some math-related questions, it assesses a variety of knowledge that, in aggregate, is mostly not math-specific. We conduct all evaluations using the Eleuther AI LM Evaluation Harness \cite{gao_framework_2024} and use an 8-shot chain-of-thought (CoT) prompting format for GSM8K, as is standard.

\paragraph{Baselines}
We compare MathNeuro to three identification methods computed using forward passes:

(a) \textit{Wanda} \cite{sun_simple_2023}: We calculate parameter importance for math inputs and choose the top K\% of parameters without filtering out those important for other unrelated tasks. 

(b) \textit{Language Activation Probability Entropy (LAPE)} \cite{tang_language-specific_2024}: LAPE finds language-specific neurons by thresholding activation probabilities
as calculated by samples for each language under consideration. We use GSM8K, MMLU, and RACE for calculating task-specific neurons using this method. Using LAPE allows us to determine if existing activation-only parameter identification methods can isolate math-specific parameters. 

(c) \textit{Random Parameter Identification}: As a sanity check, we randomly select the same number of parameters as those identified by MathNeuro when using MMLU or RACE as $\mathcal{D}_{\text{non-math}}$.

\begin{figure*}[t]
    \centering
    \setlength{\abovecaptionskip}{0pt}
    \includegraphics[width=1\linewidth]{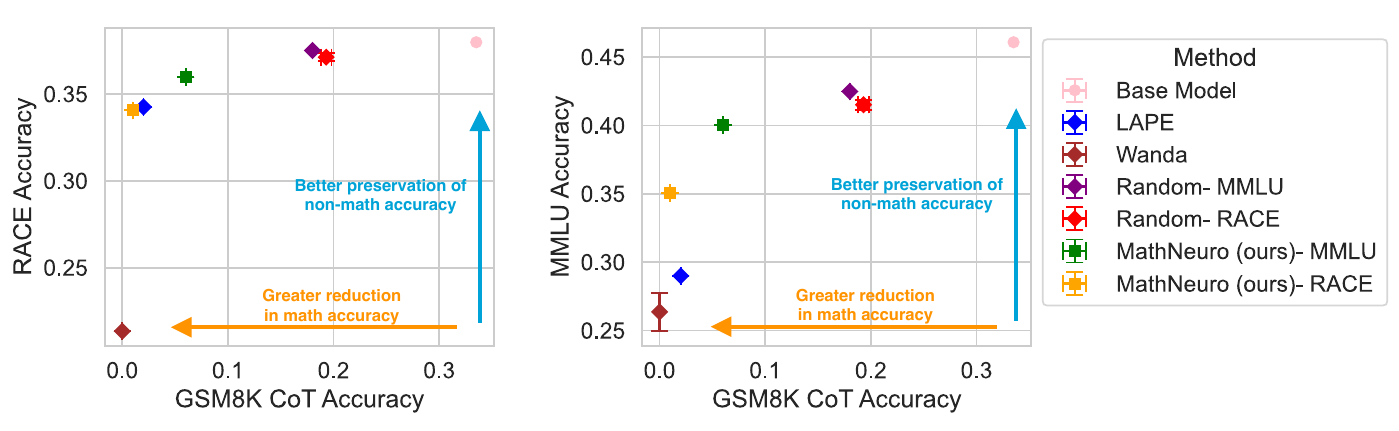}
    \caption{Effect of \textbf{\textit{pruning}} identified parameters on math and non-math performance for Llama 3.2 1B IT with $\mathrm{TopK}_{math} = \mathrm{TopK}_{non-math} = 15\%$. Ideal methods fall in the top left of the plot. MMLU and RACE denote the dataset used as $\mathcal{D}_{\text{non-math}}$. Horizontal and vertical lines represent 95\% confidence intervals for each point.}
    \label{fig:MathNeuro_llama1b_prune}
\end{figure*}
\subsection{Pruning Top Math Parameters} \label{deleting_params}
To test if the four parameter identification methods (MathNeuro and three baselines) identify parameters important for math reasoning, we identify important parameters using each method for each model and prune them (set them to 0). We then compare each model's GSM8K, RACE, and MMLU accuracy to their own unedited performance. We do this five times for each model with different random subsets of 500 samples from each dataset to identify the average performance of each method. We identify the top .01, .1, .5, 1, 2.5, 5, 10 and 15\% of parameters for each comparison and report the parameter proportion with the best performance. Appendix \ref{sec:pruning_figures} explores how parameter proportion impacts GSM8K performance; notably, this hyperparameter does not impact performance for Wanda and LAPE exhibits a performance curve similar to MathNeuro.

Figures \ref{fig:MathNeuro_llama1b_prune}, \ref{fig:MathNeuro_phi15_prune}, \ref{fig:MathNeuro_gemma2b_prune}, \ref{fig:MathNeuro_llama3b_prune} and \ref{fig:MathNeuro_llama8b_prune} show results from this experiment. An ideal method would fall in the top left of these plots, meaning math performance (GSM8K) is deleted while non-math performance (RACE and MMLU) is maintained. As seen in the figures, MathNeuro and Wanda eliminate math performance across models, while LAPE performs inconsistently across models, often displaying wide confidence intervals, suggesting its performance is sensitive to which data points are used. However, while Wanda also destroys each model's ability to perform non-math tasks, MathNeuro effectively isolates math-specific parameters across models, as shown in non-math performance decreases that are similar to the effect of random pruning. 

\begin{figure*}[t]
    \centering
    \setlength{\abovecaptionskip}{0pt}
    \includegraphics[width=1\linewidth]{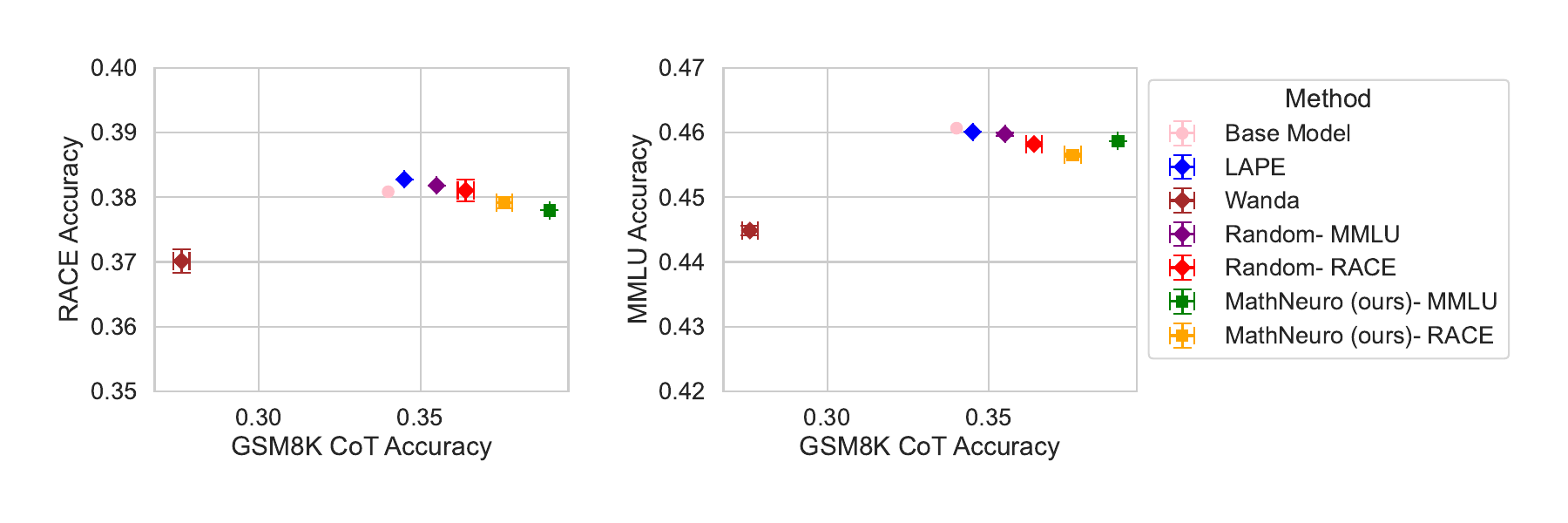}
    \caption{Effect of \textbf{\textit{scaling}} identified parameters by 1.1 on math and non-math performance for Llama 3.2 1B IT with $\mathrm{TopK}_{math} = \mathrm{TopK}_{non-math} = 5\%$. Ideal methods fall in the top right of the plot. MMLU and RACE denote the dataset used as $\mathcal{D}_{\text{non-math}}$. Horizontal and vertical lines represent 95\% confidence intervals for each point.}
    \label{fig:MathNeuro_llama1b_scale}
\end{figure*}
\subsection{Scaling Top Math Parameters} 
\vspace{-\topsep}
\label{scaling_params}
We next evaluate performance when more highly activating math-specific parameters by scaling the weights by a universal factor.  For smaller models, we find the scalar 1.1 works best, while for larger models (Llama 3.1 8B IT), a smaller factor (1.01) works better. While we leave a rigorous study of this hyperparameter to future work due to its computational expense, see Appendix \ref{sec:hyperparameters} for our experimentation with scale factors. As in Section \ref{deleting_params}, we scale the parameters each method identifies based on 500 random samples from each relevant dataset and repeat the process five times, reporting the parameter proportion that performs best.
\newline \indent Figures \ref{fig:MathNeuro_llama1b_scale}, \ref{fig:MathNeuro_phi15_scale}, \ref{fig:MathNeuro_gemma2b_scale}, \ref{fig:MathNeuro_llama3b_scale}, and \ref{fig:MathNeuro_llama8b_scale} display results from this experiment. An ideal method would fall in the top right of these plots, meaning GSM8K accuracy increases while non-math performance is maintained. As shown in these figures, scaling parameters identified by MathNeuro results in a GSM8K performance increase of 4-17\% across models, while scaling Wanda-identified parameters tends to either harm or slightly improve performance. As with pruning, LAPE performs inconsistently in the scaling setting across models. Scaling random parameters can help for some models, although the effect is not consistent across models. Each parameter identification method does not harm performance on RACE or MMLU, suggesting scaling's impact tends to be localized to math performance. 
\subsection{MathNeuro with a Single Sample} \label{single_sample}
\begin{figure*}[t]
    \centering
    \setlength{\abovecaptionskip}{0pt}
    \includegraphics[width=1\linewidth]{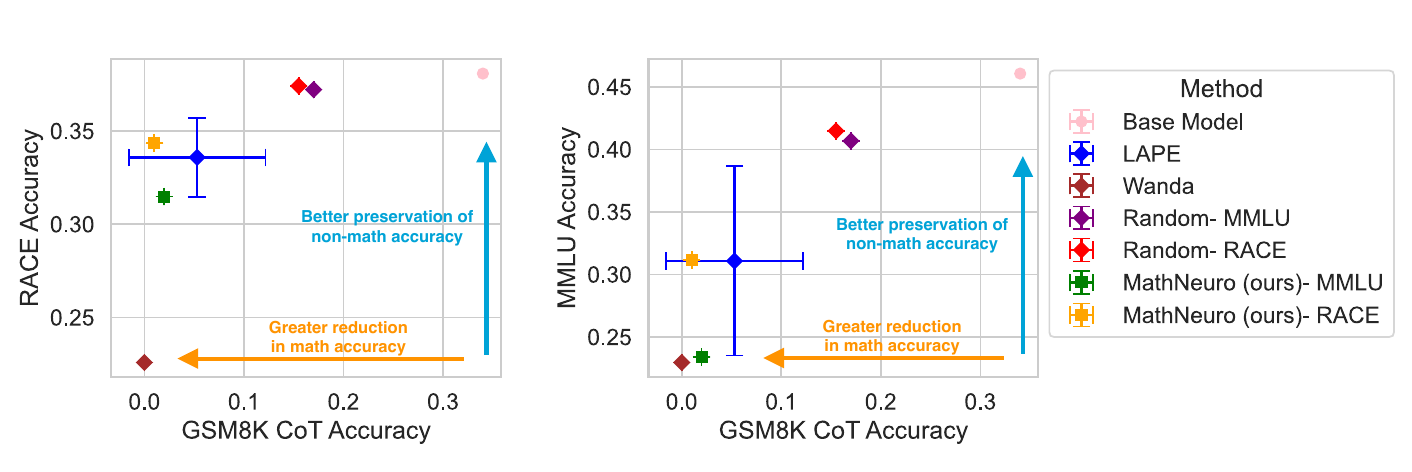}
    \caption{Effect of \textbf{\textit{pruning}} identified parameters on math and non-math performance for Llama 3.2 1B IT with $\mathrm{TopK}_{math} = \mathrm{TopK}_{non-math} = 10\%$ \textbf{\textit{based on one sample}}. Ideal methods fall in the top left of the plot. MMLU and RACE denote the dataset used as $\mathcal{D}_{\text{non-math}}$. Horizontal/vertical lines show each point's 95\% confidence intervals.}
    \label{fig:MathNeuro_llama1b_prune_one_sample}
\end{figure*}

\begin{figure*}[t]
    \centering
    \setlength{\abovecaptionskip}{0pt}
    \includegraphics[width=1\linewidth]{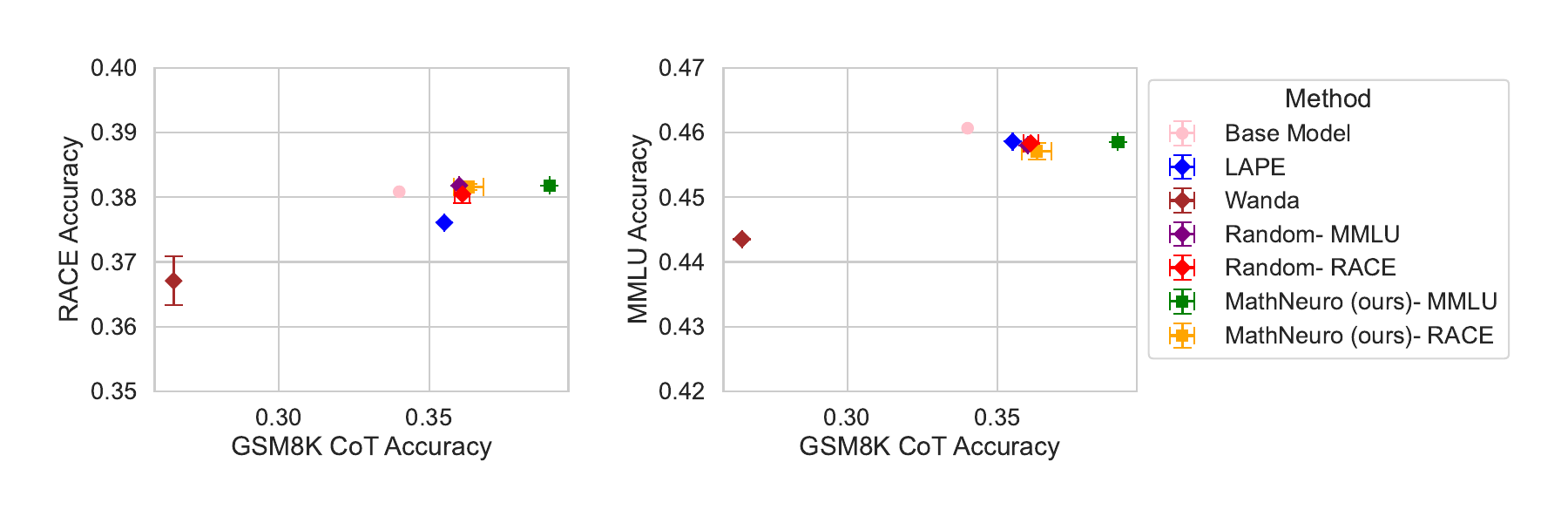}
    \caption{Effect of \textbf{\textit{scaling}} identified parameters by 1.1 on math and non-math performance for Llama 3.2 1B IT with $\mathrm{TopK}_{math} = \mathrm{TopK}_{non-math} = 2.5\%$ \textbf{\textit{based on one sample}}. Ideal methods fall in the top right. MMLU and RACE denote the dataset used as $\mathcal{D}_{\text{non-math}}$. Horizontal/vertical lines show each point's 95\% confidence intervals.}
    \label{fig:MathNeuro_llama1b_scale_one_sample}
\end{figure*}

\vspace{-\topsep}
If a method can identify math-specific parameters using a single sample, then it could inform math interventions for settings where data are limited such as for assessing a specific math operation or topic. To test this with MathNeuro and the baselines, we conduct experiments to identify parameters based on a single math and non-math input. We then prune or scale parameters identified by each method and run each experiment five times using different random samples from each dataset. 
As shown in Figures \ref{fig:MathNeuro_llama1b_prune_one_sample}, \ref{fig:MathNeuro_phi15_prune_one_sample}, \ref{fig:MathNeuro_gemma2b_prune_one_sample}, \ref{fig:MathNeuro_llama3b_prune_one_sample} and \ref{fig:MathNeuro_llama8b_prune_one_sample}, MathNeuro performs best at isolating math-specific parameters when pruning using a single sample, as shown in lower drops in non-math performance relative to Wanda. However, these performance drops are larger than when using more samples, suggesting additional samples help MathNeuro more effectively isolate math-specific parameters. LAPE again performs inconsistently across models in the one sample pruning setting.
\newline \indent As shown in Figures \ref{fig:MathNeuro_llama1b_scale_one_sample}, \ref{fig:MathNeuro_phi15_scale_one_sample}, \ref{fig:MathNeuro_gemma2b_scale_one_sample}, \ref{fig:MathNeuro_llama3b_scale_one_sample}, and \ref{fig:MathNeuro_llama8b_scale_one_sample}, we see similar or smaller, but still meaningful, boosts in GSM8K accuracy when scaling parameters MathNeuro identifies using one math and non-math sample. While random scaling sometimes helps as observed in Section \ref{scaling_params}, the effect is again not consistent across models. In some cases, LAPE and Wanda increase GSM8K accuracy, though the effects are not consistent across models, and they also sometimes decrease performance. For all methods, there is no meaningful performance drop in MMLU or RACE accuracy, suggesting scaling's impact on non-math performance is still minor. 
\subsection{MathNeuro Parameter Consistency, Number, Location and Qualitative Impact}
\begin{figure}[t]
    \centering
    \includegraphics[width=1\linewidth]{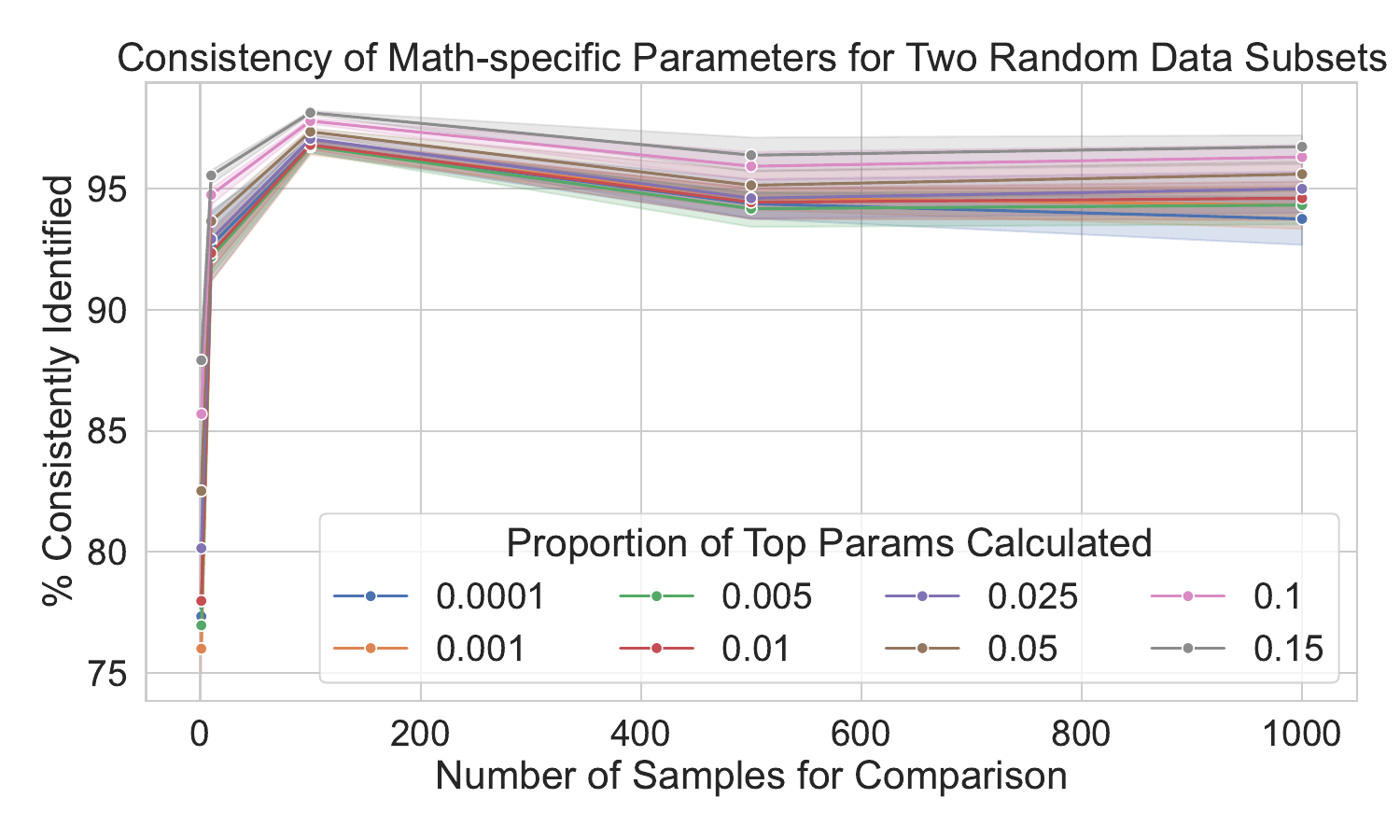}
    \caption{Consistency of math-specific parameters identified by MathNeuro for Llama 3.2 1B IT when identifying using GSM8K compared to RACE.}
    \label{fig:race_consistency}
\end{figure}

Next, we conduct experiments to explore if MathNeuro identifies the same set of parameters as math-specific across different random subsets of math and non-math data and the number and location of these parameters. We report results for these experiments in the sections below using RACE as the non-math dataset and equivalent results for using MMLU as the non-math dataset in Appendix \ref{sec:mmlu_param_location_number}. We also conduct a qualitative evaluation of model outputs after pruning or scaling parameters MathNeuro identifies to explore how math and non-math outputs are affected by the method. All experiments are conducted using Llama 3.2 1B IT. 
\paragraph{Consistency of Math-specific Parameters} \label{consistency}
We next explore if MathNeuro consistently identifies the same parameters as math-specific across different random subsets from a math and non-math dataset. This allows us to identify if math reasoning is in fact reliably concentrated in a subset of model parameters like the experiments above suggest. We first identify math-specific parameters using MathNeuro on two different random subsets from a math and non-math dataset. Next, we calculate the percentage overlap between the parameters identified in both subsets. We do this five times for different sample sizes (1, 10, 100, 500, and 1,000) and for calculating different proportions of top parameters from each dataset. This allows us to construct confidence intervals and see how parameter identification consistency varies when calculating based on different sample sizes and top parameter proportions. As shown in Figure \ref{fig:race_consistency} and Appendix \ref{sec:mmlu_param_location_number}, with 100 or more samples, roughly 95\% or more of the parameters MathNeuro identifies overlap between two random subsets regardless of the proportion of top parameters calculated, which shows that the method is able to consistently identify the most important parameters for math performance and that these parameters are largely invariant with regard to the subset of data used to calculate them. 

\paragraph{Number and Location of Math-specific Parameters} \label{param_location_and_number}
We next examine the proportion of parameters MathNeuro identifies as math-specific. We first identify math-specific parameters using random subsets from each dataset. Next, we calculate the percentage of the top K\% of parameters that are identified as math-specific using those subsets. We repeat this five times for different sample sizes and top K\%s to construct confidence intervals.  

\begin{figure}[t]
    \centering
    \includegraphics[width=1\linewidth]{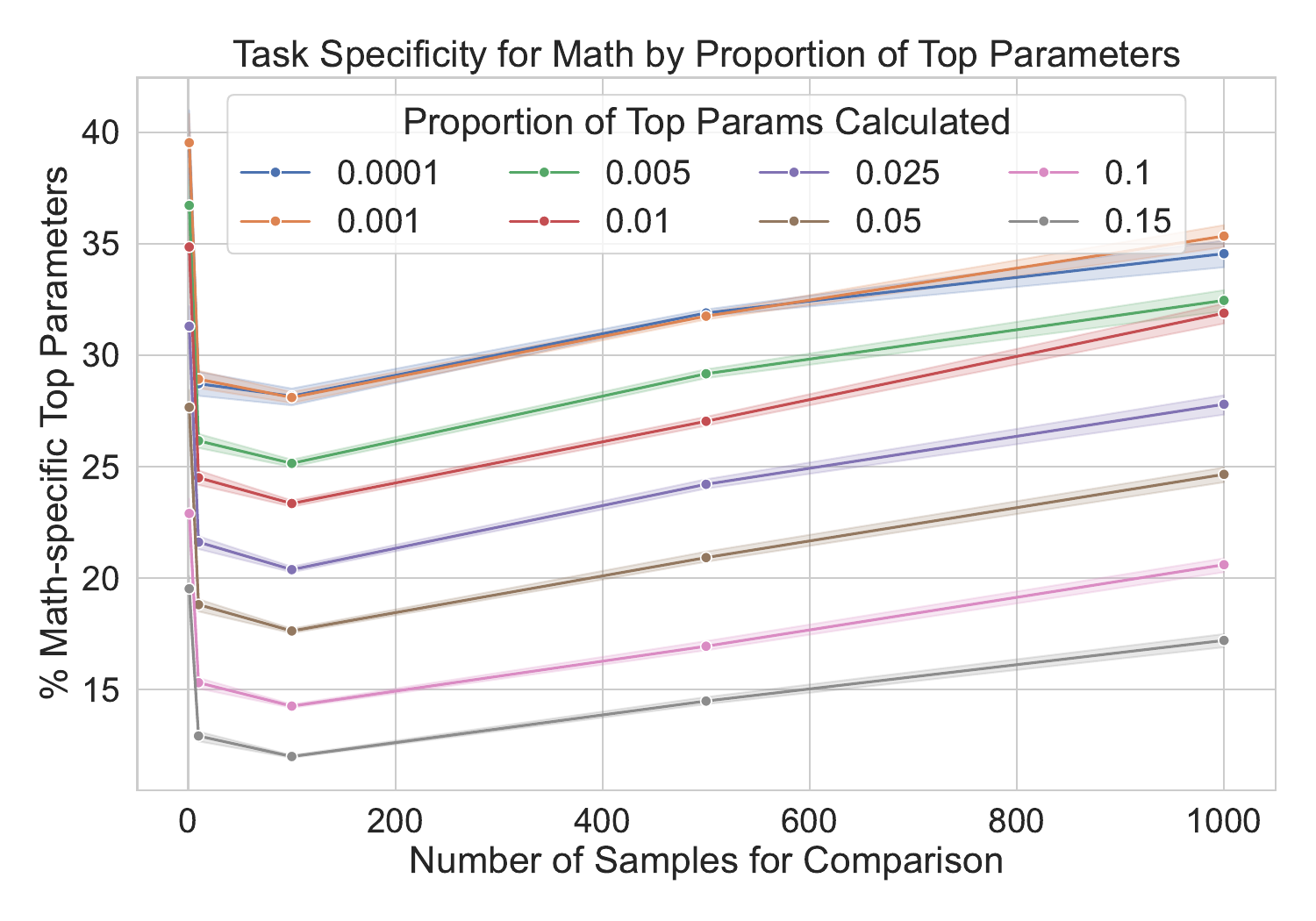}
    \caption{Percentage of top parameters that are math-specific as identified by MathNeuro for Llama 3.2 1B IT when identifying using GSM8K compared to RACE.}
    \label{fig:race_unique}
\end{figure}

As shown in Figure \ref{fig:race_unique} and Appendix \ref{sec:mmlu_param_location_number}, while the most parameters are identified as math-specific when calculating importance with one sample due to randomness, the amount of math-specific parameters identified by MathNeuro generally increases with the number of samples considered for all proportions of top parameters calculated. The relatively high amount of overlap in top parameters between tasks displayed in these figures is likely why MathNeuro performs better than existing parameter identification methods that do not filter out parameters important for other tasks. The percentage of math-specific parameters in the top K\% of parameters declines as the proportion of top parameters calculated increases because as this proportion increases, more of the model's top parameters are considered. These top parameters are likely more task-invariant than those found when considering a smaller percentage of the model's top parameters. 

\begin{figure}[t]
    \centering
    \includegraphics[width=1\linewidth]{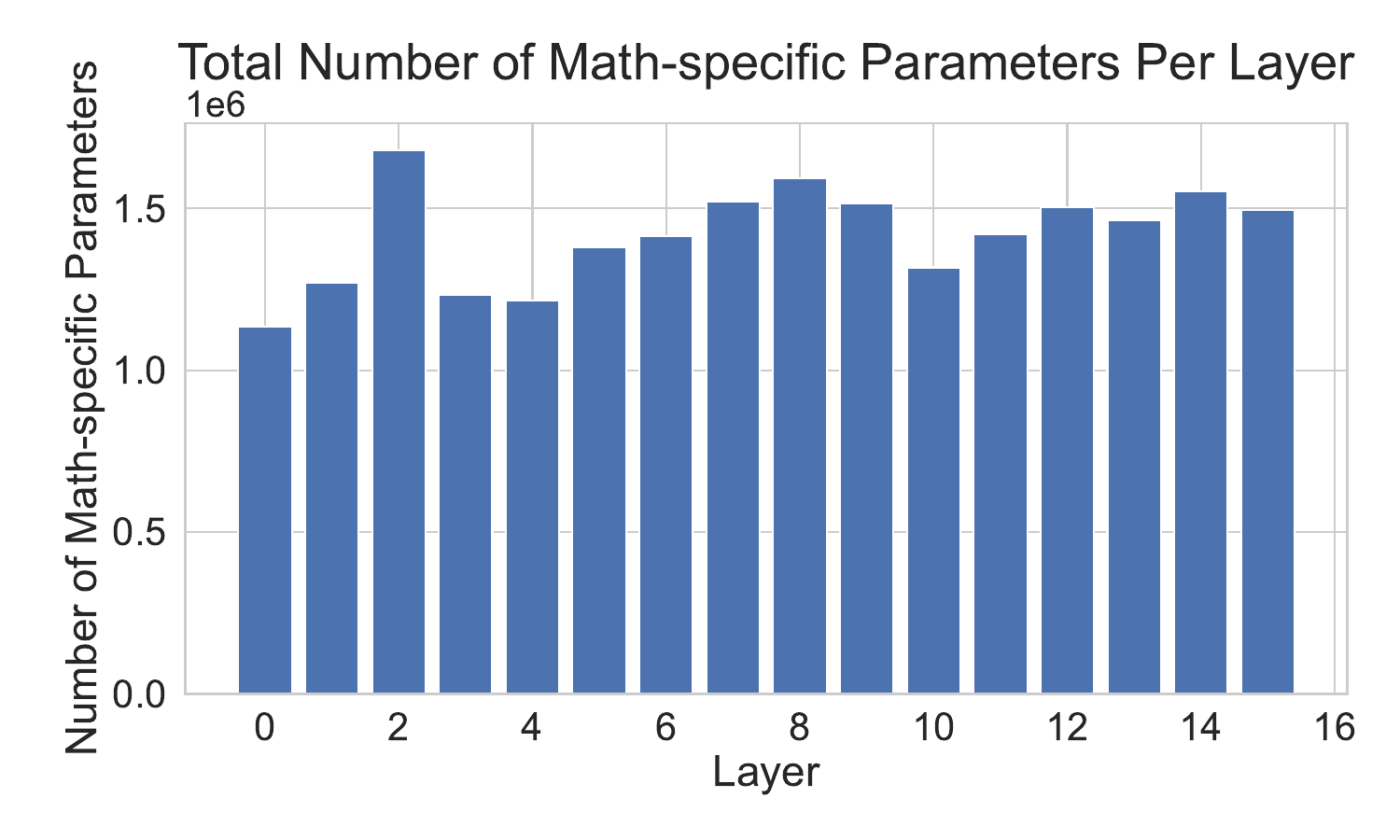}
    \caption{Distribution of math-specific parameters identified by MathNeuro for Llama 3.2 1B IT when identifying using GSM8K compared to RACE.}
    \label{fig:race_distribution}
\end{figure}

\indent To explore where math-specific parameters are located within a model, we sum the number of parameters MathNeuro identifies in each decoder block for Llama 3.2 1B IT. To do this, we calculate the top 15\% of parameters, which is the parameter proportion for which MathNeuro performs best for this model. As shown in Figure \ref{fig:race_distribution} and Appendix \ref{sec:mmlu_param_location_number}, the number of math-specific parameters MathNeuro identifies is relatively consistent across decoder blocks when using either RACE or MMLU for parameter identification. This suggests that models encode math reasoning by distributing the capability throughout their parametric knowledge rather than concentrating it in a few layers. The parameters identified in these figures correspond to just 1.51\% and 1.84\% of the model's overall parameters when calculating parameter importance based on MMLU and RACE, respectively, despite being responsible for nearly all of its math performance.

\paragraph{Qualitative Analysis}
To verify model outputs are still coherent after pruning or scaling, we conduct a qualitative analysis of outputs before and after pruning or scaling parameters MathNeuro identifies. As shown in Tables \ref{tab:gsm8k_outputs},  \ref{tab:race_outputs}, and \ref{tab:mmlu_outputs}, only the output for answering a GSM8K question becomes incoherent after pruning math-specific parameters. Before pruning, the model correctly solves the GSM8K problem; after pruning, it fails to generate an effective CoT. The pruned model effectively responds to RACE questions, although it gets the question wrong before and after pruning. The pruned model generates coherent output to MMLU questions, though it gets the answer right when using RACE as $\mathcal{D}_{\text{non-math}}$ and wrong when using MMLU as $\mathcal{D}_{\text{non-math}}$. These findings confirm our quantitative findings showing the model can still perform non-math tasks after pruning math-specific parameters, although it experiences a performance drop similar to that obtained from random pruning. As shown in Tables \ref{tab:gsm8k_outputs_scaling}, \ref{tab:race_outputs}, and \ref{tab:mmlu_outputs_scaling}, the scaled model's outputs for RACE and MMLU questions remain mostly unmodified, while it correctly solves a GSM8K question after scaling based on MMLU as $\mathcal{D}_{\text{non-math}}$ that it solved incorrectly before scaling. These findings parallel our quantitative findings that math reasoning increases post-scaling while non-math performance remains unchanged. 

\subsection{Impact of MathNeuro on Unseen Tasks} \label{unseen_tasks_experiments}
To explore if math-specific parameters MathNeuro identifies are consistently important across unseen math tasks and unimportant for unseen general language or non-math reasoning tasks, we repeat the pruning experiments reported in Section \ref{deleting_params} using GSM8K as the math dataset and MMLU or RACE as the non-math dataset and evaluate the pruned model on 8 unseen tasks (see Appendix \ref{sec:unseen_tasks} for implementation details). Unseen tasks include 5 that are non-math reasoning or general language tasks (HellaSwag, MuTual, PIQA, WikiText, and WinoGrande) \cite{zellers_hellaswag_2019, cui_mutual_2020, bisk_piqa_2020, merity_pointer_2016, sakaguchi_winogrande_2021} and 3 that are math tasks, both in domain (EGSM; \citealt{christ_mathwell_2024}) and out of domain (MATH, MATHQA) \cite{hendrycks_measuring_2021-1, amini_mathqa_2019}. The figures in Appendix \ref{sec:unseen_tasks} display this experiment's results where the Wanda baseline represents the lowest possible bound on performance for a task given that it deletes all of a model's top parameters. Across models, pruning parameters identified by MathNeuro effectively deletes math performance on both in domain and out of domain tasks while mostly maintaining performance on general language and non-math reasoning tasks. These findings expand those discussed in Section \ref{deleting_params} by showing that math-specific parameters MathNeuro identifies are universally important across math reasoning tasks and unimportant across non-math tasks regardless of the math dataset used for identification. This generalizability across math tasks allays concerns that parameters identified in one dataset are really targeted towards characteristics of that dataset and, instead, shows MathNeuro isolates parameters important for math reasoning broadly. 

%% file: sections/conclusion.tex
Although math reasoning is an active area of LLM research, few works have explored how it is encoded within LLM parameters and if it is a skill that can be isolated within a model. We introduce MathNeuro, a forward-only identification method we use to isolate math-specific parameters in LLMs. Through comprehensive experiments, we demonstrate MathNeuro's effectiveness by showing pruning or scaling the parameters it identifies can delete or reinforce a LLM's math reasoning ability, respectively, despite its simplicity and ease of calculation. Future work should build on this method by applying it to other domains and developing interventions for math-specific parameters that improve a model's performance on mathematical reasoning without catastrophic forgetting.


%% file: sections/limitations.tex
While we comprehensively evaluate MathNeuro using several math and non-math datasets used in other works and focus our evaluations on math reasoning specifically, there are many other natural language and mathematical reasoning tasks models could be evaluated on. Future work should consider extending MathNeuro to these additional tasks and explore if MathNeuro can isolate parameters important for non-math tasks. While we used five recent models for our experiments, future work should also include additional models, especially those of larger sizes (>8B). Additionally, due to computational expense, we were unable to conduct a full hyperparameter sweep for an optimal universal scaling factor for parameters identified by MathNeuro, though the rough grid search we report in Appendix \ref{sec:hyperparameters} highlights that larger scale factors tend to work better for smaller models and smaller scale factors tend to work better for larger models.

%% file: sections/ethics.tex
All data used in this paper come from open-access datasets and, therefore, should not contain any private sensitive information.

%% file: sections/pruning_figures.tex
\begin{figure*}[t]
    \centering
    \includegraphics[width=1\linewidth]{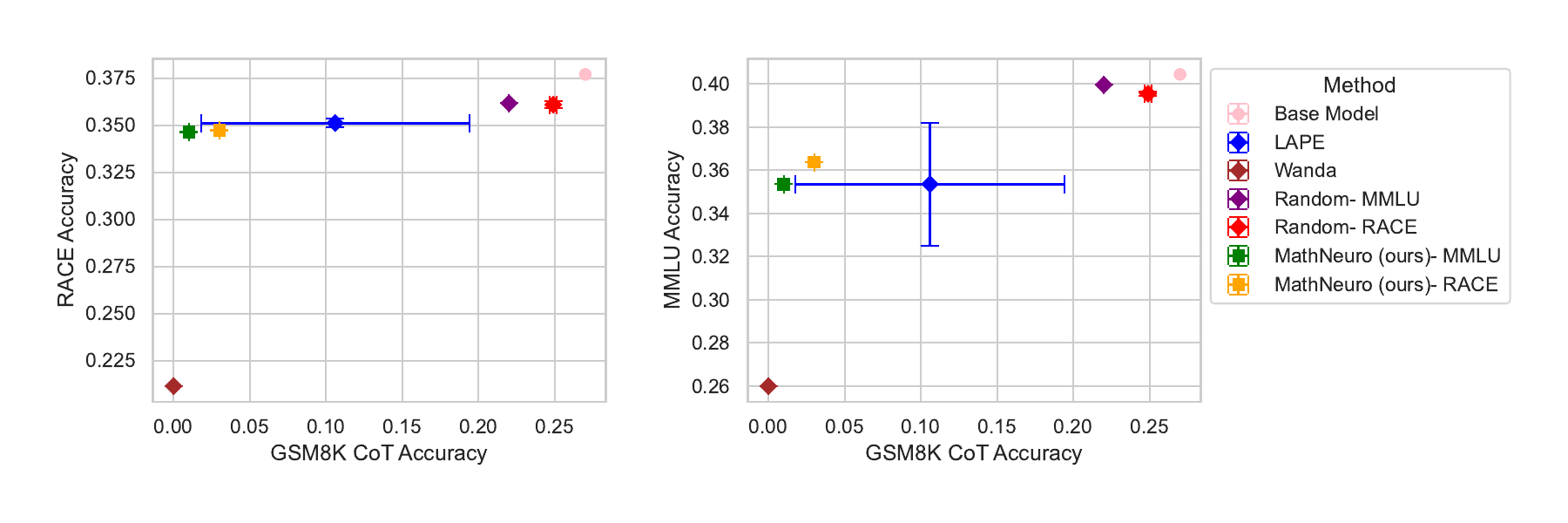}
    \caption{Effect of \textbf{\textit{pruning}} identified parameters on math and non-math performance for Phi 1.5 based on calculating the top 5\% of parameters. Ideal methods should fall in the top left of the plot. MMLU and RACE denote that a point was calculated using MMLU or RACE, respectively, as $\mathcal{D}_{\text{non-math}}$. Horizontal and vertical lines represent 95\% confidence intervals for each point on the plot.}
    \label{fig:MathNeuro_phi15_prune}
\end{figure*}
\begin{figure*}[t]
    \centering
    \includegraphics[width=1\linewidth]{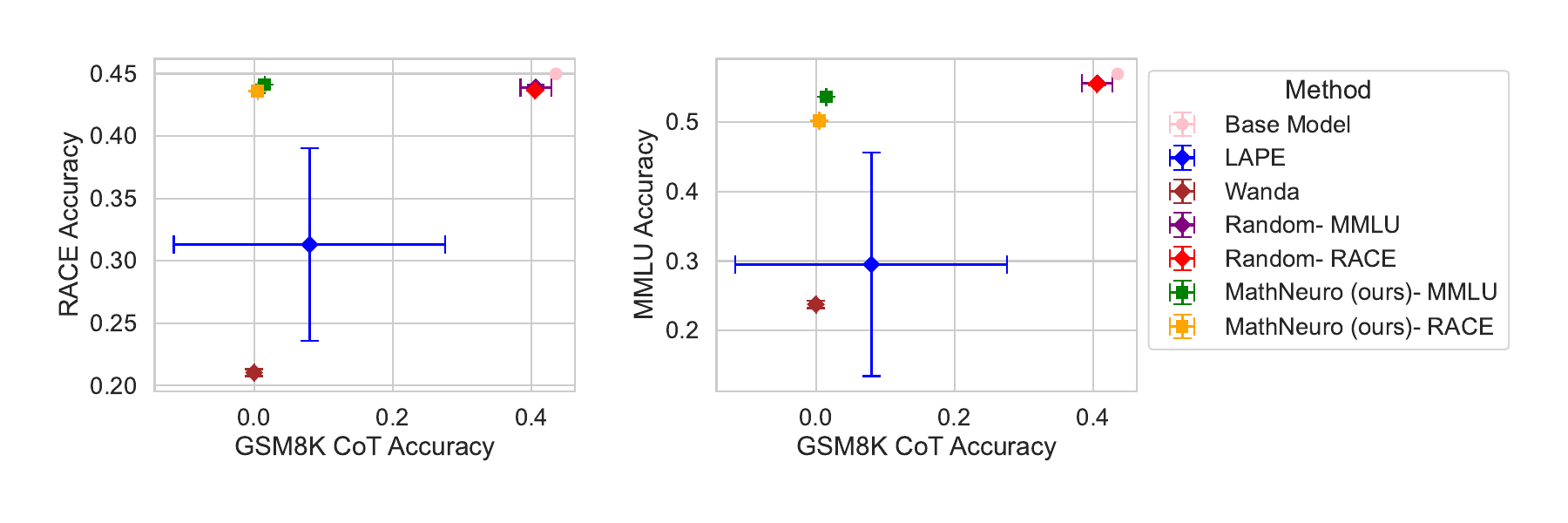}
    \caption{Effect of \textbf{\textit{pruning}} identified parameters on math and non-math performance for Gemma 2 2B IT based on calculating the top 5\% of parameters. Ideal methods should fall in the top left of the plot. MMLU and RACE denote that a point was calculated using MMLU or RACE, respectively, as $\mathcal{D}_{\text{non-math}}$. Horizontal and vertical lines represent 95\% confidence intervals for each point on the plot.}
    \label{fig:MathNeuro_gemma2b_prune}
\end{figure*}
\begin{figure*}[t]
    \centering
    \includegraphics[width=1\linewidth]{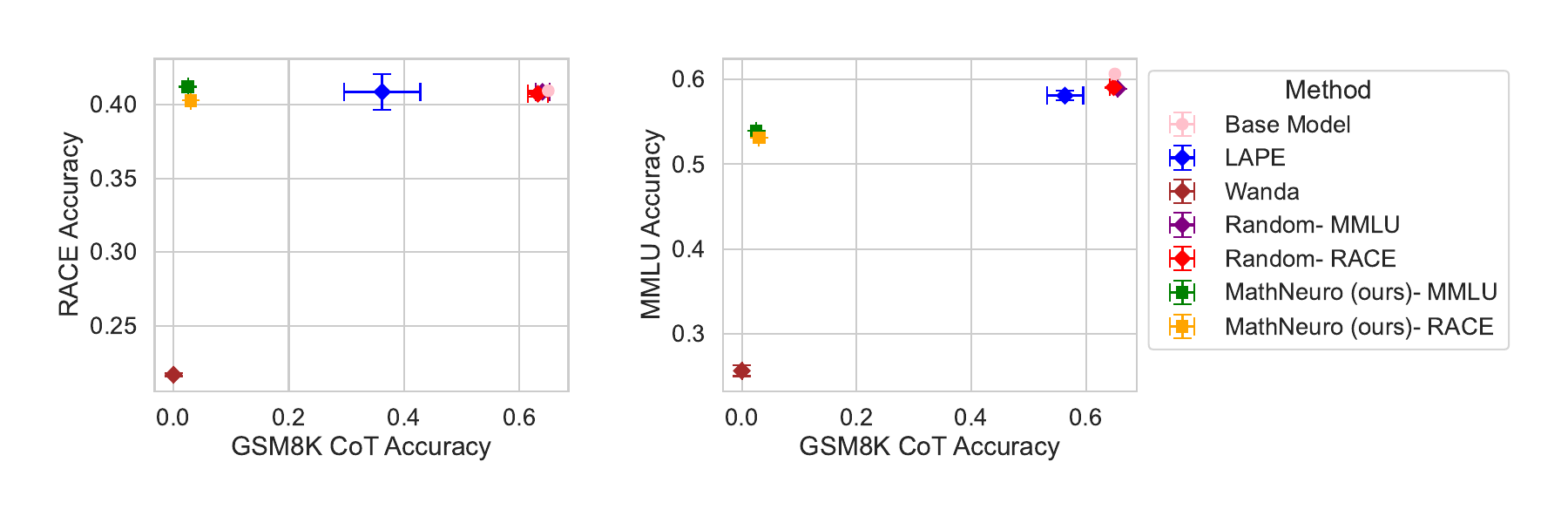}
    \caption{Effect of \textbf{\textit{pruning}} identified parameters on math and non-math performance for Llama 3.2 3B IT based on calculating the top 2.5\% (left) and 1\% (right) of parameters. Ideal methods should fall in the top left of the plot. MMLU and RACE denote that a point was calculated using MMLU or RACE, respectively, as $\mathcal{D}_{\text{non-math}}$. Horizontal and vertical lines represent 95\% confidence intervals for each point on the plot.}
    \label{fig:MathNeuro_llama3b_prune}
\end{figure*}
\begin{figure*}[t]
    \centering
    \includegraphics[width=1\linewidth]{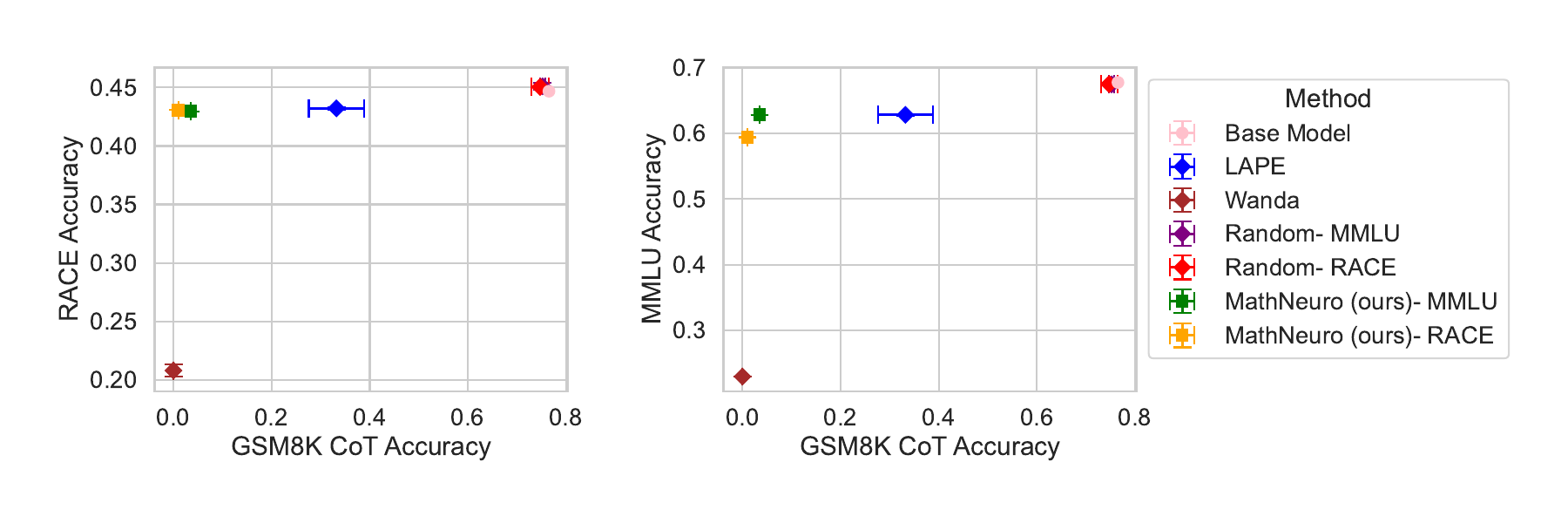}
    \caption{Effect of \textbf{\textit{pruning}} identified parameters on math and non-math performance for Llama 3.1 8B IT based on calculating the top 1\% of parameters. Ideal methods should fall in the top left of the plot. MMLU and RACE denote that a point was calculated using MMLU or RACE, respectively, as $\mathcal{D}_{\text{non-math}}$. Horizontal and vertical lines represent 95\% confidence intervals for each point on the plot.}
    \label{fig:MathNeuro_llama8b_prune}
\end{figure*}
\begin{figure*}[t]
    \centering
    \includegraphics[width=1\linewidth]{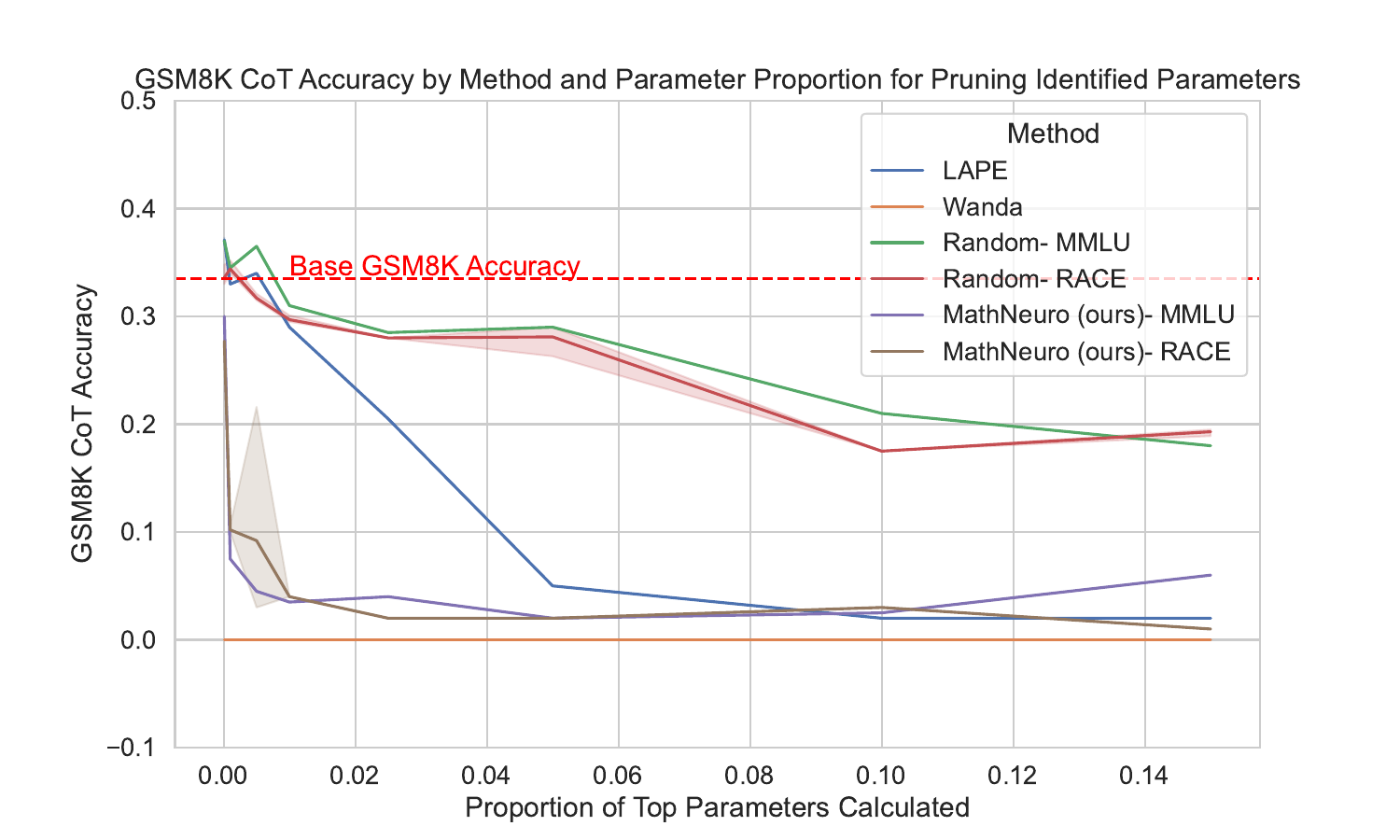}
    \caption{Impact of parameter proportion on GSM8K performance for \textbf{\textit{pruning}} parameters identified by each method for Llama 3.2 1B IT when considering the top .01, .1, .5, 1, 2.5, 5, 10 and 15\% of parameters.}
    \label{fig:pruning_by_param_prop}
\end{figure*}

Figures \ref{fig:MathNeuro_phi15_prune}, \ref{fig:MathNeuro_gemma2b_prune}, \ref{fig:MathNeuro_llama3b_prune} and \ref{fig:MathNeuro_llama8b_prune} show the results for pruning parameters identified by each method for Phi 1.5, Gemma 2 2B IT, Llama 3.2 3B IT, and Llama 3.1 8B IT, respectively. As shown in these figures, the results for each model closely mirror those discussed in Section \ref{deleting_params}, where MathNeuro performs the best at isolating math-specific parameters as shown by destroying GSM8K performance while having low drops in MMLU and RACE performance that are similar to the impact of pruning random model parameters. 
\newline \indent Figure \ref{fig:pruning_by_param_prop} shows the impact of parameter proportion on GSM8K performance when pruning parameters identified by each method for Llama 3.2 1B IT. As shown in the figure, GSM8K performance declines with increasing proportion of parameters considered when using MathNeuro until the parameter proportion reaches .05, at which point the effectiveness of the method levels off. GSM8K performance begins to increase after the top 10\% of parameters are considered due to the top 15\% of model parameters being more invariant across tasks, as shown in Section \ref{param_location_and_number}. For comparison methods, Wanda deletes math performance regardless of parameter proportion, LAPE exhbits a similar performance trend to MathNeuro, and pruning random parameters tends to hurt performance as the proportion of top parameters considered increases, which is expected. The other four models show similar trends when considering different proportions of top parameters. 

%% file: sections/scaling_figures.tex
\begin{figure*}[t]
    \centering
    \includegraphics[width=1\linewidth]{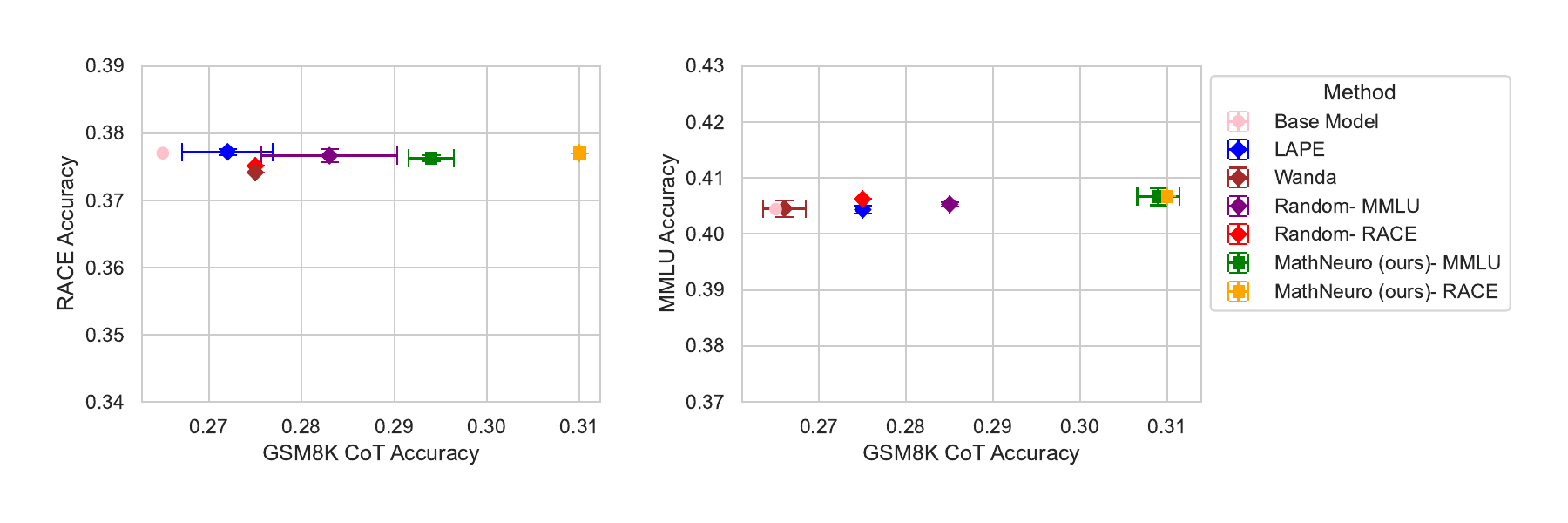}
    \caption{Effect of \textbf{\textit{scaling}} identified parameters by 1.1 on math and non-math performance for Phi 1.5 based on calculating the top .1\% (left) and .01\% (right) of parameters. Ideal methods should fall in the top right of the plot. MMLU and RACE denote that a point was calculated using MMLU or RACE, respectively, as $\mathcal{D}_{\text{non-math}}$. Horizontal and vertical lines represent 95\% confidence intervals for each point on the plot.}
    \label{fig:MathNeuro_phi15_scale}
\end{figure*}
\begin{figure*}[t]
    \centering
    \includegraphics[width=1\linewidth]{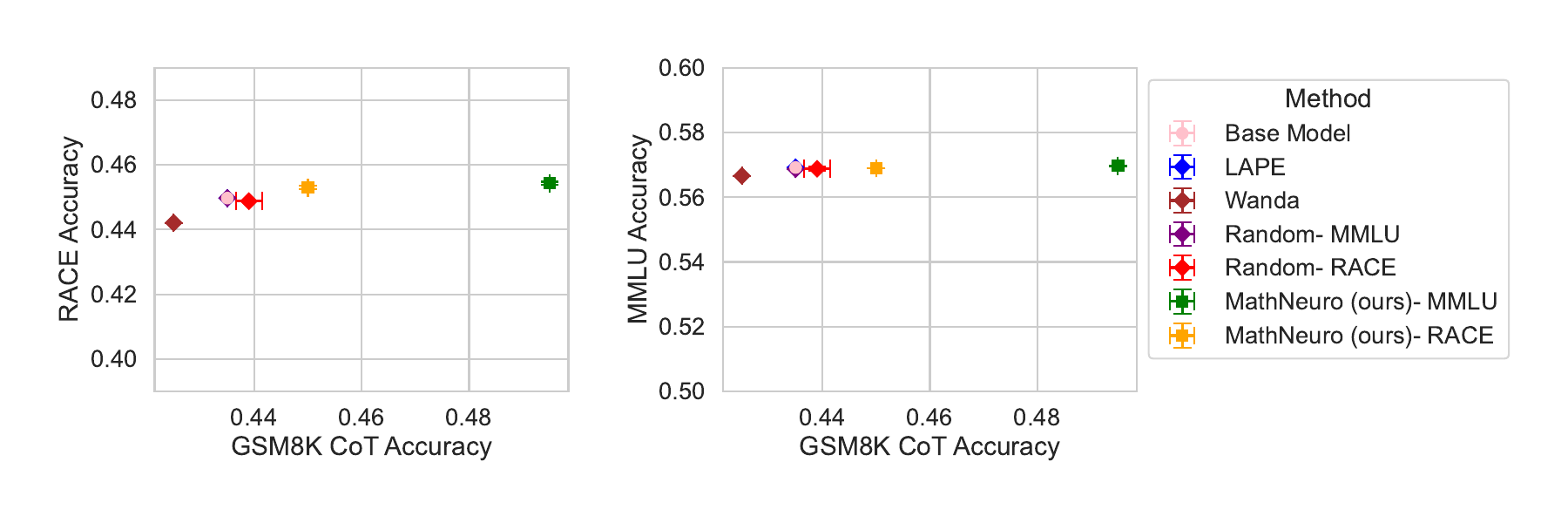}
    \caption{Effect of \textbf{\textit{scaling}} identified parameters by 1.1 on math and non-math performance for Gemma 2 2B IT based on calculating the top 5\% of parameters. Ideal methods should fall in the top right of the plot. MMLU and RACE denote that a point was calculated using MMLU or RACE, respectively, as $\mathcal{D}_{\text{non-math}}$. Horizontal and vertical lines represent 95\% confidence intervals for each point on the plot.}
    \label{fig:MathNeuro_gemma2b_scale}
\end{figure*}
\begin{figure*}[t]
    \centering
    \includegraphics[width=1\linewidth]{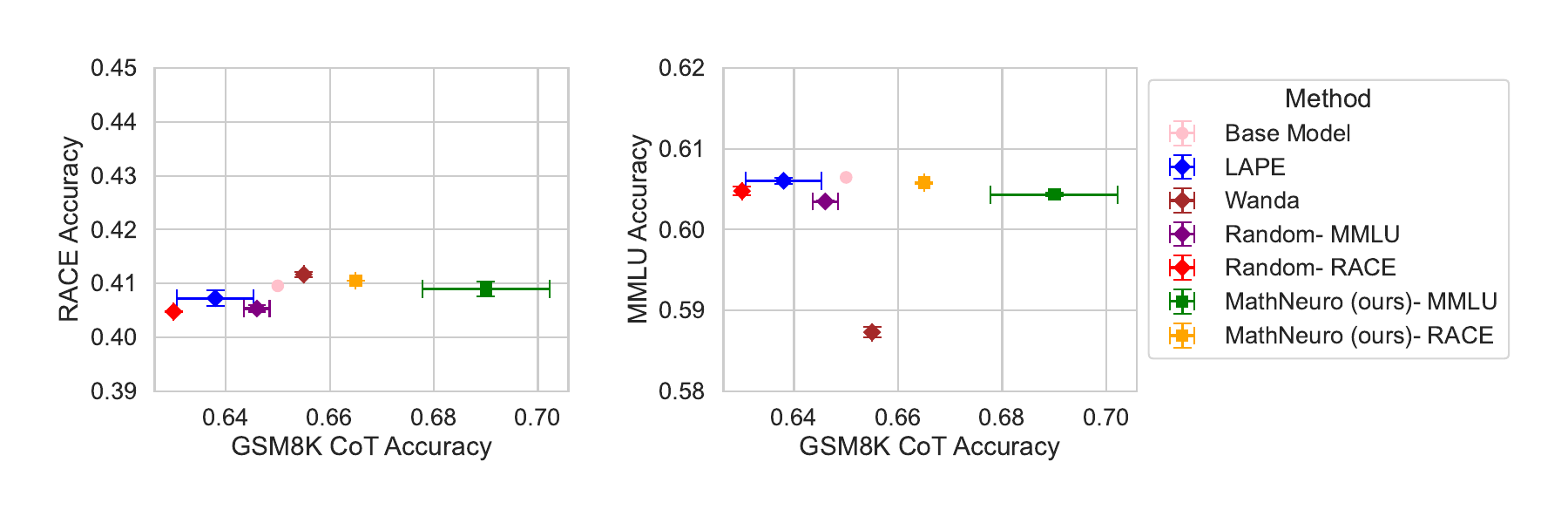}
    \caption{Effect of \textbf{\textit{scaling}} identified parameters by 1.1 on math and non-math performance for Llama 3.2 3B IT based on calculating the top 5\% of parameters. Ideal methods should fall in the top right of the plot. MMLU and RACE denote that a point was calculated using MMLU or RACE, respectively, as $\mathcal{D}_{\text{non-math}}$. Horizontal and vertical lines represent 95\% confidence intervals for each point on the plot.}
    \label{fig:MathNeuro_llama3b_scale}
\end{figure*}
\begin{figure*}[t]
    \centering
    \includegraphics[width=1\linewidth]{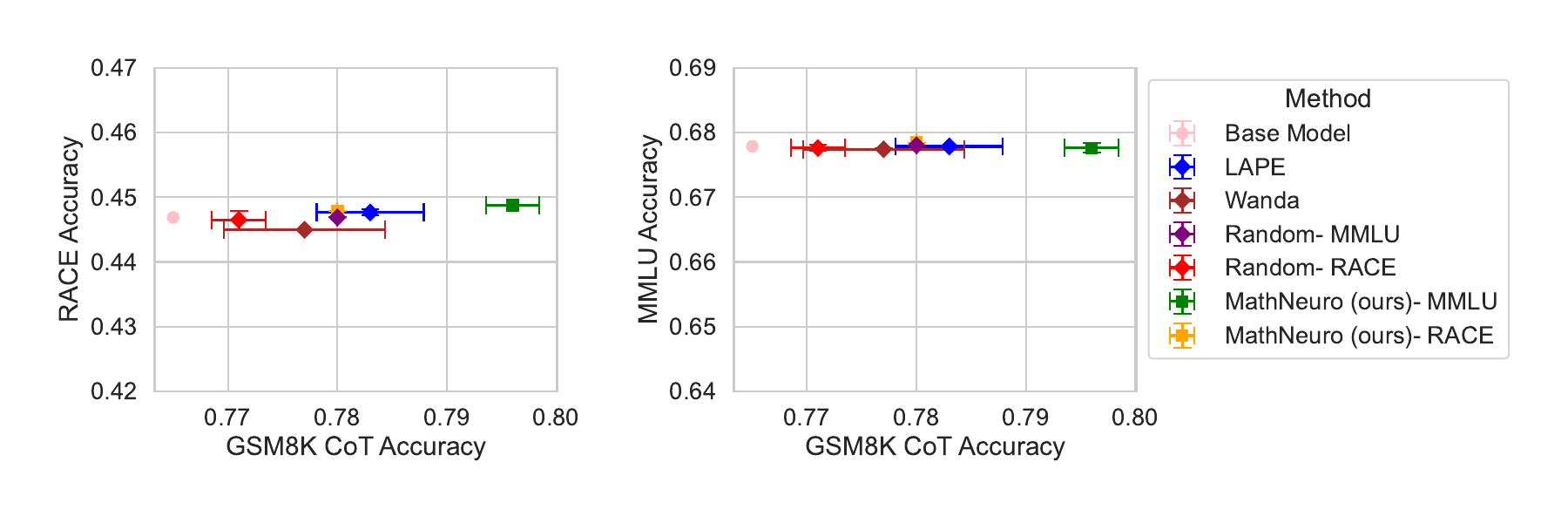}
    \caption{Effect of \textbf{\textit{scaling}} identified parameters by 1.01 on math and non-math performance for Llama 3.1 8B IT based on calculating the top .5\% of parameters. Ideal methods should fall in the top right of the plot. MMLU and RACE denote that a point was calculated using MMLU or RACE, respectively, as $\mathcal{D}_{\text{non-math}}$. Horizontal and vertical lines represent 95\% confidence intervals for each point on the plot.}
    \label{fig:MathNeuro_llama8b_scale}
\end{figure*}

Figures \ref{fig:MathNeuro_phi15_scale}, \ref{fig:MathNeuro_gemma2b_scale}, \ref{fig:MathNeuro_llama3b_scale} and \ref{fig:MathNeuro_llama8b_scale} show the results for scaling parameters identified by each method for Phi 1.5, Gemma 2 2B IT, Llama 3.2 3B IT, and Llama 3.1 8B IT, respectively. As shown in these figures, the results for each model closely mirror those discussed in Section \ref{scaling_params}, where scaling parameters identified by MathNeuro consistently increases GSM8K performance by 3-6 percentage points across models, representing a 4-17\% overall increase in math performance depending on the model. 

%% file: sections/one_sample_pruning_figures.tex
\begin{figure*}[t]
    \centering
    \includegraphics[width=1\linewidth]{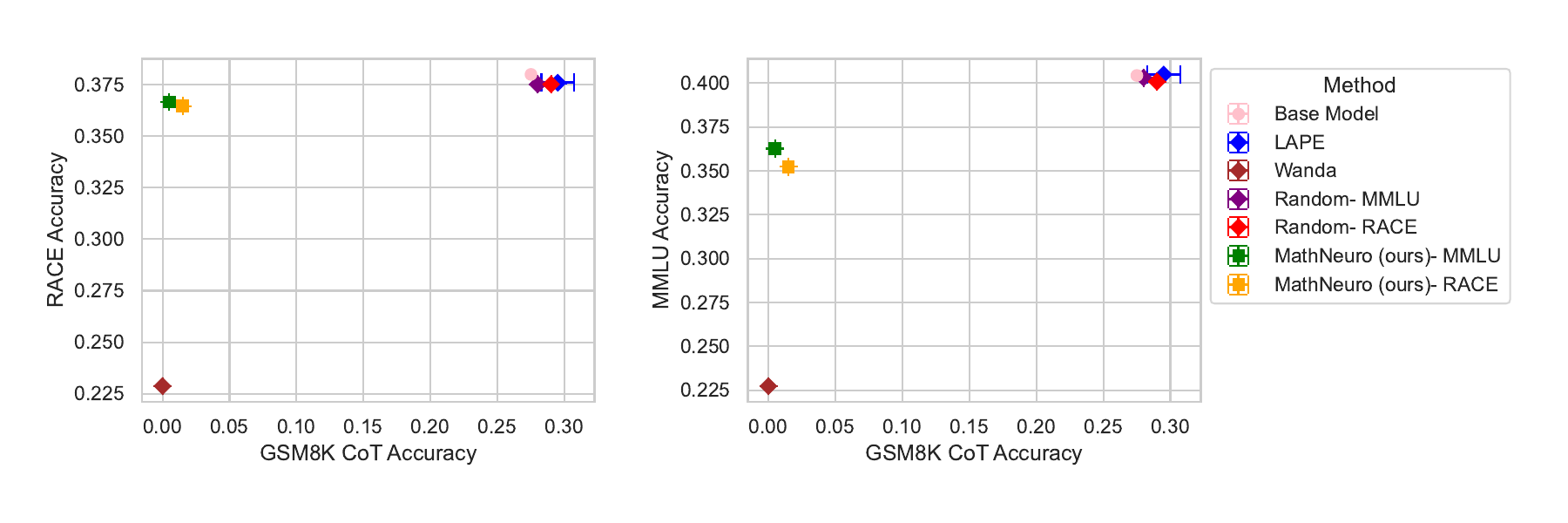}
    \caption{Effect of \textbf{\textit{pruning}} identified parameters on math and non-math performance for Phi 1.5 for calculating the top .5\% of parameters \textbf{\textit{based on one sample}}. Ideal methods should fall in the top left of the plot. MMLU and RACE denote that a point was calculated using MMLU or RACE, respectively, as $\mathcal{D}_{\text{non-math}}$. Horizontal and vertical lines represent 95\% confidence intervals for each point on the plot.}
    \label{fig:MathNeuro_phi15_prune_one_sample}
\end{figure*}
\begin{figure*}[t]
    \centering
    \includegraphics[width=1\linewidth]{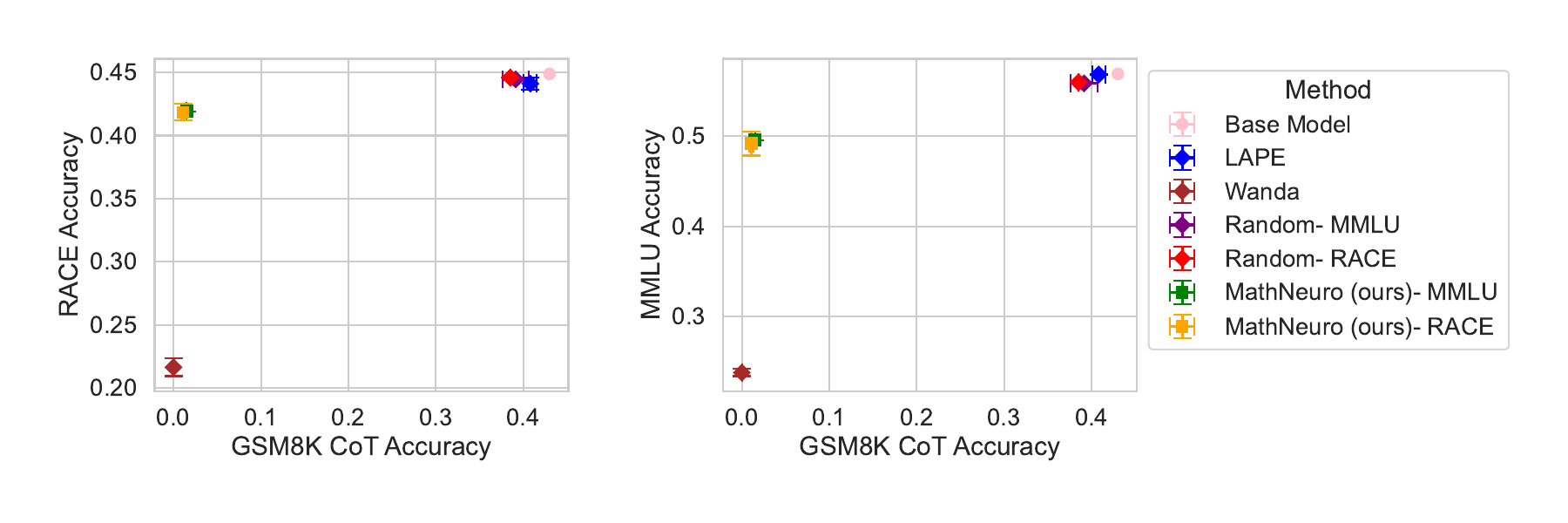}
    \caption{Effect of \textbf{\textit{pruning}} identified parameters on math and non-math performance for Gemma 2 2B IT for calculating the top 2.5\% of parameters \textbf{\textit{based on one sample}}. Ideal methods should fall in the top left of the plot. MMLU and RACE denote that a point was calculated using MMLU or RACE, respectively, as $\mathcal{D}_{\text{non-math}}$. Horizontal and vertical lines represent 95\% confidence intervals for each point on the plot.}
    \label{fig:MathNeuro_gemma2b_prune_one_sample}
\end{figure*}
\begin{figure*}[t]
    \centering
    \includegraphics[width=1\linewidth]{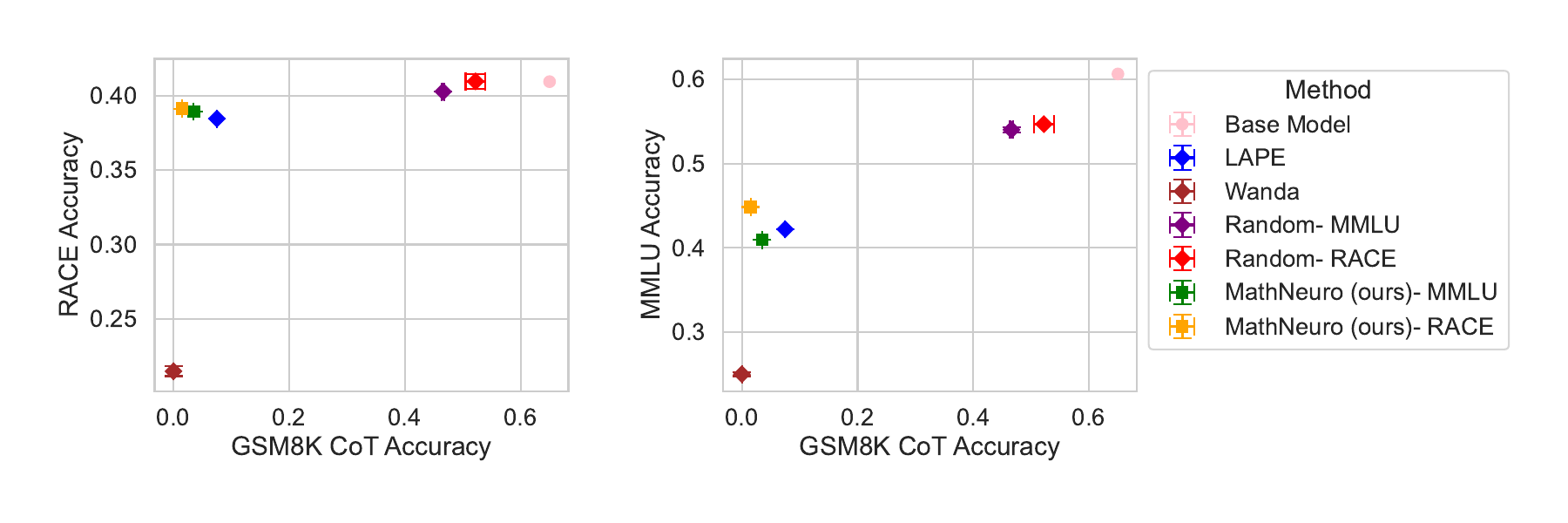}
    \caption{Effect of \textbf{\textit{pruning}} identified parameters on math and non-math performance for Llama 3.2 3B IT for calculating the top 10\% of parameters \textbf{\textit{based on one sample}}. Ideal methods should fall in the top left of the plot. MMLU and RACE denote that a point was calculated using MMLU or RACE, respectively, as $\mathcal{D}_{\text{non-math}}$. Horizontal and vertical lines represent 95\% confidence intervals for each point on the plot.}
    \label{fig:MathNeuro_llama3b_prune_one_sample}
\end{figure*}
\begin{figure*}[t]
    \centering
    \includegraphics[width=1\linewidth]{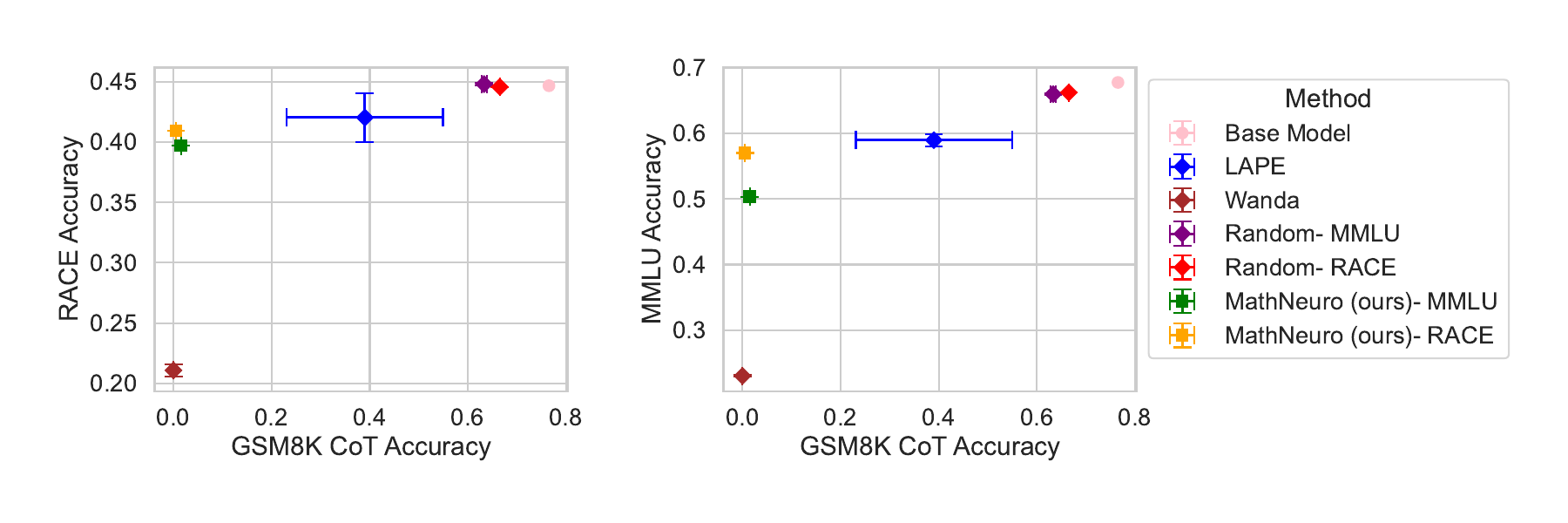}
    \caption{Effect of \textbf{\textit{pruning}} identified parameters on math and non-math performance for Llama 3.1 8B IT for calculating the top 5\% of parameters \textbf{\textit{based on one sample}}. Ideal methods should fall in the top left of the plot. MMLU and RACE denote that a point was calculated using MMLU or RACE, respectively, as $\mathcal{D}_{\text{non-math}}$. Horizontal and vertical lines represent 95\% confidence intervals for each point on the plot.}
    \label{fig:MathNeuro_llama8b_prune_one_sample}
\end{figure*}

Figures \ref{fig:MathNeuro_phi15_prune_one_sample}, \ref{fig:MathNeuro_gemma2b_prune_one_sample}, \ref{fig:MathNeuro_llama3b_prune_one_sample} and \ref{fig:MathNeuro_llama8b_prune_one_sample} show the results for pruning parameters identified by each method for Phi 1.5, Gemma 2 2B IT, Llama 3.2 3B IT, and Llama 3.1 8B IT, respectively, when calculating parameter importance based on a single sample. As shown in these figures, the results for each model closely mirror those discussed in Section \ref{single_sample}, where MathNeuro still performs the best at isolating math-specific parameters as shown by destroying GSM8K performance while having lower drops in MMLU and RACE performance than Wanda. However, as reported in Section \ref{single_sample}, these results suggest additional samples help the method more effectively isolate math-specific parameters because the non-math drops in performance are larger than those shown in Appendix \ref{sec:pruning_figures}, where we used 500 samples to calculate parameter importance. 

%% file: sections/one_sample_scaling_figures.tex
\begin{figure*}[t]
    \centering
    \includegraphics[width=1\linewidth]{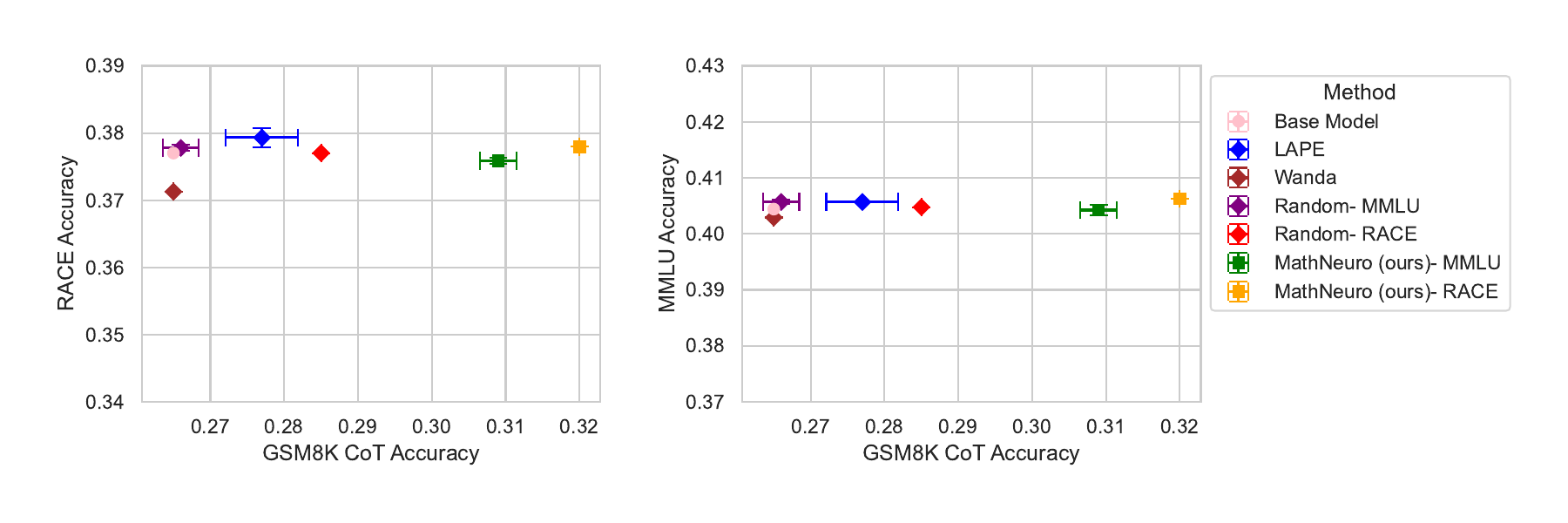}
    \caption{Effect of \textbf{\textit{scaling}} identified parameters by 1.1 on math and non-math performance for Phi 1.5 for calculating the top .1\% of parameters \textbf{\textit{based on one sample}}. Ideal methods should fall in the top right of the plot. MMLU and RACE denote that a point was calculated using MMLU or RACE, respectively, as $\mathcal{D}_{\text{non-math}}$. Horizontal and vertical lines represent 95\% confidence intervals for each point on the plot.}
    \label{fig:MathNeuro_phi15_scale_one_sample}
\end{figure*}
\begin{figure*}[t]
    \centering
    \includegraphics[width=1\linewidth]{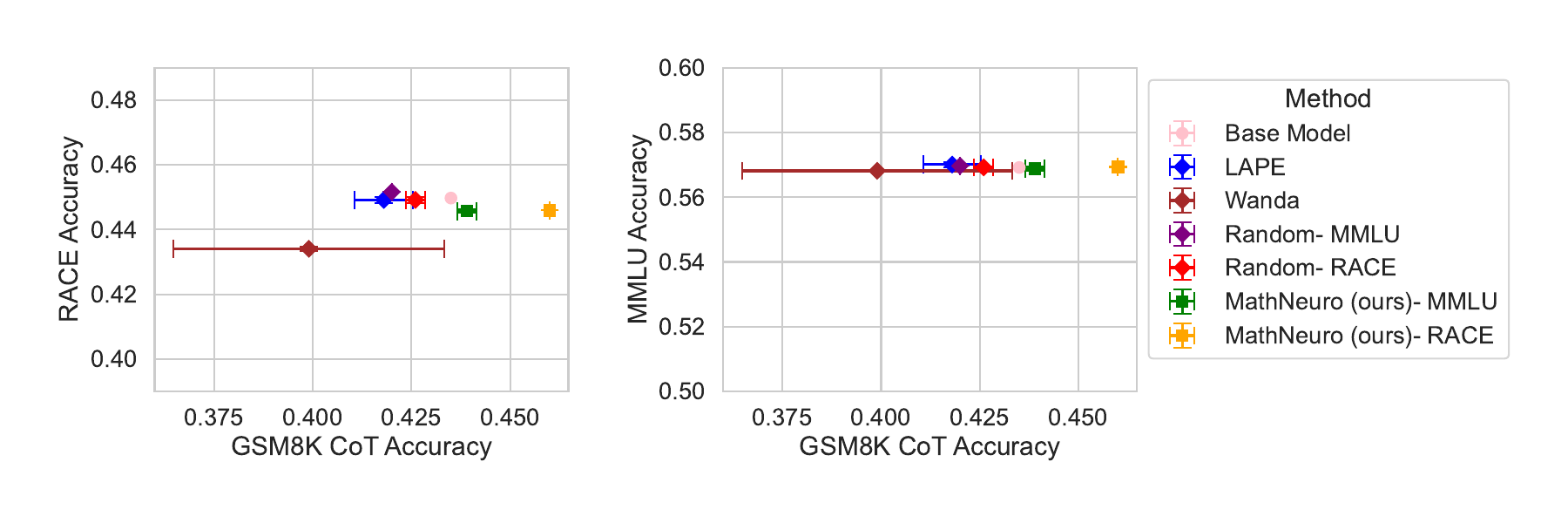}
    \caption{Effect of \textbf{\textit{scaling}} identified parameters by 1.1 on math and non-math performance for Gemma 2 2B IT for calculating the top 2.5\% of parameters \textbf{\textit{based on one sample}}. Ideal methods should fall in the top right of the plot. MMLU and RACE denote that a point was calculated using MMLU or RACE, respectively, as $\mathcal{D}_{\text{non-math}}$. Horizontal and vertical lines represent 95\% confidence intervals for each point on the plot.}
    \label{fig:MathNeuro_gemma2b_scale_one_sample}
\end{figure*}
\begin{figure*}[t]
    \centering
    \includegraphics[width=1\linewidth]{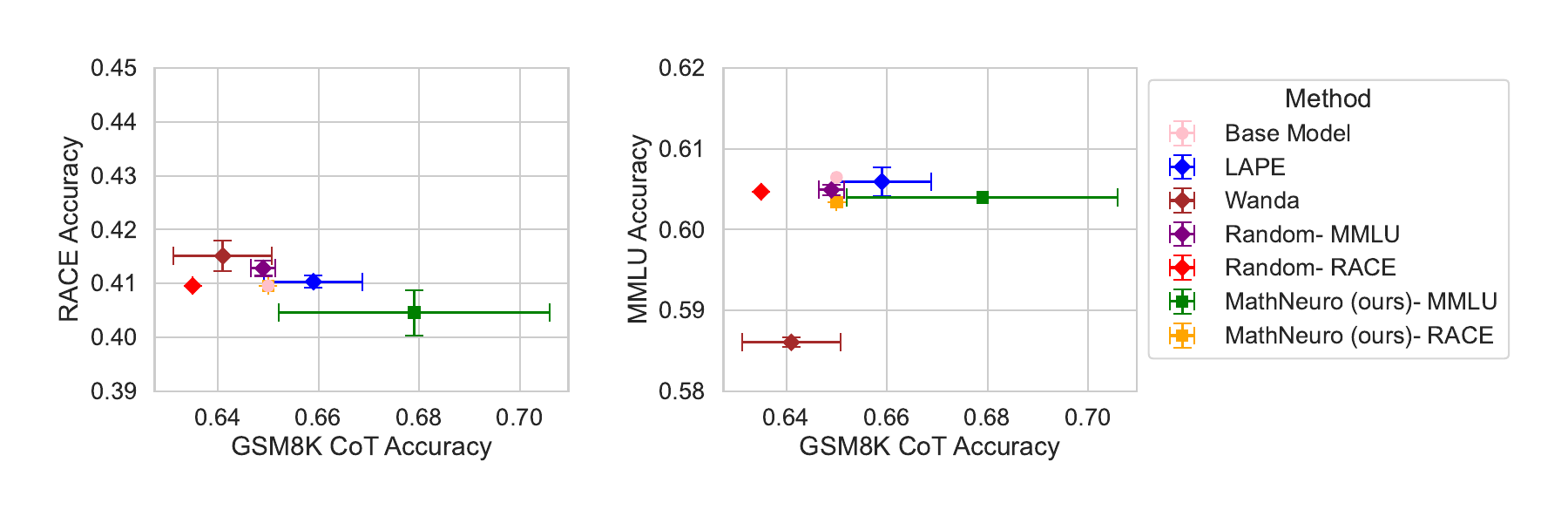}
    \caption{Effect of \textbf{\textit{scaling}} identified parameters by 1.1 on math and non-math performance for Llama 3.2 3B IT for calculating the top 5\% of parameters \textbf{\textit{based on one sample}}. Ideal methods should fall in the top right of the plot. MMLU and RACE denote that a point was calculated using MMLU or RACE, respectively, as $\mathcal{D}_{\text{non-math}}$. Horizontal and vertical lines represent 95\% confidence intervals for each point on the plot.}
    \label{fig:MathNeuro_llama3b_scale_one_sample}
\end{figure*}
\begin{figure*}[t]
    \centering
    \includegraphics[width=1\linewidth]{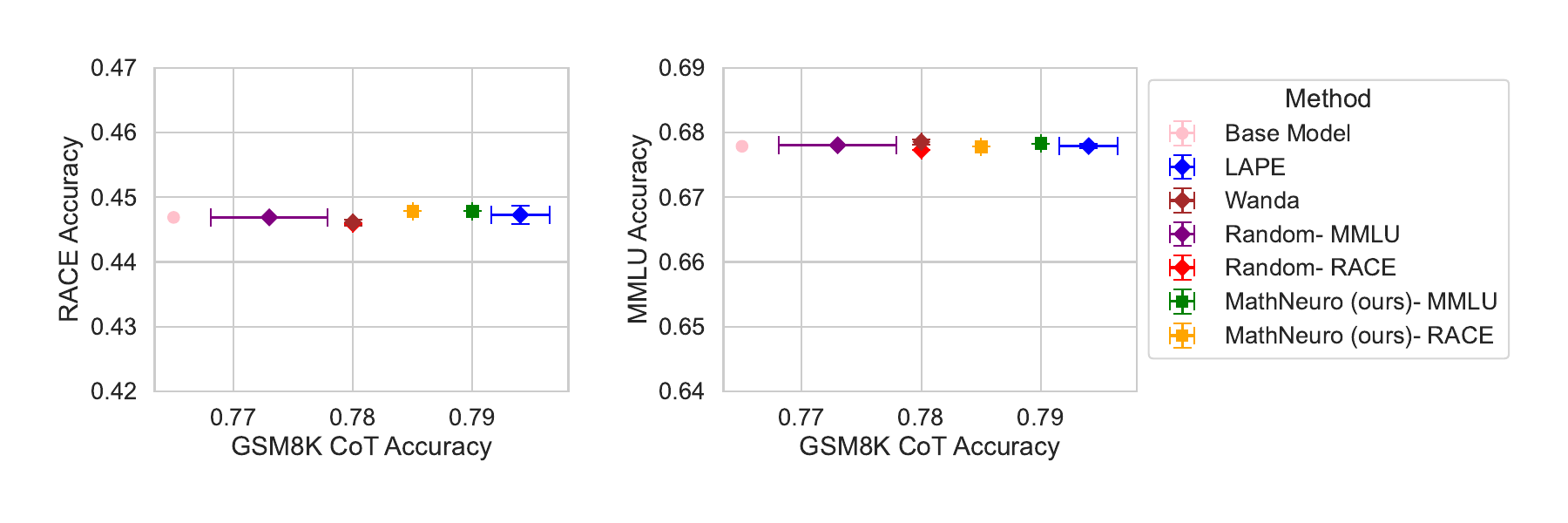}
    \caption{Effect of \textbf{\textit{scaling}} identified parameters by 1.01 on math and non-math performance for Llama 3.1 8B IT for calculating the top 1\% of parameters \textbf{\textit{based on one sample}}. Ideal methods should fall in the top right of the plot. MMLU and RACE denote that a point was calculated using MMLU or RACE, respectively, as $\mathcal{D}_{\text{non-math}}$. Horizontal and vertical lines represent 95\% confidence intervals for each point on the plot.}
\label{fig:MathNeuro_llama8b_scale_one_sample}
\end{figure*}

Figures \ref{fig:MathNeuro_phi15_scale_one_sample}, \ref{fig:MathNeuro_gemma2b_scale_one_sample}, \ref{fig:MathNeuro_llama3b_scale_one_sample} and \ref{fig:MathNeuro_llama8b_scale_one_sample} show the results for scaling parameters identified by each method for Phi 1.5, Gemma 2 2B IT, Llama 3.2 3B IT, and Llama 3.1 8B IT, respectively, when calculating parameter importance based on one sample. As shown in these figures, the results for each model closely mirror those discussed in Section \ref{single_sample}, where scaling parameters identified by MathNeuro using a single sample consistently increases GSM8K performance across models. These increases are either similar to those reported in Appendix \ref{sec:scaling_figures}, or smaller but still meaningful. For some models, the comparison methods can increase GSM8K performance when calculating parameter importance based off a single sample, but MathNeuro is the only method for which a meaningful positive increase is consistent across models. 


%% file: sections/unseen_tasks.tex
\begin{figure*}[t]
    \centering
    \includegraphics[width=1\linewidth]{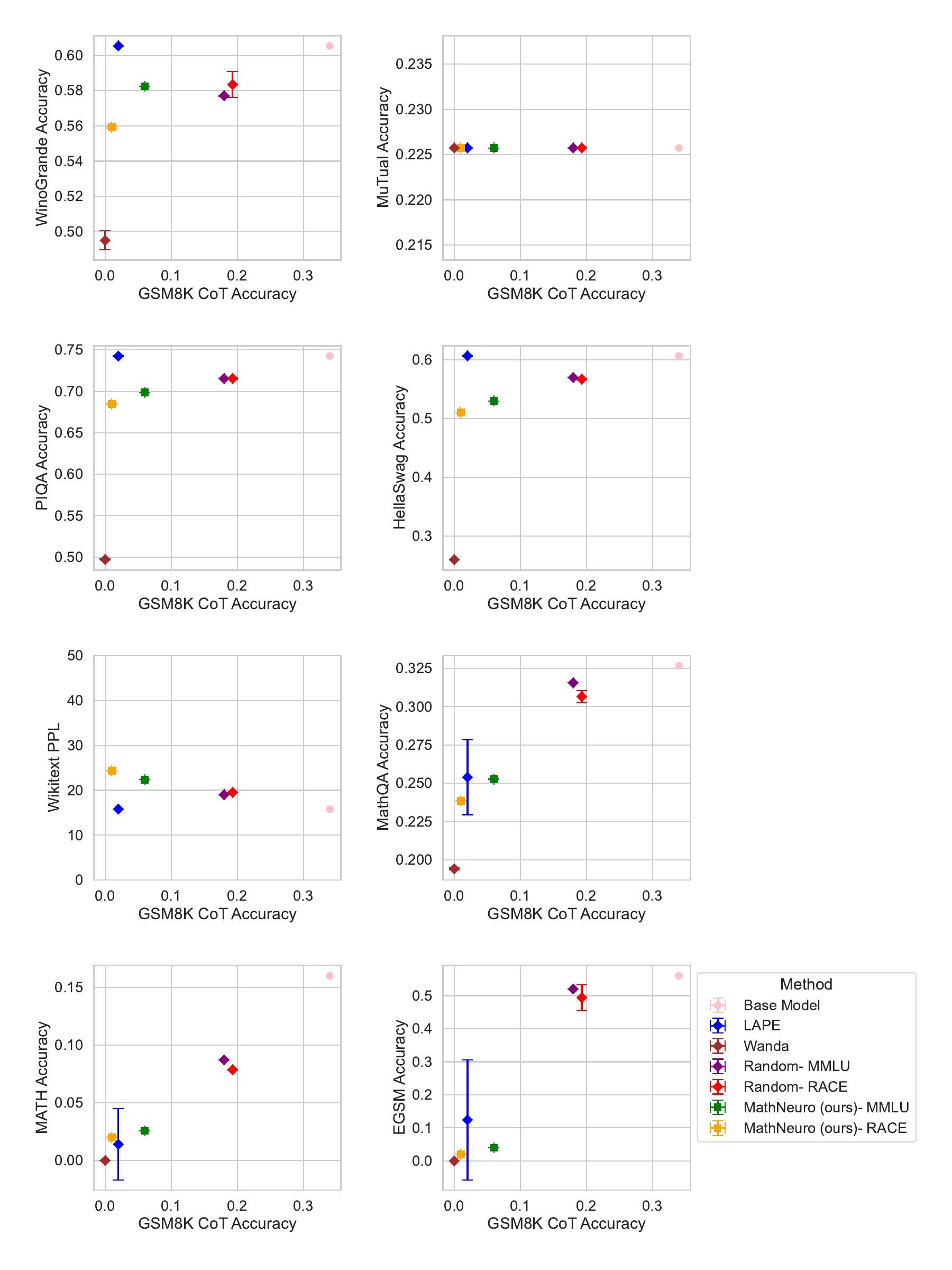}
    \caption{Effect of \textbf{\textit{pruning}} identified parameters on performance for unseen math and non-math tasks for Llama 3.2 1B IT when using GSM8K as $\mathcal{D}_{\text{math}}$ and MMLU or Race as  $\mathcal{D}_{\text{non-math}}$. Ideal methods for the first two rows of figures should fall in the top left of the plot, while ideal methods for the last two rows of figures should fall in the bottom left of the plot. Wanda results are not pictured in the Wikitext figures because PPL increased dramatically when using this method. MMLU and RACE denote that a point was calculated using MMLU or RACE, respectively, as $\mathcal{D}_{\text{non-math}}$. Horizontal and vertical lines represent 95\% confidence intervals for each point on the plot.}
    \label{fig:MathNeuro_llama1b_comprehensive}
\end{figure*}
\begin{figure*}[t]
    \centering
    \includegraphics[width=1\linewidth]{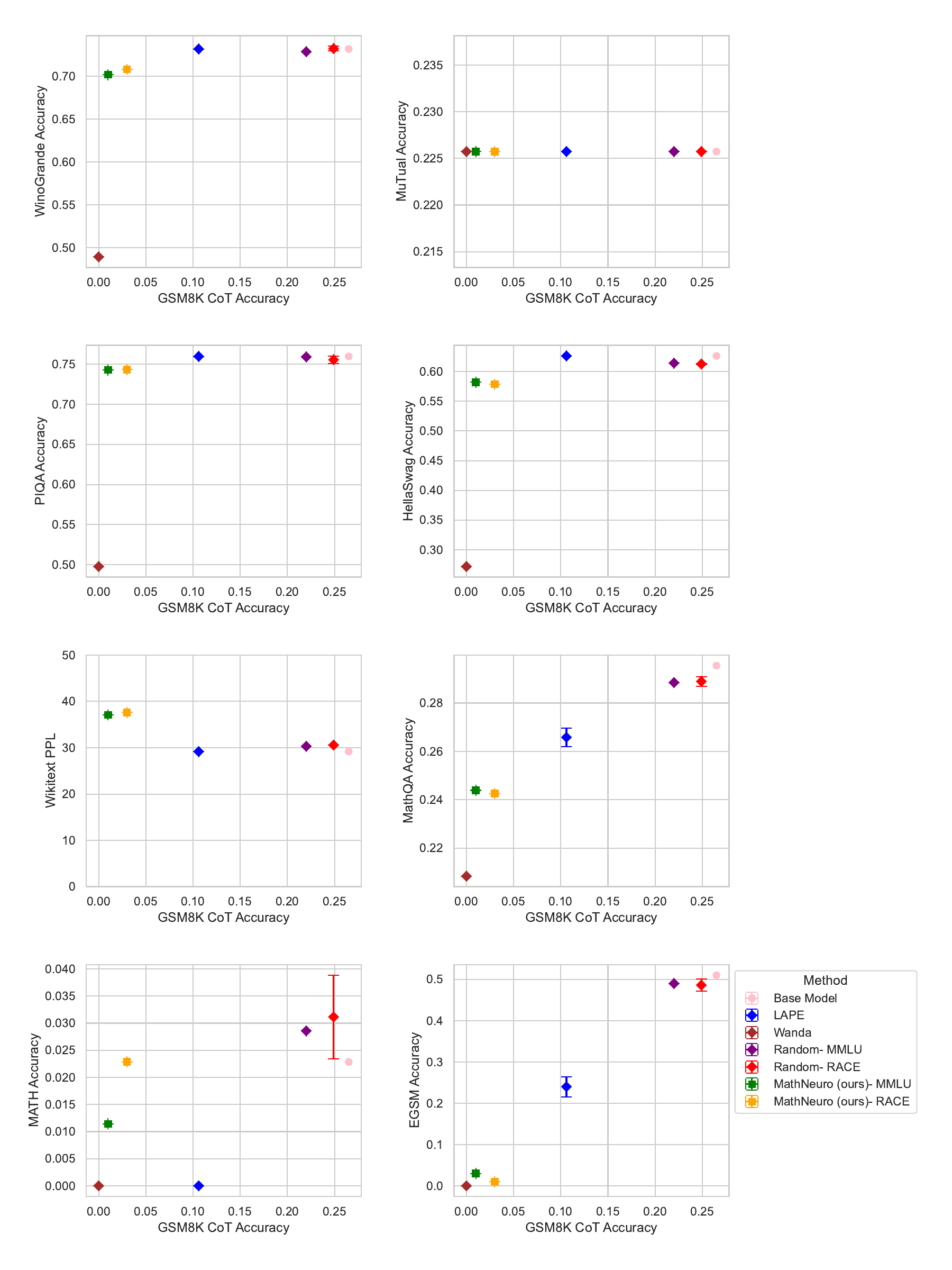}
    \caption{Effect of \textbf{\textit{pruning}} identified parameters on performance for unseen math and non-math tasks for Phi 1.5 when using GSM8K as $\mathcal{D}_{\text{math}}$ and MMLU or Race as  $\mathcal{D}_{\text{non-math}}$. Ideal methods for the first two rows of figures should fall in the top left of the plot, while ideal methods for the last two rows of figures should fall in the bottom left of the plot. Wanda results are not pictured in the Wikitext figures because PPL increased dramatically when using this method. MMLU and RACE denote that a point was calculated using MMLU or RACE, respectively, as $\mathcal{D}_{\text{non-math}}$. Horizontal and vertical lines represent 95\% confidence intervals for each point on the plot.}
    \label{fig:MathNeuro_phi15_comprehensive}
\end{figure*}
\begin{figure*}[t]
    \centering
    \includegraphics[width=1\linewidth]{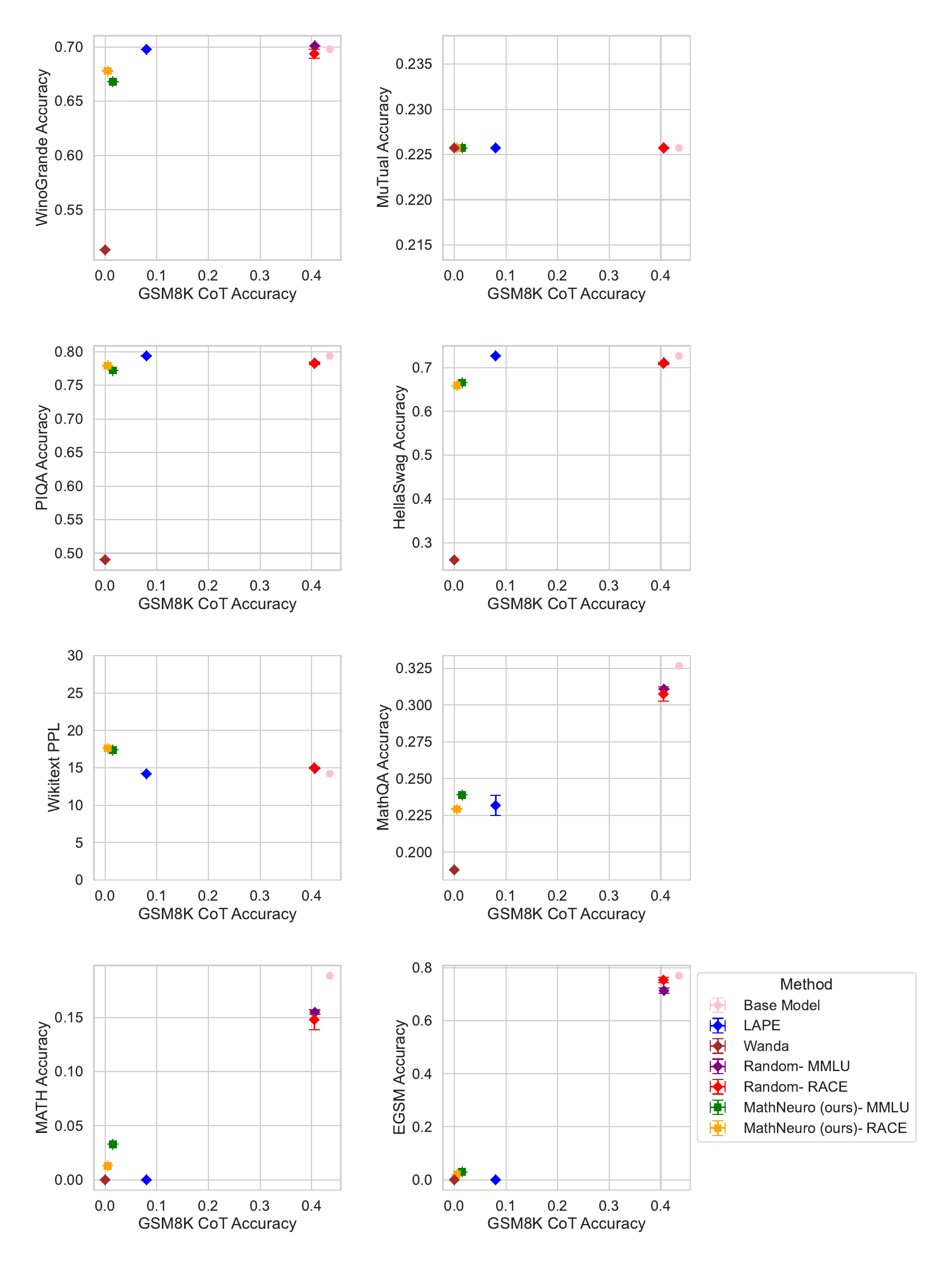}
    \caption{Effect of \textbf{\textit{pruning}} identified parameters on performance for unseen math and non-math tasks for Gemma 2 2B IT when using GSM8K as $\mathcal{D}_{\text{math}}$ and MMLU or Race as  $\mathcal{D}_{\text{non-math}}$. Ideal methods for the first two rows of figures should fall in the top left of the plot, while ideal methods for the last two rows of figures should fall in the bottom left of the plot. Wanda results are not pictured in the Wikitext figures because PPL increased dramatically when using this method. MMLU and RACE denote that a point was calculated using MMLU or RACE, respectively, as $\mathcal{D}_{\text{non-math}}$. Horizontal and vertical lines represent 95\% confidence intervals for each point on the plot.}
    \label{fig:MathNeuro_gemma_comprehensive}
\end{figure*}
\begin{figure*}[t]
    \centering
    \includegraphics[width=1\linewidth]{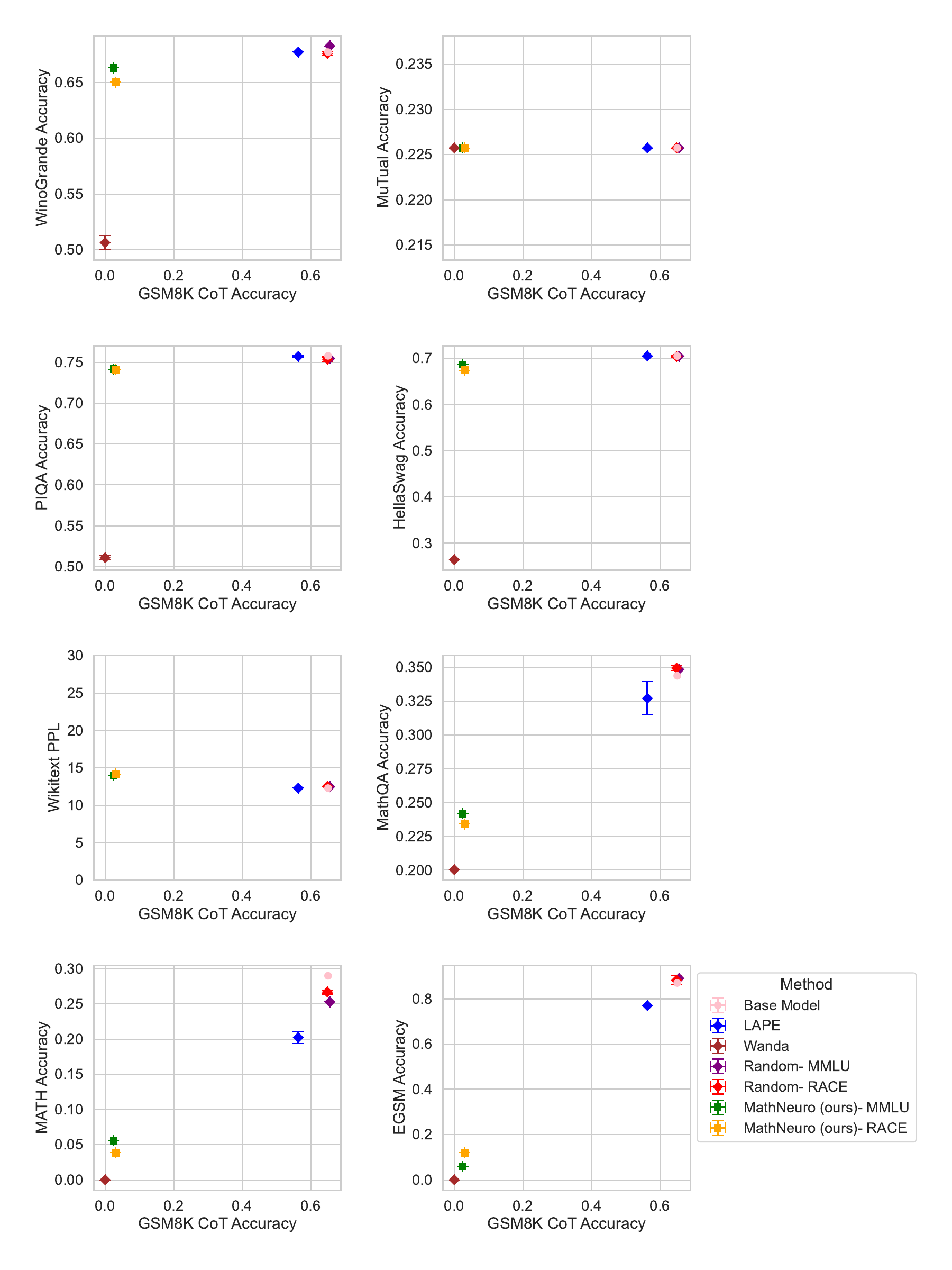}
    \caption{Effect of \textbf{\textit{pruning}} identified parameters on performance for unseen math and non-math tasks for Llama 3.2 3B IT when using GSM8K as $\mathcal{D}_{\text{math}}$ and MMLU or Race as  $\mathcal{D}_{\text{non-math}}$. Ideal methods for the first two rows of figures should fall in the top left of the plot, while ideal methods for the last two rows of figures should fall in the bottom left of the plot. Wanda results are not pictured in the Wikitext figures because PPL increased dramatically when using this method. MMLU and RACE denote that a point was calculated using MMLU or RACE, respectively, as $\mathcal{D}_{\text{non-math}}$. Horizontal and vertical lines represent 95\% confidence intervals for each point on the plot.}
    \label{fig:MathNeuro_llama3b_comprehensive}
\end{figure*}
\begin{figure*}[t]
    \centering
    \includegraphics[width=1\linewidth]{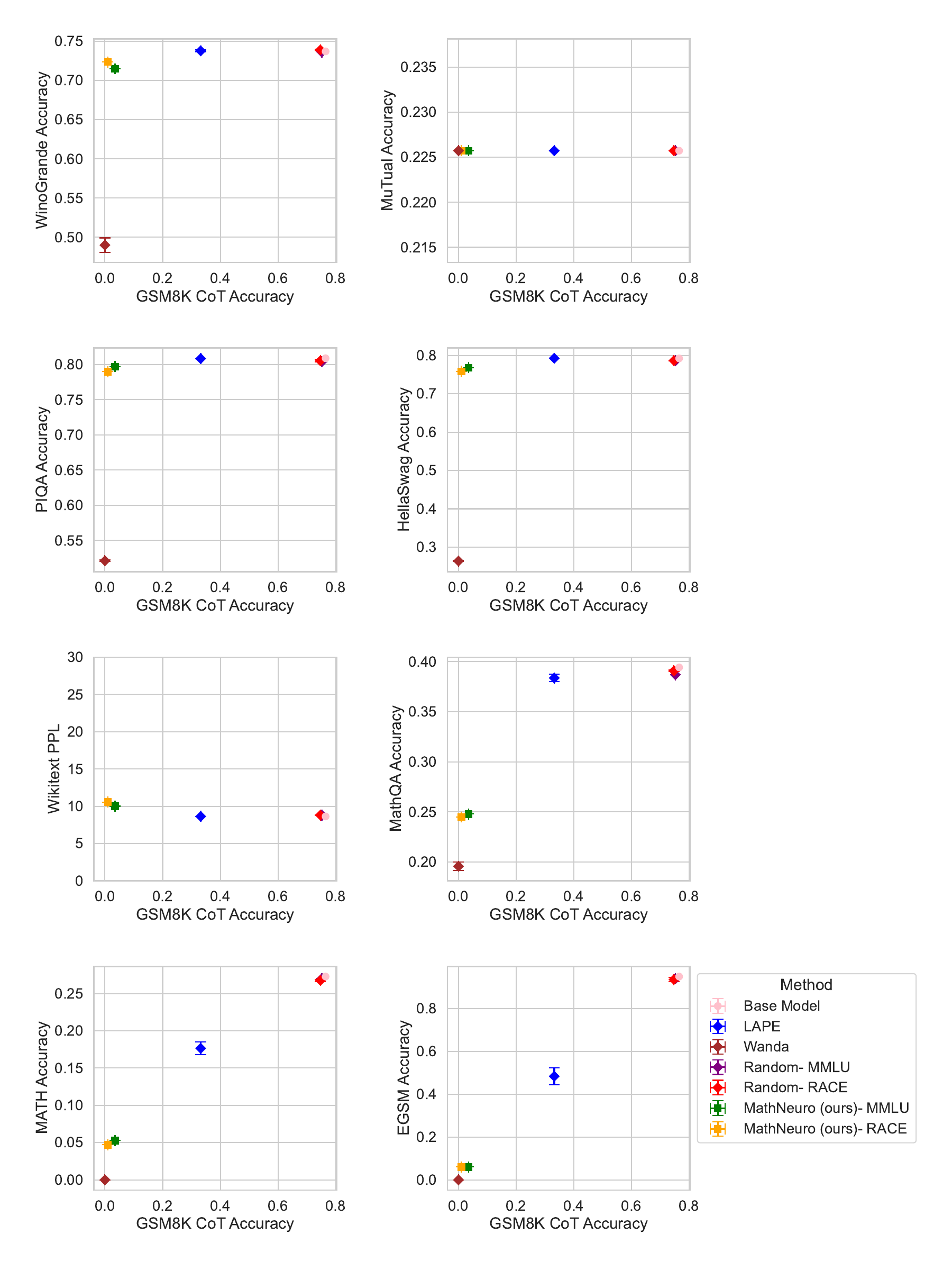}
    \caption{Effect of \textbf{\textit{pruning}} identified parameters on performance for unseen math and non-math tasks for Llama 3.1 8B IT when using GSM8K as $\mathcal{D}_{\text{math}}$ and MMLU or Race as  $\mathcal{D}_{\text{non-math}}$. Ideal methods for the first two rows of figures should fall in the top left of the plot, while ideal methods for the last two rows of figures should fall in the bottom left of the plot. Wanda results are not pictured in the Wikitext figures because PPL increased dramatically when using this method. MMLU and RACE denote that a point was calculated using MMLU or RACE, respectively, as $\mathcal{D}_{\text{non-math}}$. Horizontal and vertical lines represent 95\% confidence intervals for each point on the plot.}
    \label{fig:MathNeuro_llama8b_comprehensive}
\end{figure*}

As discussed in Section \ref{unseen_tasks_experiments}, Figures \ref{fig:MathNeuro_llama1b_comprehensive}, \ref{fig:MathNeuro_phi15_comprehensive}, \ref{fig:MathNeuro_gemma_comprehensive}, \ref{fig:MathNeuro_llama3b_comprehensive}, and \ref{fig:MathNeuro_llama8b_comprehensive} display the results for evaluating unseen task performance after pruning parameters identified by MathNeuro using GSM8K as $\mathcal{D}_{\text{math}}$ and MMLU or Race as  $\mathcal{D}_{\text{non-math}}$. In these figures, MuTual performance is unimpaired regardless of the parameter identification method used, suggesting that performance on this task is consistently equal to random guessing regardless of which parameters are pruned. All tasks are implemented using their default implementation in the Eleuther AI LM Evaluation Harness \cite{gao_framework_2024}. For MATH, we use Eleuther's implementation of the 4-shot CoT Minerva prompt and for EGSM we use a fork of the Eleuther AI LM Evaluation Harness where we implemented the task using GSM8K's 8-shot CoT prompt given they are both grade school math datasets. For all tasks except MATH and EGSM, we run evaluations using the full testing split for each task given they are in multiple-choice format. For MATH and EGSM, which require long-form responses, we follow our other experiments by using the same set of random samples from each dataset for experimental efficiency. For MATH, we use 700 samples and for EGSM we use 100 due to its smaller size. 

%% file: sections/math_results.tex
\subsection{MATH Pruning Experiments}
\begin{figure*}[t]
    \centering
    \includegraphics[width=1\linewidth]{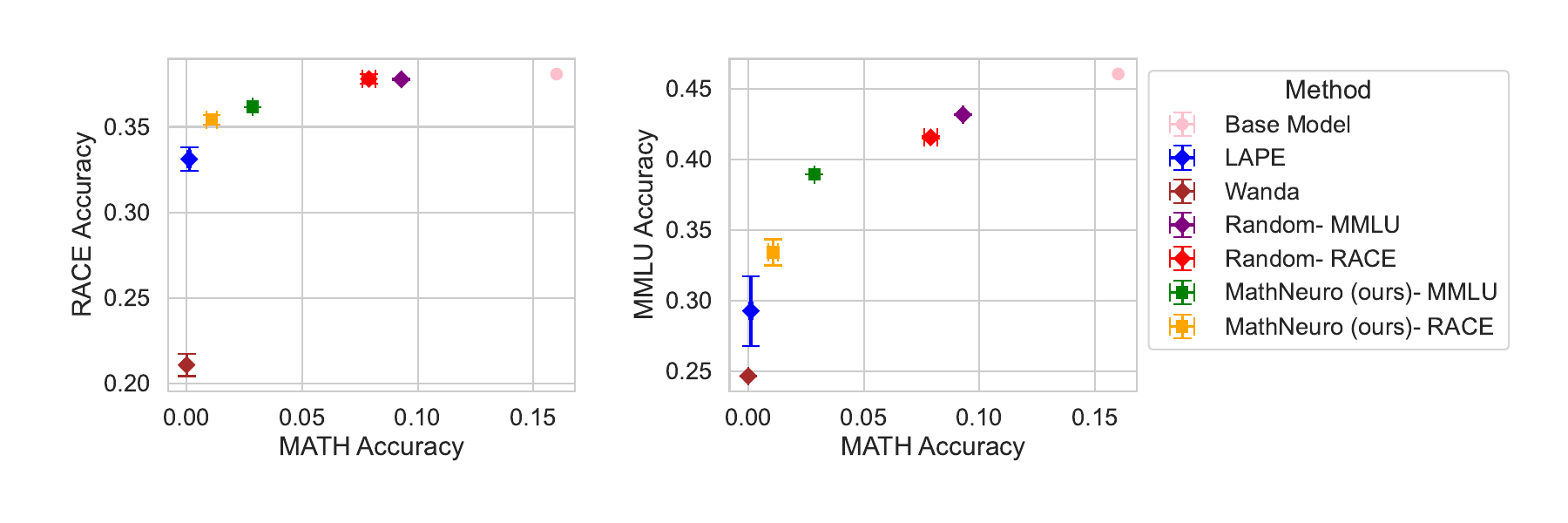}
    \caption{Effect of \textbf{\textit{pruning}} identified parameters on math and non-math performance for Llama 3.2 1B IT based on calculating the top 15\% of parameters using the MATH dataset as $\mathcal{D}_{\text{math}}$. Ideal methods should fall in the top left of the plot. MMLU and RACE denote that a point was calculated using MMLU or RACE, respectively, as $\mathcal{D}_{\text{non-math}}$. Horizontal and vertical lines represent 95\% confidence intervals for each point on the plot.}
    \label{fig:MathNeuro_llama1b_prune_MATH}
\end{figure*}

\begin{figure*}[t]
    \centering
    \includegraphics[width=1\linewidth]{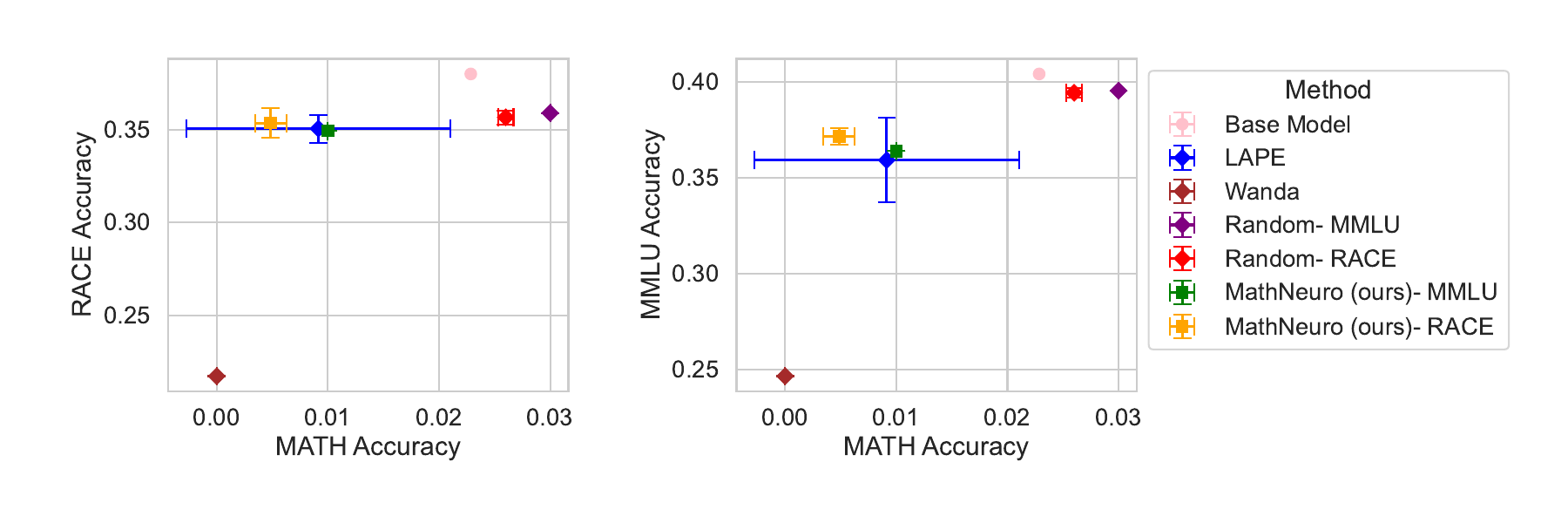}
    \caption{Effect of \textbf{\textit{pruning}} identified parameters on math and non-math performance for Phi 1.5 based on calculating the top 5\% of parameters using the MATH dataset as $\mathcal{D}_{\text{math}}$. Ideal methods should fall in the top left of the plot. MMLU and RACE denote that a point was calculated using MMLU or RACE, respectively, as $\mathcal{D}_{\text{non-math}}$. Horizontal and vertical lines represent 95\% confidence intervals for each point on the plot.}
    \label{fig:MathNeuro_phi15_prune_MATH}
\end{figure*}

\begin{figure*}[t]
    \centering
    \includegraphics[width=1\linewidth]{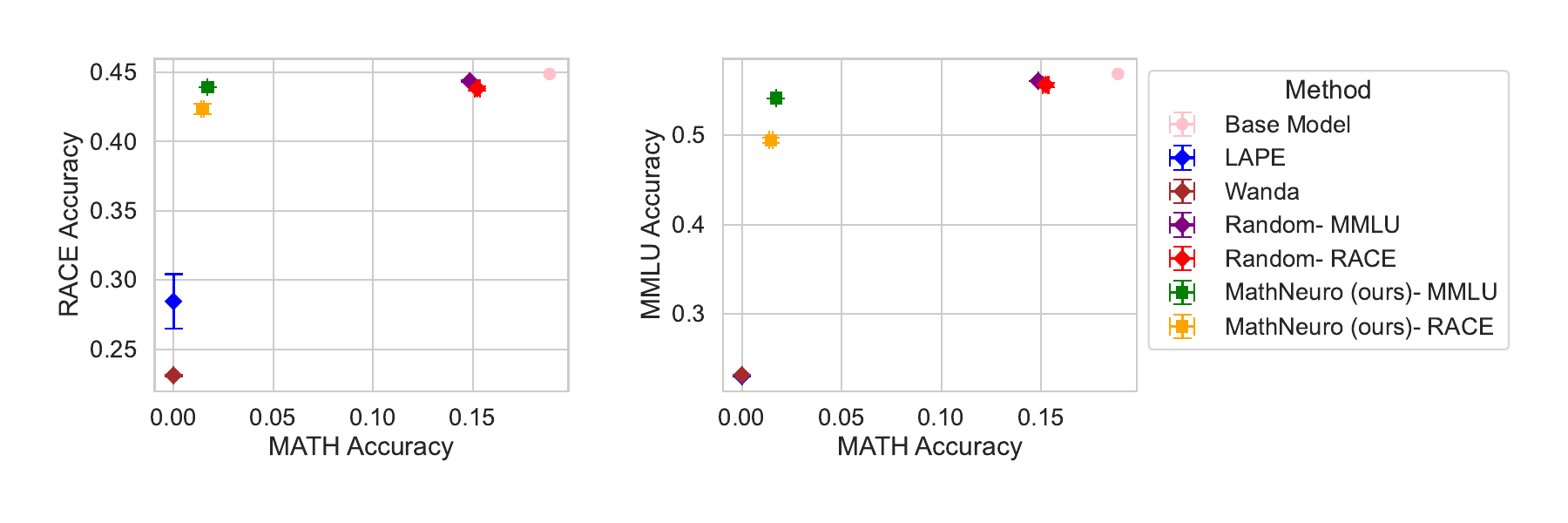}
    \caption{Effect of \textbf{\textit{pruning}} identified parameters on math and non-math performance for Gemma 2 2B IT based on calculating the top 5\% of parameters using the MATH dataset as $\mathcal{D}_{\text{math}}$.  Ideal methods should fall in the top left of the plot. MMLU and RACE denote that a point was calculated using MMLU or RACE, respectively, as $\mathcal{D}_{\text{non-math}}$. Horizontal and vertical lines represent 95\% confidence intervals for each point on the plot.}
    \label{fig:MathNeuro_gemma2b_prune_MATH}
\end{figure*}

\begin{figure*}[t]
    \centering
    \includegraphics[width=1\linewidth]{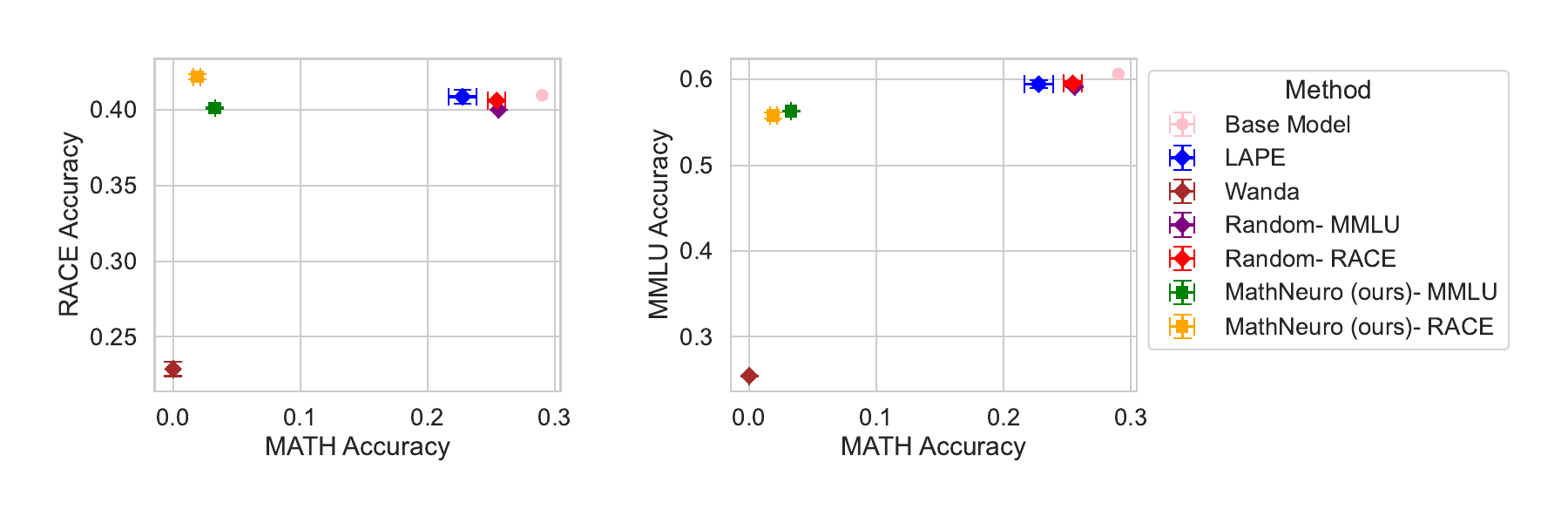}
    \caption{Effect of \textbf{\textit{pruning}} identified parameters on math and non-math performance for Llama 3.2 3B IT based on calculating the top 2.5\% (left) and 1\% (right) of parameters using the MATH dataset as $\mathcal{D}_{\text{math}}$.  Ideal methods should fall in the top left of the plot. MMLU and RACE denote that a point was calculated using MMLU or RACE, respectively, as $\mathcal{D}_{\text{non-math}}$. Horizontal and vertical lines represent 95\% confidence intervals for each point on the plot.}
    \label{fig:MathNeuro_llama3b_prune_MATH}
\end{figure*}

\begin{figure*}[t]
    \centering
    \includegraphics[width=1\linewidth]{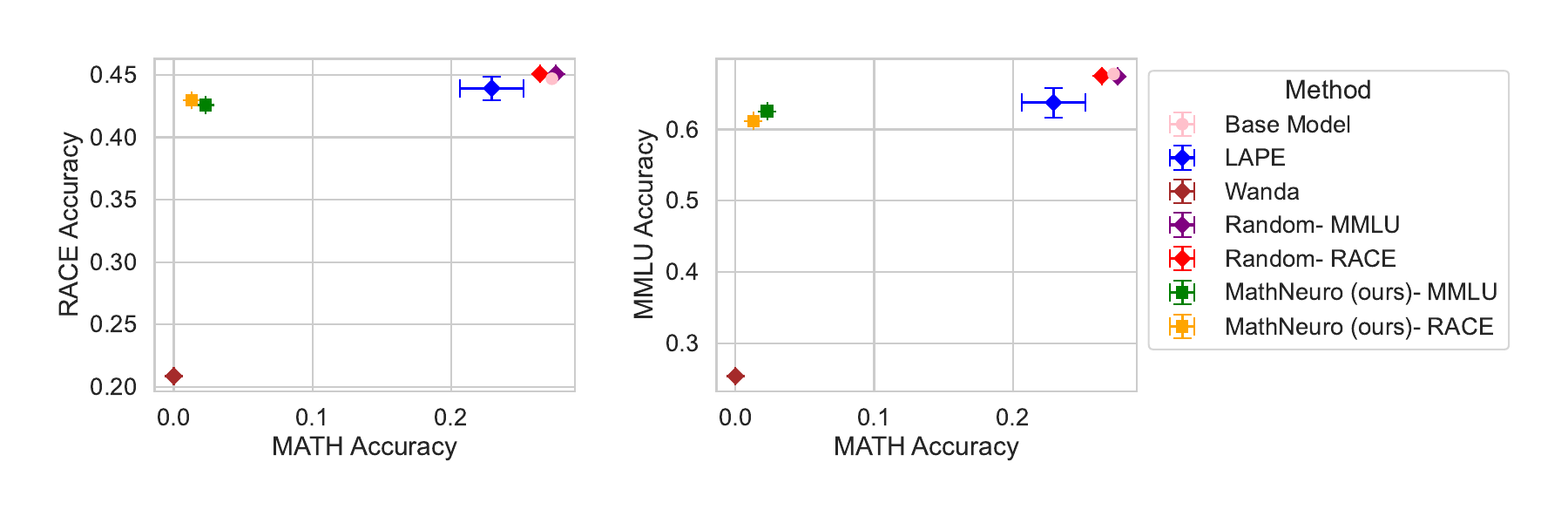}
    \caption{Effect of \textbf{\textit{pruning}} identified parameters on math and non-math performance for Llama 3.1 8B IT based on calculating the top 1\% of parameters using the MATH dataset as $\mathcal{D}_{\text{math}}$. Ideal methods should fall in the top left of the plot. MMLU and RACE denote that a point was calculated using MMLU or RACE, respectively, as $\mathcal{D}_{\text{non-math}}$. Horizontal and vertical lines represent 95\% confidence intervals for each point on the plot.}
    \label{fig:MathNeuro_llama8b_prune_MATH}
\end{figure*}

We replicate our GSM8K pruning experiments using MATH as the math dataset and MMLU or RACE as the non-math dataset. Similar to our approach with GSM8K, we subset the MATH testing split to the same 700 random samples for each model and experimental run for experimental efficiency. As with our other experiments, we use the Eleuther AI LM Evaluation Harness \cite{gao_framework_2024} for implementing the MATH evaluations, using their default implementation of the Minerva MATH prompt with 4-shot CoT examples. For each model, we use the exact same hyperparameters as those reported for the pruning experiments in Section \ref{deleting_params} and Appendix \ref{sec:pruning_figures}. As shown in Figures \ref{fig:MathNeuro_llama1b_prune_MATH}, \ref{fig:MathNeuro_phi15_prune_MATH}, \ref{fig:MathNeuro_gemma2b_prune_MATH}, \ref{fig:MathNeuro_llama3b_prune_MATH}, and \ref{fig:MathNeuro_llama8b_prune_MATH}, our GSM8K pruning results replicate when using MATH as the math dataset, as pruning math-specific parameters MathNeuro identifies effectively deletes MATH performance while leaving MMLU and RACE performance largely unaltered. These results buttress those reported in Section \ref{unseen_tasks_experiments} by showing that MathNeuro's effectiveness in isolating math-specific parameters holds when using a different math dataset. 

\begin{figure*}[t]
    \centering
    \includegraphics[width=1\linewidth]{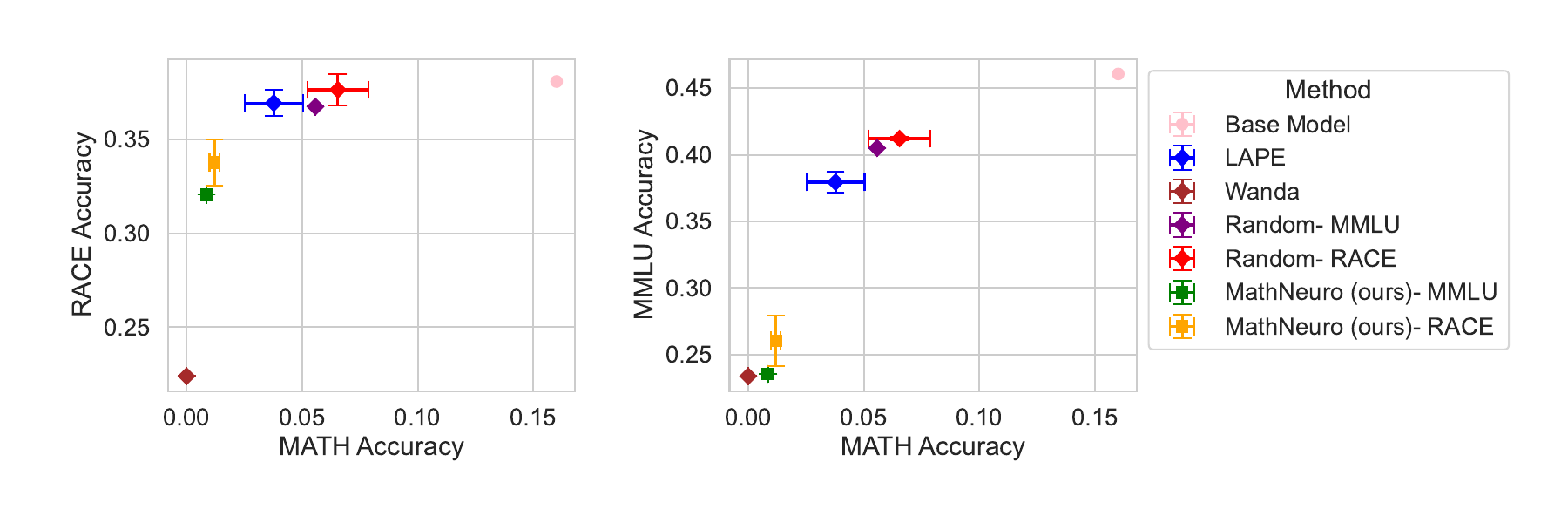}
    \caption{Effect of \textbf{\textit{pruning}} identified parameters on math and non-math performance for Llama 3.2 1B IT based on calculating the top 15\% of parameters using the MATH dataset as $\mathcal{D}_{\text{math}}$ \textbf{based on one sample}. Ideal methods should fall in the top left of the plot. MMLU and RACE denote that a point was calculated using MMLU or RACE, respectively, as $\mathcal{D}_{\text{non-math}}$. Horizontal and vertical lines represent 95\% confidence intervals for each point on the plot.}
    \label{fig:MathNeuro_llama1b_prune_MATH_one_sample}
\end{figure*}

\begin{figure*}[t]
    \centering
    \includegraphics[width=1\linewidth]{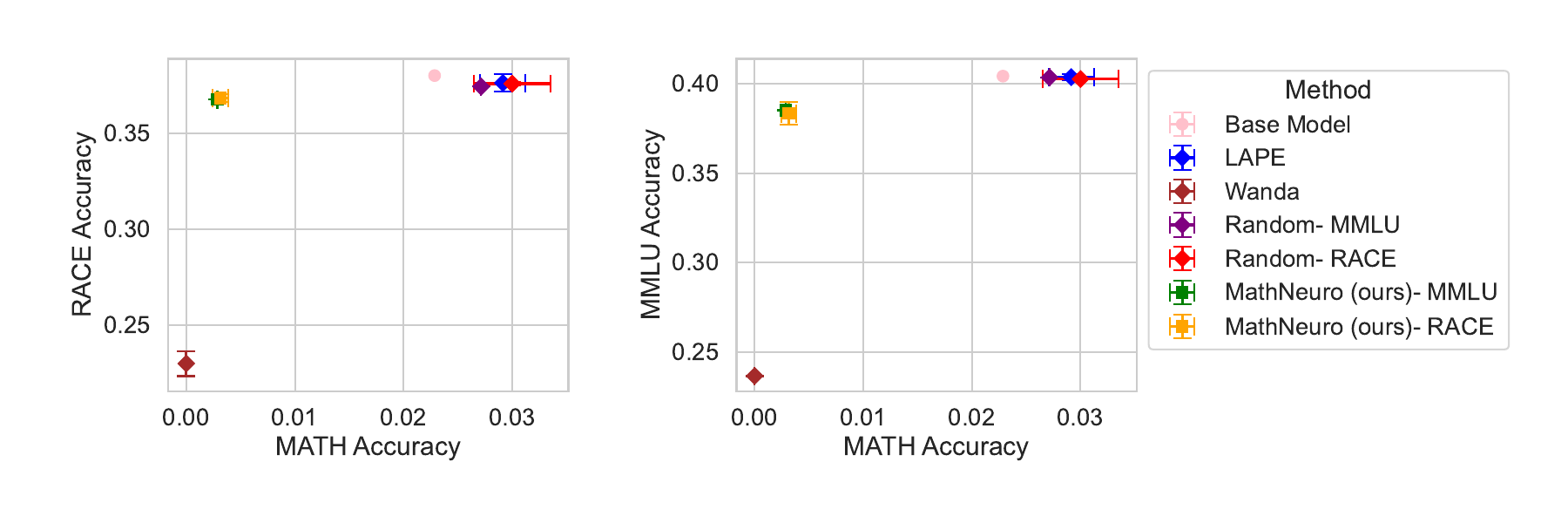}
    \caption{Effect of \textbf{\textit{pruning}} identified parameters on math and non-math performance for Phi 1.5 based on calculating the top 5\% of parameters using the MATH dataset as $\mathcal{D}_{\text{math}}$ \textbf{based on one sample}. Ideal methods should fall in the top left of the plot. MMLU and RACE denote that a point was calculated using MMLU or RACE, respectively, as $\mathcal{D}_{\text{non-math}}$. Horizontal and vertical lines represent 95\% confidence intervals for each point on the plot.}
    \label{fig:MathNeuro_phi15_prune_MATH_one_sample}
\end{figure*}

\begin{figure*}[t]
    \centering
    \includegraphics[width=1\linewidth]{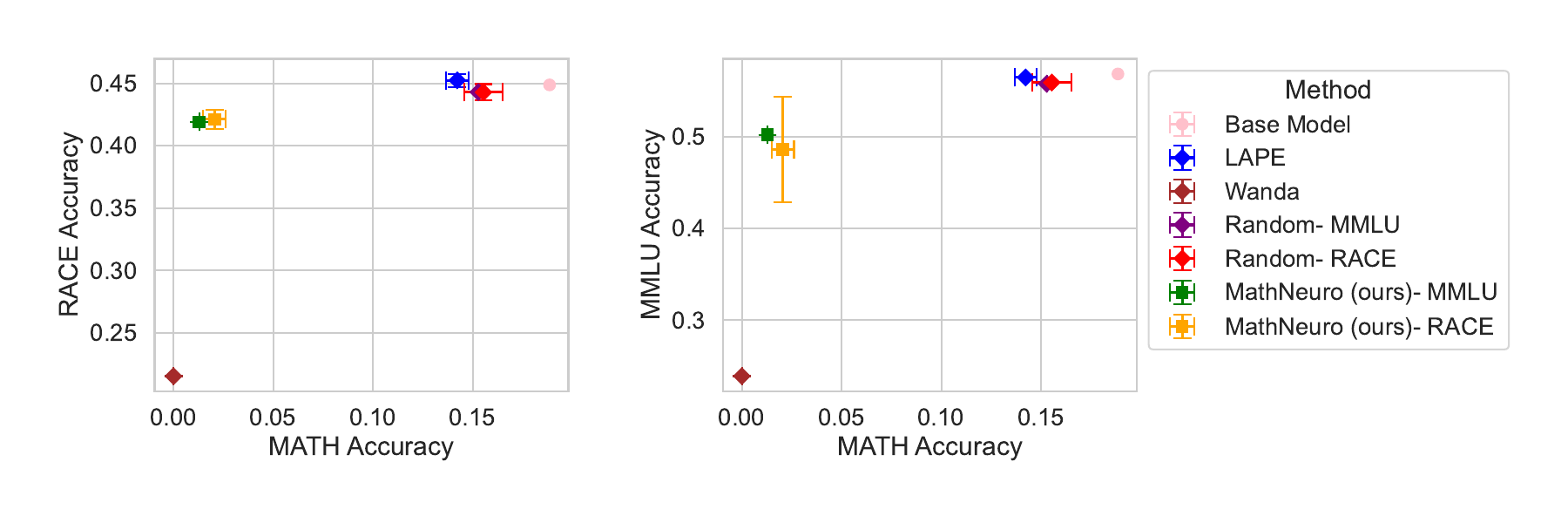}
    \caption{Effect of \textbf{\textit{pruning}} identified parameters on math and non-math performance for Gemma 2 2B IT based on calculating the top 5\% of parameters using the MATH dataset as $\mathcal{D}_{\text{math}}$ \textbf{based on one sample}.  Ideal methods should fall in the top left of the plot. MMLU and RACE denote that a point was calculated using MMLU or RACE, respectively, as $\mathcal{D}_{\text{non-math}}$. Horizontal and vertical lines represent 95\% confidence intervals for each point on the plot.}
    \label{fig:MathNeuro_gemma2b_prune_MATH_one_sample}
\end{figure*}

\begin{figure*}[t]
    \centering
    \includegraphics[width=1\linewidth]{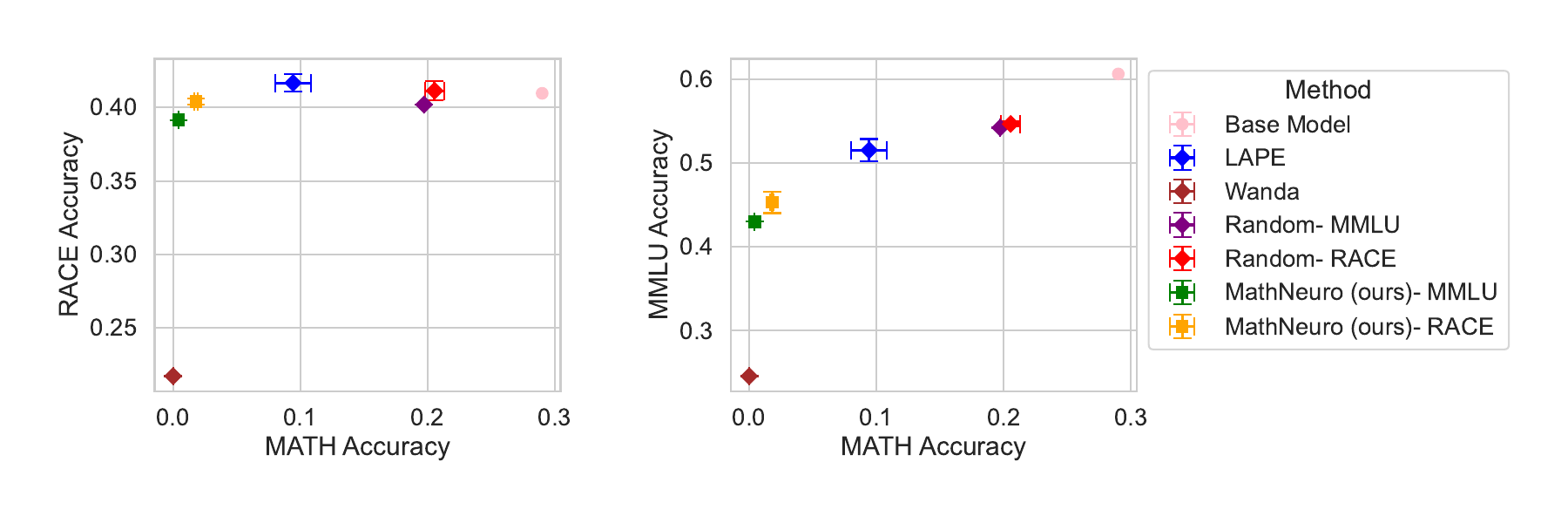}
    \caption{Effect of \textbf{\textit{pruning}} identified parameters on math and non-math performance for Llama 3.2 3B IT based on calculating the top 2.5\% (left) and 1\% (right) of parameters using the MATH dataset as $\mathcal{D}_{\text{math}}$ \textbf{based on one sample}.  Ideal methods should fall in the top left of the plot. MMLU and RACE denote that a point was calculated using MMLU or RACE, respectively, as $\mathcal{D}_{\text{non-math}}$. Horizontal and vertical lines represent 95\% confidence intervals for each point on the plot.}
    \label{fig:MathNeuro_llama3b_prune_MATH_one_sample}
\end{figure*}

\begin{figure*}[t]
    \centering
    \includegraphics[width=1\linewidth]{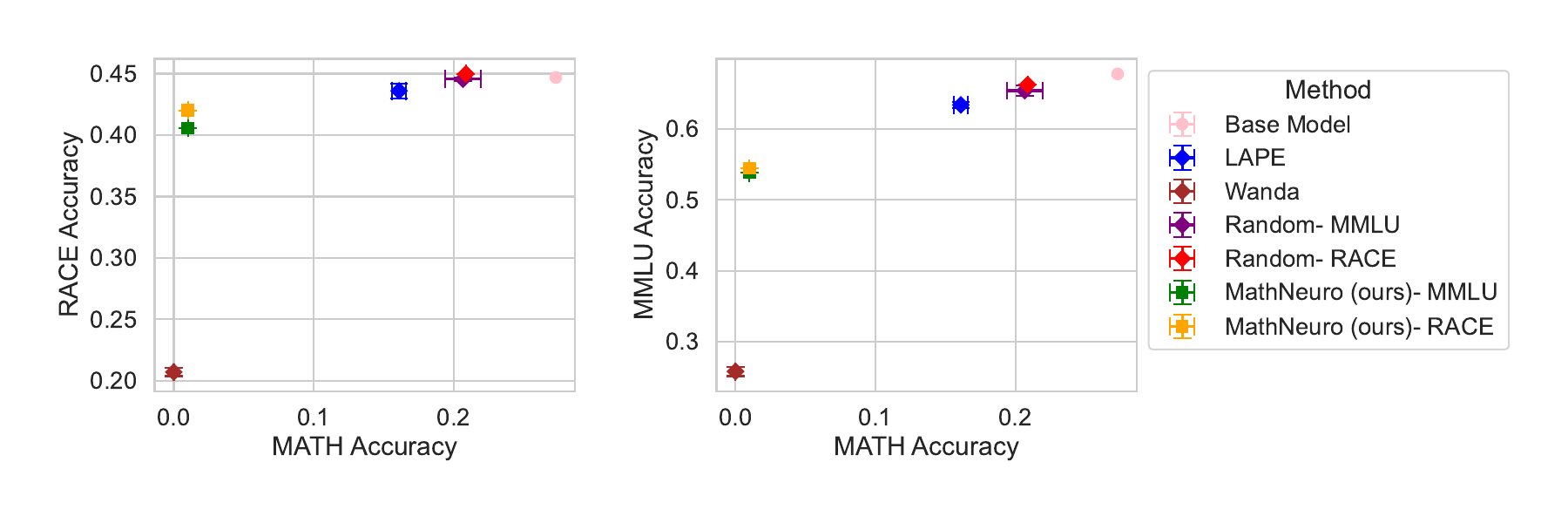}
    \caption{Effect of \textbf{\textit{pruning}} identified parameters on math and non-math performance for Llama 3.1 8B IT based on calculating the top 1\% of parameters using the MATH dataset as $\mathcal{D}_{\text{math}}$ \textbf{based on one sample}. Ideal methods should fall in the top left of the plot. MMLU and RACE denote that a point was calculated using MMLU or RACE, respectively, as $\mathcal{D}_{\text{non-math}}$. Horizontal and vertical lines represent 95\% confidence intervals for each point on the plot.}
    \label{fig:MathNeuro_llama8b_prune_MATH_one_sample}
\end{figure*}

Using the same hyperparameters, we also replicate our one sample pruning experiments using MATH as $\mathcal{D}_{\text{math}}$. Results are shown in Figures \ref{fig:MathNeuro_llama1b_prune_MATH_one_sample}, \ref{fig:MathNeuro_phi15_prune_MATH_one_sample}, \ref{fig:MathNeuro_gemma2b_prune_MATH_one_sample}, \ref{fig:MathNeuro_llama3b_prune_MATH_one_sample}, and \ref{fig:MathNeuro_llama8b_prune_MATH_one_sample}, where MathNeuro still performs best at isolating math-specific parameters when using a single MATH sample for parameter identification.

\subsection{MATH Scaling Experiments}
\begin{figure*}[t]
    \centering
    \includegraphics[width=1\linewidth]{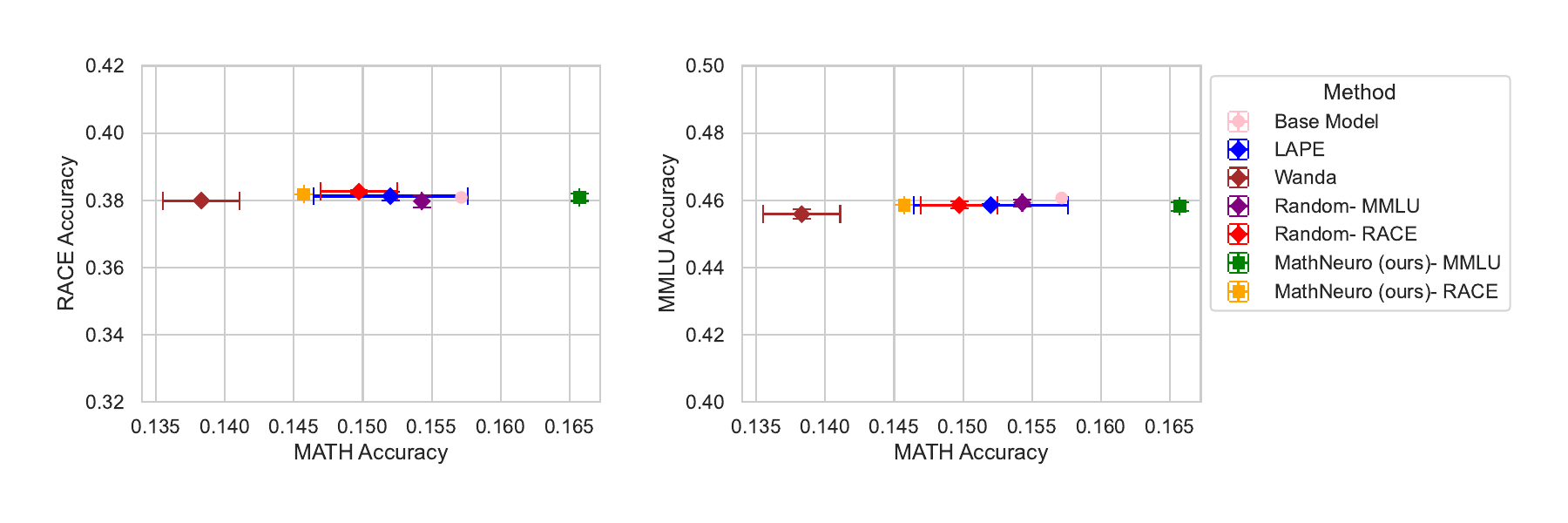}
    \caption{Effect of \textbf{\textit{scaling}} identified parameters by 1.025 on math and non-math performance for Llama 3.2 1B IT based on calculating the top .5\% of parameters using the MATH dataset as $\mathcal{D}_{\text{math}}$. Ideal methods should fall in the top right of the plot. MMLU and RACE denote that a point was calculated using MMLU or RACE, respectively,  as $\mathcal{D}_{\text{non-math}}$. Horizontal and vertical lines represent 95\% confidence intervals for each point on the plot.}
    \label{fig:MathNeuro_llama1b_scale_MATH}
\end{figure*}

\begin{figure*}[t]
    \centering
    \includegraphics[width=1\linewidth]{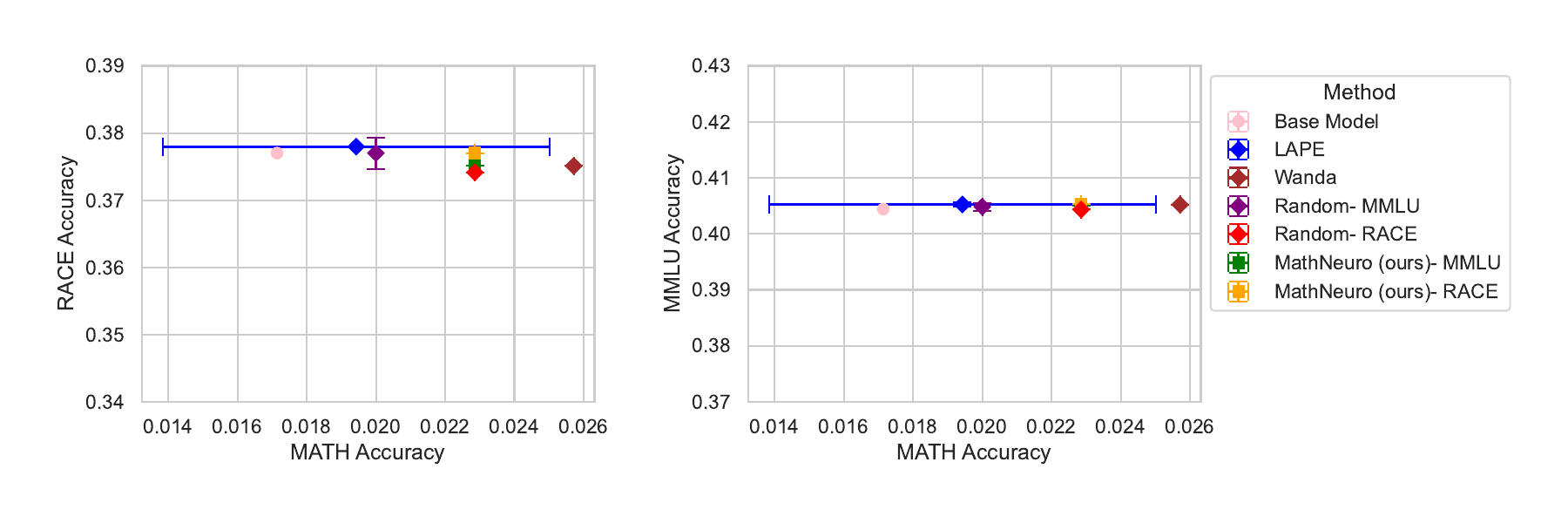}
    \caption{Effect of \textbf{\textit{scaling}} identified parameters by 1.025 on math and non-math performance for Phi 1.5 based on calculating the top 15\% of parameters using the MATH dataset as $\mathcal{D}_{\text{math}}$. Ideal methods should fall in the top right of the plot. MMLU and RACE denote that a point was calculated using MMLU or RACE, respectively, as $\mathcal{D}_{\text{non-math}}$. Horizontal and vertical lines represent 95\% confidence intervals for each point on the plot.}
    \label{fig:MathNeuro_phi15_scale_MATH}
\end{figure*}

\begin{figure*}[t]
    \centering
    \includegraphics[width=1\linewidth]{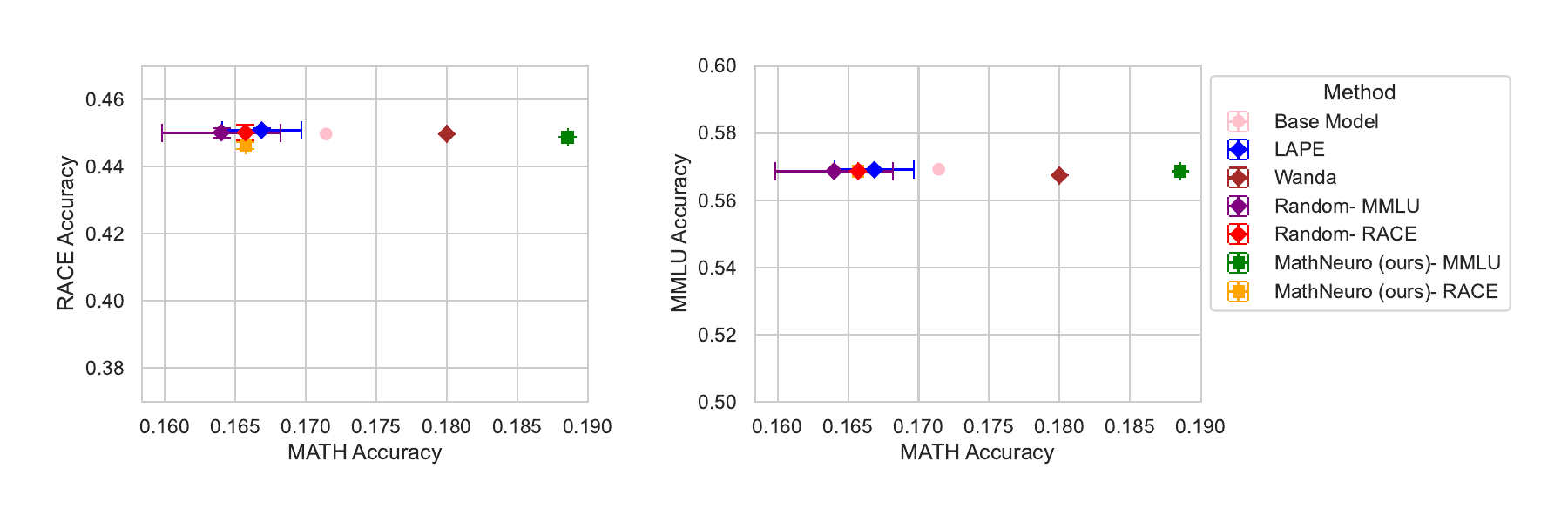}
    \caption{Effect of \textbf{\textit{scaling}} identified parameters by 1.05 on math and non-math performance for Gemma 2 2B IT based on calculating the top .5\% of parameters using the MATH dataset as $\mathcal{D}_{\text{math}}$. Ideal methods should fall in the top right of the plot. MMLU and RACE denote that a point was calculated using MMLU or RACE, respectively, as $\mathcal{D}_{\text{non-math}}$. Horizontal and vertical lines represent 95\% confidence intervals for each point on the plot.}
    \label{fig:MathNeuro_gemma2b_scale_MATH}
\end{figure*}

\begin{figure*}[t]
    \centering
    \includegraphics[width=1\linewidth]{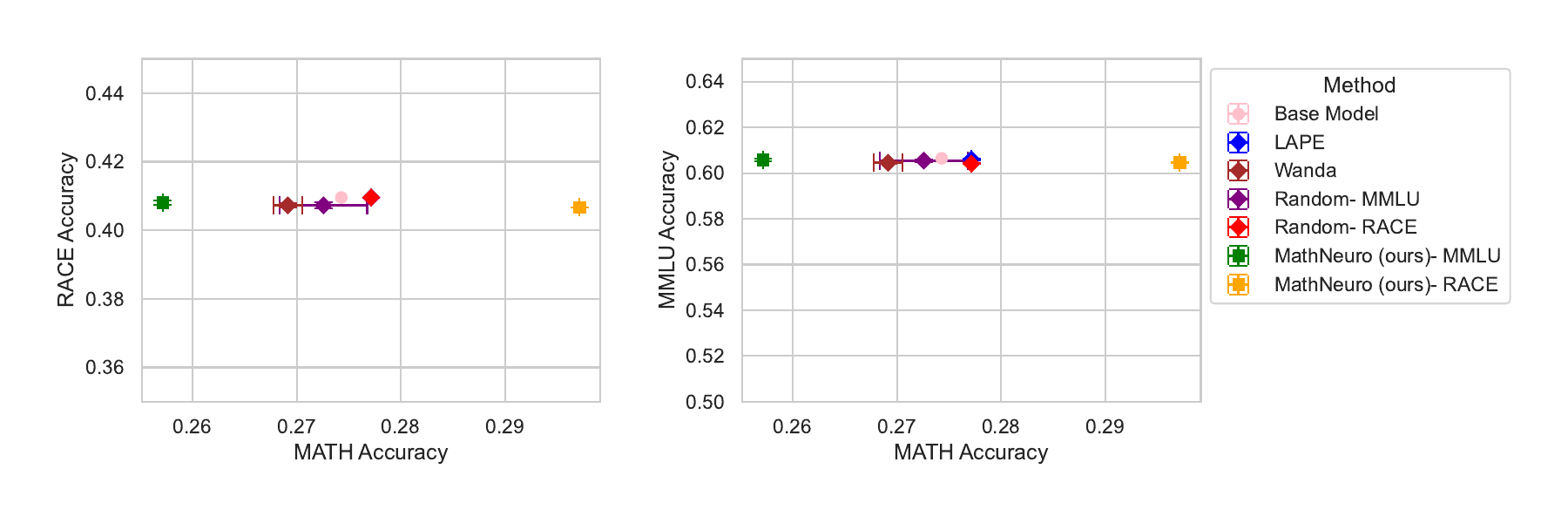}
    \caption{Effect of \textbf{\textit{scaling}} identified parameters by 1.025 on math and non-math performance for Llama 3.2 3B IT based on calculating the top 10\% of parameters using the MATH dataset as $\mathcal{D}_{\text{math}}$. Ideal methods should fall in the top right of the plot. MMLU and RACE denote that a point was calculated using MMLU or RACE, respectively, as $\mathcal{D}_{\text{non-math}}$. Horizontal and vertical lines represent 95\% confidence intervals for each point on the plot.}
    \label{fig:MathNeuro_llama3b_scale_MATH}
\end{figure*}

\begin{figure*}[t]
    \centering
    \includegraphics[width=1\linewidth]{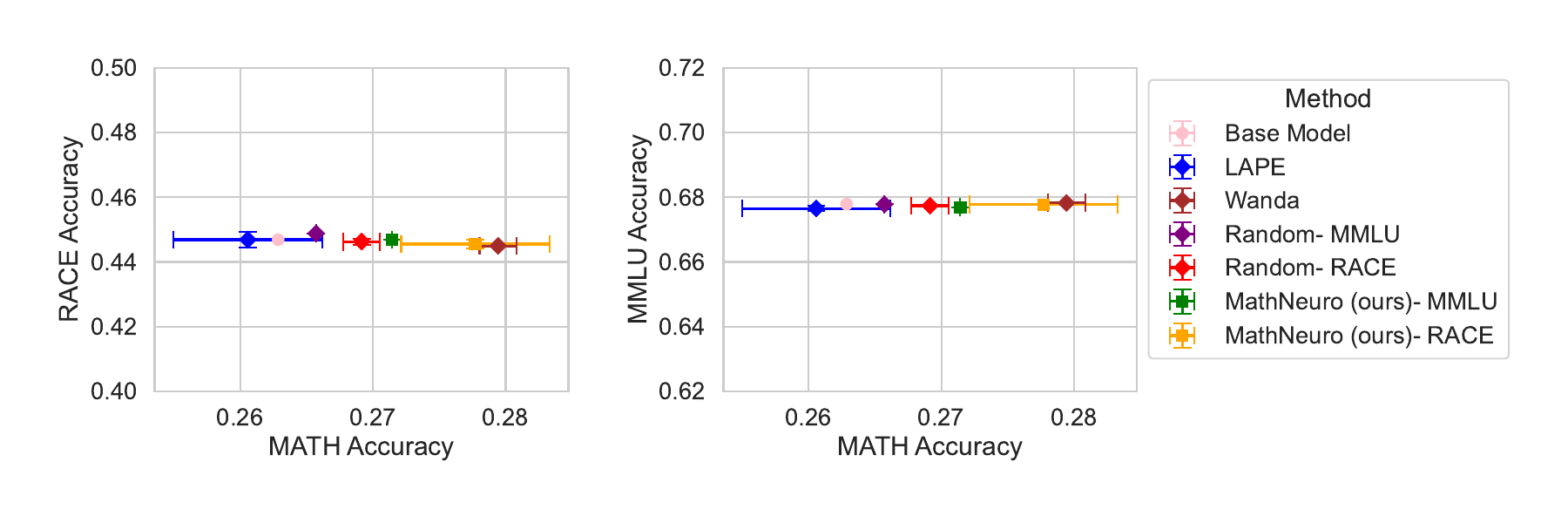}
    \caption{Effect of \textbf{\textit{scaling}} identified parameters by 1.01 on math and non-math performance for Llama 3.1 8B IT based on calculating the top 2.5\% of parameters using the MATH dataset as $\mathcal{D}_{\text{math}}$. Ideal methods should fall in the top right of the plot. MMLU and RACE denote that a point was calculated using MMLU or RACE, respectively, as $\mathcal{D}_{\text{non-math}}$. Horizontal and vertical lines represent 95\% confidence intervals for each point on the plot.}
    \label{fig:MathNeuro_llama8b_scale_MATH}
\end{figure*}

\begin{figure*}[t]
    \centering
    \includegraphics[width=1\linewidth]{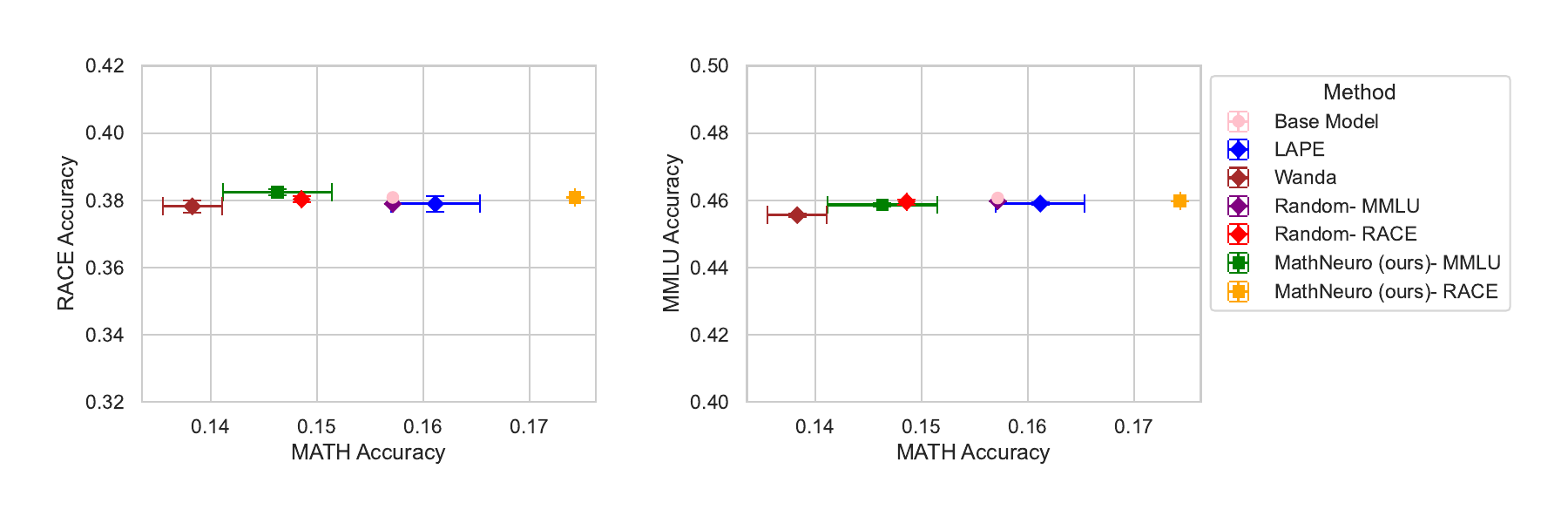}
    \caption{Effect of \textbf{\textit{scaling}} identified parameters by 1.025 on math and non-math performance for Llama 3.2 1B IT based on calculating the top .5\% of parameters using the MATH dataset as $\mathcal{D}_{\text{math}}$ \textbf{based on one sample}. Ideal methods should fall in the top right of the plot. MMLU and RACE denote that a point was calculated using MMLU or RACE, respectively,  as $\mathcal{D}_{\text{non-math}}$. Horizontal and vertical lines represent 95\% confidence intervals for each point on the plot.}
    \label{fig:MathNeuro_llama1b_scale_MATH_one_sample}
\end{figure*}

\begin{figure*}[t]
    \centering
    \includegraphics[width=1\linewidth]{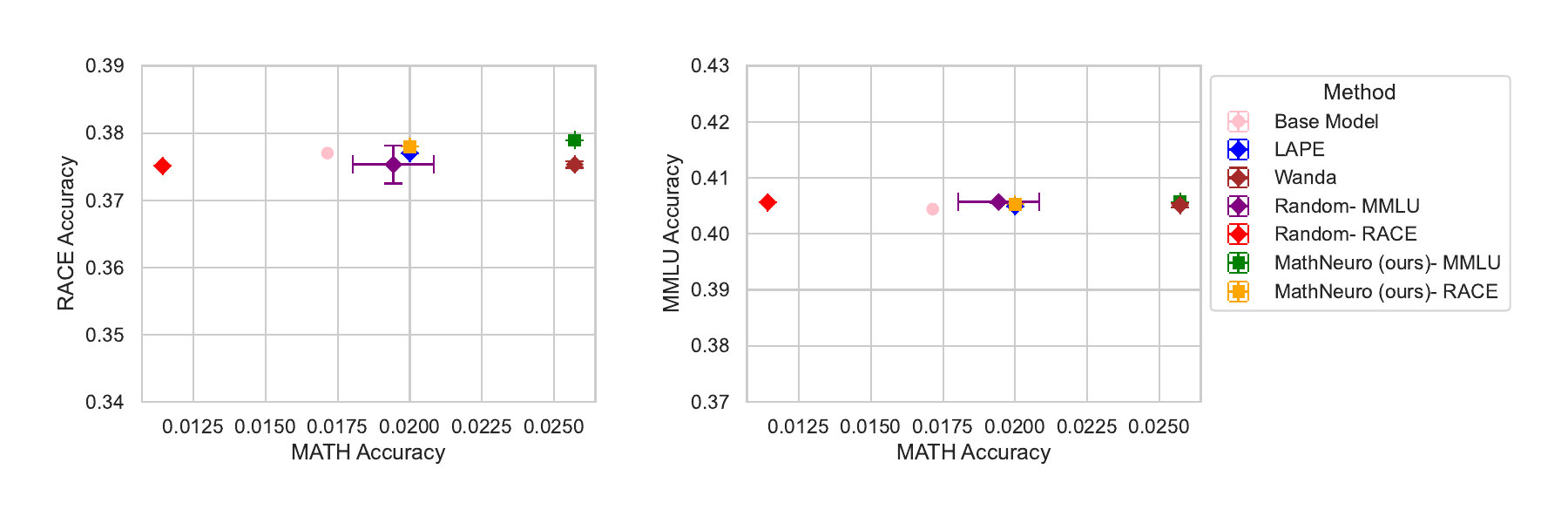}
    \caption{Effect of \textbf{\textit{scaling}} identified parameters by 1.025 on math and non-math performance for Phi 1.5 based on calculating the top 10\% of parameters using the MATH dataset as $\mathcal{D}_{\text{math}}$ \textbf{based on one sample}. Ideal methods should fall in the top right of the plot. MMLU and RACE denote that a point was calculated using MMLU or RACE, respectively, as $\mathcal{D}_{\text{non-math}}$. Horizontal and vertical lines represent 95\% confidence intervals for each point on the plot.}
    \label{fig:MathNeuro_phi15_scale_MATH_one_sample}
\end{figure*}

\begin{figure*}[t]
    \centering
    \includegraphics[width=1\linewidth]{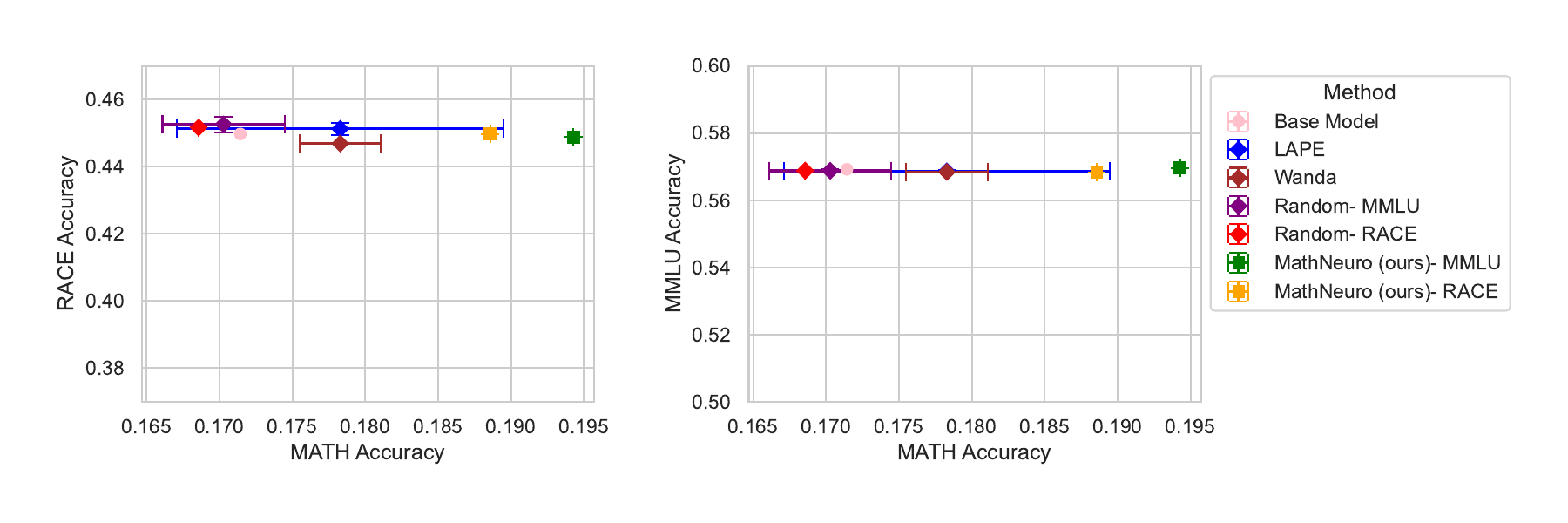}
    \caption{Effect of \textbf{\textit{scaling}} identified parameters by 1.05 on math and non-math performance for Gemma 2 2B IT based on calculating the top 1\% of parameters using the MATH dataset as $\mathcal{D}_{\text{math}}$ \textbf{based on one sample}. Ideal methods should fall in the top right of the plot. MMLU and RACE denote that a point was calculated using MMLU or RACE, respectively, as $\mathcal{D}_{\text{non-math}}$. Horizontal and vertical lines represent 95\% confidence intervals for each point on the plot.}
    \label{fig:MathNeuro_gemma2b_scale_MATH_one_sample}
\end{figure*}

\begin{figure*}[t]
    \centering
    \includegraphics[width=1\linewidth]{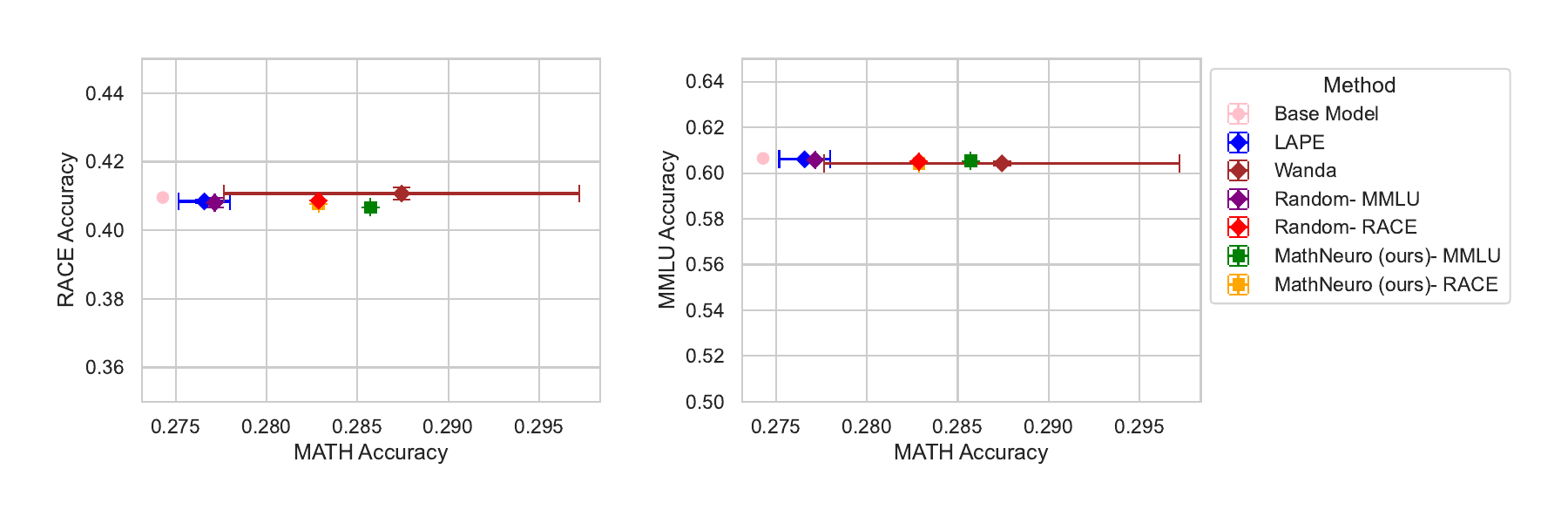}
    \caption{Effect of \textbf{\textit{scaling}} identified parameters by 1.025 on math and non-math performance for Llama 3.2 3B IT based on calculating the top .01\% of parameters using the MATH dataset as $\mathcal{D}_{\text{math}}$ \textbf{based on one sample}. Ideal methods should fall in the top right of the plot. MMLU and RACE denote that a point was calculated using MMLU or RACE, respectively, as $\mathcal{D}_{\text{non-math}}$. Horizontal and vertical lines represent 95\% confidence intervals for each point on the plot.}
    \label{fig:MathNeuro_llama3b_scale_MATH_one_sample}
\end{figure*}

\begin{figure*}[t]
    \centering
    \includegraphics[width=1\linewidth]{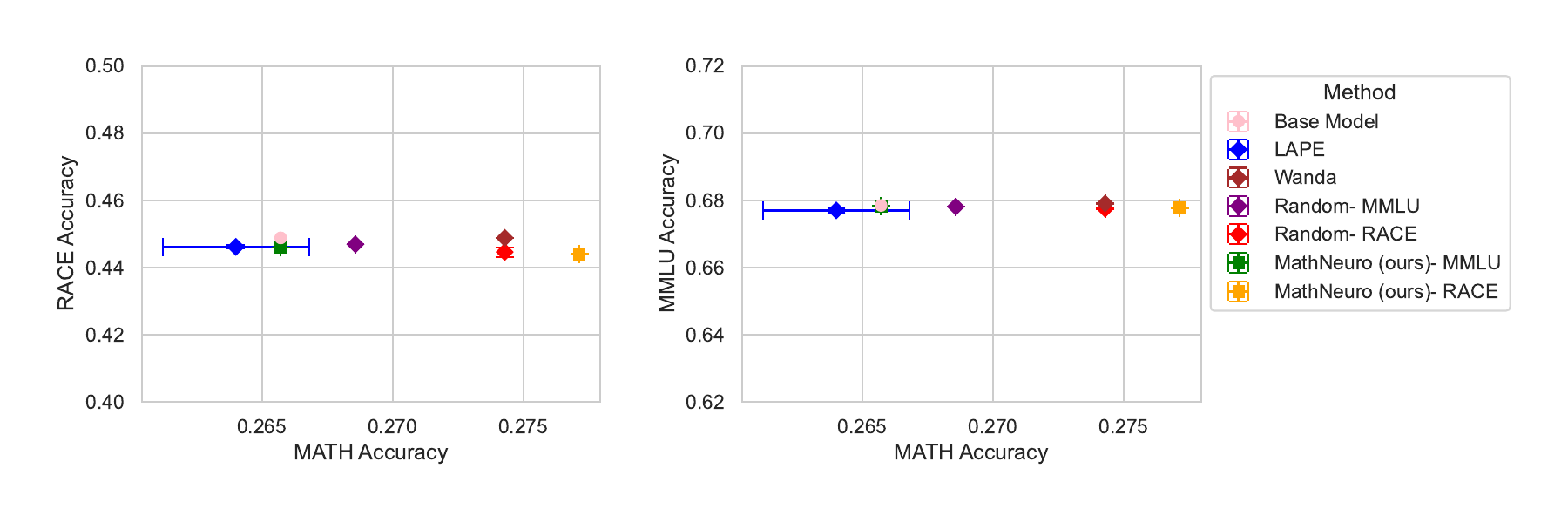}
    \caption{Effect of \textbf{\textit{scaling}} identified parameters by 1.01 on math and non-math performance for Llama 3.1 8B IT based on calculating the top 10\% of parameters using the MATH dataset as $\mathcal{D}_{\text{math}}$ \textbf{based on one sample}. Ideal methods should fall in the top right of the plot. MMLU and RACE denote that a point was calculated using MMLU or RACE, respectively, as $\mathcal{D}_{\text{non-math}}$. Horizontal and vertical lines represent 95\% confidence intervals for each point on the plot.}
    \label{fig:MathNeuro_llama8b_scale_MATH_one_sample}
\end{figure*}

We also replicate our GSM8K scaling experiments using MATH as the math dataset and MMLU or RACE as the non-math dataset. Similar to our MATH pruning experiments, we subset the MATH testing split to the same random samples for each model and experimental run for experimental efficiency and implement the MATH task with the same prompting approach described above. We use a smaller sample of the MATH testing split (350 samples) for our scaling experiments given that we conduct the rough grid search for an optimal scaling factor and parameter proportion described in Appendix \ref{sec:hyperparameters} and Section \ref{scaling_params}, respectively, and using a larger set of samples would be computationally prohibitive. As shown in Figures \ref{fig:MathNeuro_llama1b_scale_MATH}, \ref{fig:MathNeuro_phi15_scale_MATH}, \ref{fig:MathNeuro_gemma2b_scale_MATH}, \ref{fig:MathNeuro_llama3b_scale_MATH}, and \ref{fig:MathNeuro_llama8b_scale_MATH}, our GSM8K scaling results replicate when using MATH as the math dataset, as scaling math-specific parameters MathNeuro identifies boosts MATH performance by a small but meaningful amount while leaving MMLU and RACE performance unchanged. These increases in performance correspond to increasing baseline MATH performance by 5-35\% depending on the model. These results show that scaling parameters identified using MathNeuro boosts math performance even when using different math datasets. 
\newline \indent Using the same scaling factors and conducting a small grid search for the optimal parameter proportion like we did in Section \ref{single_sample}, we also replicate our one sample scaling experiments using MATH as $\mathcal{D}_{\text{math}}$.  Results are shown in Figures \ref{fig:MathNeuro_llama1b_scale_MATH_one_sample}, \ref{fig:MathNeuro_phi15_scale_MATH_one_sample}, \ref{fig:MathNeuro_gemma2b_scale_MATH_one_sample}, \ref{fig:MathNeuro_llama3b_scale_MATH_one_sample}, and \ref{fig:MathNeuro_llama8b_scale_MATH_one_sample}, where scaling MathNeuro-identified parameters using a single MATH sample for parameter identification still results in a small but meaningful boost in MATH performance while leaving non-math performance unaltered. 




%% file: sections/sample_outputs.tex
Tables \ref{tab:gsm8k_outputs}, \ref{tab:gsm8k_outputs_scaling}, \ref{tab:race_outputs}, \ref{tab:mmlu_outputs}, and \ref{tab:mmlu_outputs_scaling} display sample outputs from Llama 3.2 1B IT before and after pruning or scaling parameters identified by MathNeuro. The tables display outputs for a GSM8K, RACE, or MMLU question. 
\begin{table}[t]
    \centering\resizebox{\linewidth}{!}{
    \begin{tabular}{p{1\linewidth}} \toprule
        \textbf{Question} \\
        Betty is saving money for a new wallet which costs \$100. Betty has only half of the money she needs. Her parents decided to give her \$15 for that purpose, and her grandparents twice as much as her parents. How much more money does Betty need to buy the wallet?\\ \midrule 
        \textbf{Response Before Pruning (correct)}\\ 
        Let's think step by step. Betty has half of the money she needs, which is $100 / 2 = $50. She received \$15 from her parents, so now she has \$50 + \$15 = \$65. Her grandparents gave her twice as much as her parents, which is 2 x \$15 = \$30. Now she has \$65 + \$30 = \$95. The wallet costs \$100, so Betty needs \$100 - \$95 = \$5 more to buy the wallet.
        \\ \midrule \textbf{Response After Pruning Based on RACE as $\mathcal{D}_{\text{non-math}}$ (incorrect)}\\ Let's think step by step. First, let's determine the total amount needed to buy the wallet. \$100 is the total amount needed to buy the wallet. Betty's parents gave her \$15. Her grandparents gave her \$2 as a gift. Betty's grandparents gave her \$15 as a gift. Betty's parents gave her \$15 as a gift. Betty's grandparents gave her \$15 as a gift. Betty's parents gave her \$15 as a gift. Betty's grandparents gave her \$15 as a gift. Betty's grandparents gave her \$15 as a gift. Betty's grandparents gave her \$15 as a gift. Betty's grandparents gave her \$15 as a gift. Betty's grandparents gave her \$15 as a gift. Betty's grandparents gave her \$15
        \\ \midrule \textbf{Response After Pruning Based on MMLU as $\mathcal{D}_{\text{non-math}}$ (incorrect)}\\ A: Let's think step by step. Betty needs \$100 to buy the wallet. Her parents gave her \$15 for that purpose, and her grandparents gave her \$15 twice as much as her parents. This is a 50-50 split. Betty's grandparents gave her \$15 twice as much as her parents, which is \$15. Therefore, Betty's grandparents gave her \$15 twice as much as her parents.
        B: Let's think step by step. Betty's grandparents gave her \$15 twice as much as her parents. This is a 50-50 split. Betty's grandparents gave her \$15 twice as much as her parents. Therefore, Betty's grandparents gave her \$15 twice as much as her parents.
        C: Let's think step by step. Betty's
        \\ \bottomrule 
    \end{tabular}}
    \caption{Responses to a GSM8K question before and after pruning parameters identified by MathNeuro for Llama 3 1B IT.}
    \label{tab:gsm8k_outputs}
\end{table}

\begin{table}[t]
    \centering\resizebox{\linewidth}{!}{
    \begin{tabular}{p{1\linewidth}} \toprule
        \textbf{Question} \\
        The profit from a business transaction is shared among 2 business partners, Mike and Johnson in the ratio 2:5 respectively. If Johnson got \$2500, how much will Mike have after spending some of his share on a shirt that costs \$200?\\ \midrule 
        \textbf{Response Before Scaling (incorrect)}\\ 
        Let's think step by step. Mike's share of the profit is 2 parts, and Johnson's share is 5 parts. The total number of parts is 2 + 5 = 7. Johnson's share of the profit is 5/7 of the total profit. If Johnson got \$2500, then Mike's share of the profit is 2/7 of the total profit. If Mike spent \$200 on a shirt, then he will have \$200 + (2/7) of the profit. The profit is 7/7 of the total profit, so Mike's share of the profit is 2/7 of 7/7, which is 2/7. Mike's share of the profit is \$200 + (2/7) of the profit, which is \$200 + (2/7)(7/7) = \$200 + 2 = \$202. The answer is \$202.\\ \midrule \textbf{Response After Scaling Based on RACE as $\mathcal{D}_{\text{non-math}}$ (incorrect)}\\ Let's think step by step. Mike's share of the profit is 2 parts, and Johnson's share is 5 parts. The total number of parts is 2 + 5 = 7. Johnson's share is 5/7 of the total profit. Therefore, Mike's share is 2/7 of the total profit. Mike's share is 2/7 × \$2500 = \$142.86. Mike will have \$142.86 after spending \$200 on a shirt.\\ \midrule \textbf{Response After Scaling Based on MMLU as $\mathcal{D}_{\text{non-math}}$ (correct)}\\ Let's think step by step. Mike's share of the profit is 2 parts, and Johnson's share is 5 parts. The total number of parts is 2 + 5 = 7. Johnson's share is 5/7 of the total profit. If Johnson got \$2500, then Mike's share is 2/7 of the total profit. If Mike spent \$200 on a shirt, how much profit will he have left? To find Mike's profit, we need to find the total profit first. We can set up the following proportion: 2/7 = 2500/\$x, where x is the total profit. Cross-multiplying, we get 2x = 2500(7). Dividing both sides by 2, we get x = 3500. Now that we know the total profit, we can find Mike's profit: Mike's share is 2/7 of the total profit, so Mike's profit is 2/7(3500) = 1000. Finally, we can find out how much profit Mike will have left after spending \$200 on a shirt: Mike's profit - \$200 = \$800. The answer is \$800.\\ \bottomrule 
    \end{tabular}}
    \caption{Responses to a GSM8K question before and after scaling parameters identified by MathNeuro for Llama 3 1B IT.}
    \label{tab:gsm8k_outputs_scaling}
\end{table}

\begin{table*}[t]
    \centering\resizebox{\linewidth}{!}{
    \begin{tabular}{p{1\linewidth}} \toprule
        \textbf{Prompt} \\
        Read this passage and answer the multiple choice question below it. 
    
        A newspaper reporter's job can be very interesting. He meets all types of people and lives quite a busy life. He is for news all the time,then after several years he may get a desk job,and life becomes a bit more settled. Let's look at his work a little more closely. In a day he may have to interview the prime minister of a foreign country, and the next day he may be writing about a football match. Sometimes he may be so busy that he hardly has any time to sleep. And at other times he may go on for days looking out for news materials yet return empty--handed.
        \newline In the beginning, a reporter has to cover a very wide field. After the early years he becomes more specialized in his work. For example, he may finaliy be asked to write only on court cases or politics or sports. Some reporters may become so specialized that they are asked only to write on a special thing: horse racing, for example. In most newspaper houses there is at least one special racing correspondent. Some newspapers have book reviews. Their job is delightful. They read the latest book and then write reviews on the ones they like. Then there are those who write on fdms, so they get to see them even before they are shown in the cinema. How lucky, you would say!
        A reporter's job can also be very dangerous. If there is a flood or a riot they may get hurt or even be killed. Three years ago there was a reporter whose camera was destroyed by a group of men, because they were angry with him for taking their picture. Dangerous or not, one thing is certain, and that is, their job is never dull!
        \newline
        Question: Reporters who write on films are said to be lucky because they  [blank].
        \newline
        Answer choices: ['can write anything they like', 'can see more film stars', 'can pay less than other people', 'can see the fdms before most people see them in the cinema']
        \\ \midrule 
        \textbf{Response Before Interventions (incorrect)}\\ 
         Answer: 'can see more film stars'
        \\ \midrule \textbf{Response After Pruning Based on RACE as $\mathcal{D}_{\text{non-math}}$ (incorrect)}\\ Answer: 'can see more film stars' 
        \\ \midrule \textbf{Response After Pruning Based on MMLU as $\mathcal{D}_{\text{non-math}}$ (incorrect)}\\ Answer: A. 'can write anything they like'
        \\ \midrule \textbf{Response After Scaling Based on RACE as $\mathcal{D}_{\text{non-math}}$ (incorrect)}\\ Answer: A
        \\ \midrule \textbf{Response After Scaling Based on MMLU as $\mathcal{D}_{\text{non-math}}$ (incorrect)}\\ Answer: A
        \\ \bottomrule
    \end{tabular}}
    \caption{Responses to a RACE question before and after pruning or scaling parameters identified by MathNeuro for Llama 3 1B IT.}
    \label{tab:race_outputs}
\end{table*}

\begin{table*}[t]
    \centering\resizebox{\linewidth}{!}{
    \begin{tabular}{p{1\linewidth}} \toprule
        \textbf{Prompt} \\
        Question: A 30-year-old nulliparous female presents to the office with the complaint of mood changes. She says that for the past several months she has been anxious, hyperactive, and unable to sleep 3 to 4 days prior to the onset of menses. She further reports that on the day her menses begins she becomes acutely depressed, anorectic, irritable, and lethargic. She has no psychiatric history. Physical examination findings are normal. She and her husband have been trying to conceive for over 2 years. History reveals a tuboplasty approximately 1 year ago to correct a closed fallopian tube. The most likely diagnosis is
        Answer choices: ['adjustment disorder with depressed mood', 'bipolar I disorder, mixed', 'cyclothymic personality', 'generalized anxiety disorder']
        \\ \midrule 
        \textbf{Response Before Pruning (correct; hit max number of output tokens)}\\ 
        Answer: A \newline
        Explanation: The patient's symptoms of mood changes, anxiety, hyperactivity, inability to sleep, acute depression, anorexia, irritability, and lethargy, particularly in the context of her history of tuboplasty, are characteristic of adjustment disorder with depressed mood. The patient's anxiety, hyperactivity, and acute depression in the premenstrual phase of her cycle are also consistent with this diagnosis. The mixed presentation of mood changes is also consistent with this diagnosis, as bipolar I disorder, mixed type typically involves a combination of manic and depressive episodes. Cyclothymic personality disorder is characterized by periods of hypomania and depression that last for at least 2 years. Generalized anxiety disorder is characterized by excessive worry and
        \\ \midrule 
        \textbf{Response After Pruning Based on RACE as $\mathcal{D}_{\text{non-math}}$ (correct)}\\
        Answer: 'adjustment disorder with depressed mood' \newline
        The most likely diagnosis is adjustment disorder with depressed mood. This is a condition where the person's symptoms of depression are triggered by a stressful event, but the symptoms are not severe enough to meet the criteria for a major depressive disorder. The symptoms of depression are typically present for at least two days, are persistent, and are severe enough to cause significant distress or impairment in social or work functioning. 
        The diagnosis of adjustment disorder with depressed mood is typically made based on the criteria outlined in the DSM-IV. 
        The symptoms of depression in adjustment disorder with depressed mood include:
        1. Depressed mood
        2. Loss of interest in activities
        3. Insomnia or hypersomnia
        4. Fatigue
        \\ \midrule 
        \textbf{Response After Pruning Based on MMLU as $\mathcal{D}_{\text{non-math}}$ (incorrect; hit max number of output tokens)}\\
        Answer: B \newline
        Explanation: A 30-year-old nulliparous female presenting with mood changes is a classic presentation for bipolar disorder, particularly in the early stages. The patient's symptoms of irritability, lethargy, and an acute depression during menses are classic for a manic episode. The history of tuboplasty is a significant clue that points towards a bipolar disorder, especially given the rapid onset of symptoms and the pattern of episodes. The patient's past history of depression and the fact that she and her husband have been trying to conceive for over 2 years suggests a pattern of mood instability and possibly a depressive episode. The most likely diagnosis is bipolar I disorder, mixed. The symptoms of mood changes, irritability, lethargy, and 
        \\ \bottomrule
    \end{tabular}}
    \caption{Responses to a MMLU question before and after pruning parameters identified by MathNeuro for Llama 3 1B IT.}
    \label{tab:mmlu_outputs}
\end{table*}

\begin{table*}[t]
    \centering\resizebox{\linewidth}{!}{
    \begin{tabular}{p{1\linewidth}} \toprule
        \textbf{Prompt} \\
        Question: A 30-year-old nulliparous female presents to the office with the complaint of mood changes. She says that for the past several months she has been anxious, hyperactive, and unable to sleep 3 to 4 days prior to the onset of menses. She further reports that on the day her menses begins she becomes acutely depressed, anorectic, irritable, and lethargic. She has no psychiatric history. Physical examination findings are normal. She and her husband have been trying to conceive for over 2 years. History reveals a tuboplasty approximately 1 year ago to correct a closed fallopian tube. The most likely diagnosis is
        Answer choices: ['adjustment disorder with depressed mood', 'bipolar I disorder, mixed', 'cyclothymic personality', 'generalized anxiety disorder']
        \\ \midrule 
        \textbf{Response Before Scaling (correct; hit max number of output tokens)}\\ 
        Answer: A \newline
        Explanation: The patient's symptoms of mood changes, anxiety, hyperactivity, inability to sleep, acute depression, anorexia, irritability, and lethargy, particularly in the context of her history of tuboplasty, are characteristic of adjustment disorder with depressed mood. The patient's anxiety, hyperactivity, and acute depression in the premenstrual phase of her cycle are also consistent with this diagnosis. The mixed presentation of mood changes is also consistent with this diagnosis, as bipolar I disorder, mixed type typically involves a combination of manic and depressive episodes. Cyclothymic personality disorder is characterized by periods of hypomania and depression that last for at least 2 years. Generalized anxiety disorder is characterized by excessive worry and \\
        \midrule \textbf{Response After Scaling Based on RACE as $\mathcal{D}_{\text{non-math}}$ (correct; hit max number of output tokens)}\\
        Answer: A\newline
        Explanation:  The patient's symptoms of mood changes (anxiety, hyperactivity, irritability, lethargy) and the physical symptoms (anorexia) of anorexia nervosa, which are typically seen in the context of hormonal fluctuations, are consistent with this diagnosis. The physical symptoms of anorexia nervosa are also consistent with the tubal surgery. The patient's symptoms do not meet the criteria for a manic episode (i.e., she is not hyperactive or irritable for more than one week), and her symptoms do not meet the criteria for a depressive episode (i.e., she is not depressed for more than two weeks). The patient's age and the fact that she is trying to conceive do not support\\
        \midrule \textbf{Response After Scaling Based on MMLU as $\mathcal{D}_{\text{non-math}}$ (correct; hit max number of output tokens)}\\
        Answer: A\newline
        Explanation: The patient's symptoms of mood changes, anxiety, hyperactivity, and sleep disturbances, particularly in the context of her menstrual cycle, are characteristic of premenstrual syndrome (PMS). The patient's history of tuboplasty and the timing of her symptoms suggest that she may be experiencing a cyclical pattern of mood changes, which is a hallmark of bipolar I disorder. The patient's symptoms are also consistent with a diagnosis of adjustment disorder with depressed mood, which is a type of mood disorder that occurs in response to a significant life stressor. The patient's symptoms are not consistent with cyclothymic personality or generalized anxiety disorder, which do not typically present with such a cyclical pattern of mood changes.
        The best answer
        \\ \bottomrule
    \end{tabular}}
    \caption{Responses to a MMLU question before and after scaling parameters identified by MathNeuro for Llama 3 1B IT.}
    \label{tab:mmlu_outputs_scaling}
\end{table*}

%% file: sections/hyperparameter_search.tex
Because an exhaustive grid search for the optimal scaling factor for MathNeuro would be computationally prohibitive, we used a rough bisection grid search to find a factor that worked best for each model for the GSM8K and MATH scaling experiments. For each model, we tried three scaling factors based on initial experiments that showed scale factors above 1.1 were too large: 1.01, 1.05, and 1.075. For GSM8K, for smaller models (Phi 1.5, Gemma 2 2B IT, Llama 3.2 1B IT, and Llama 3.2 3B IT), 1.075 worked best or tied with 1.05, while for Llama 3.1 8B IT, a larger model, 1.01 worked best. For the smaller models, we next tried scale factors between 1.075 and 1.05 (1.0625) and between 1.075 and the maximum scale factor we saw improved results based on initial experiments (1.1), finding that 1.1 worked best for all models except for Llama 3.2 1B IT, where 1.1 tied with the results of 1.075. For GSM8K for Llama 3.1 8B IT, we next tried a scale factor between 1.05 and the minimum scale factor we used (1.025), finding that 1.01 still worked best. Each scale factor considered increased performance across models for GSM8K. 

\begin{figure*}[t]
    \centering
    \includegraphics[width=1\linewidth]{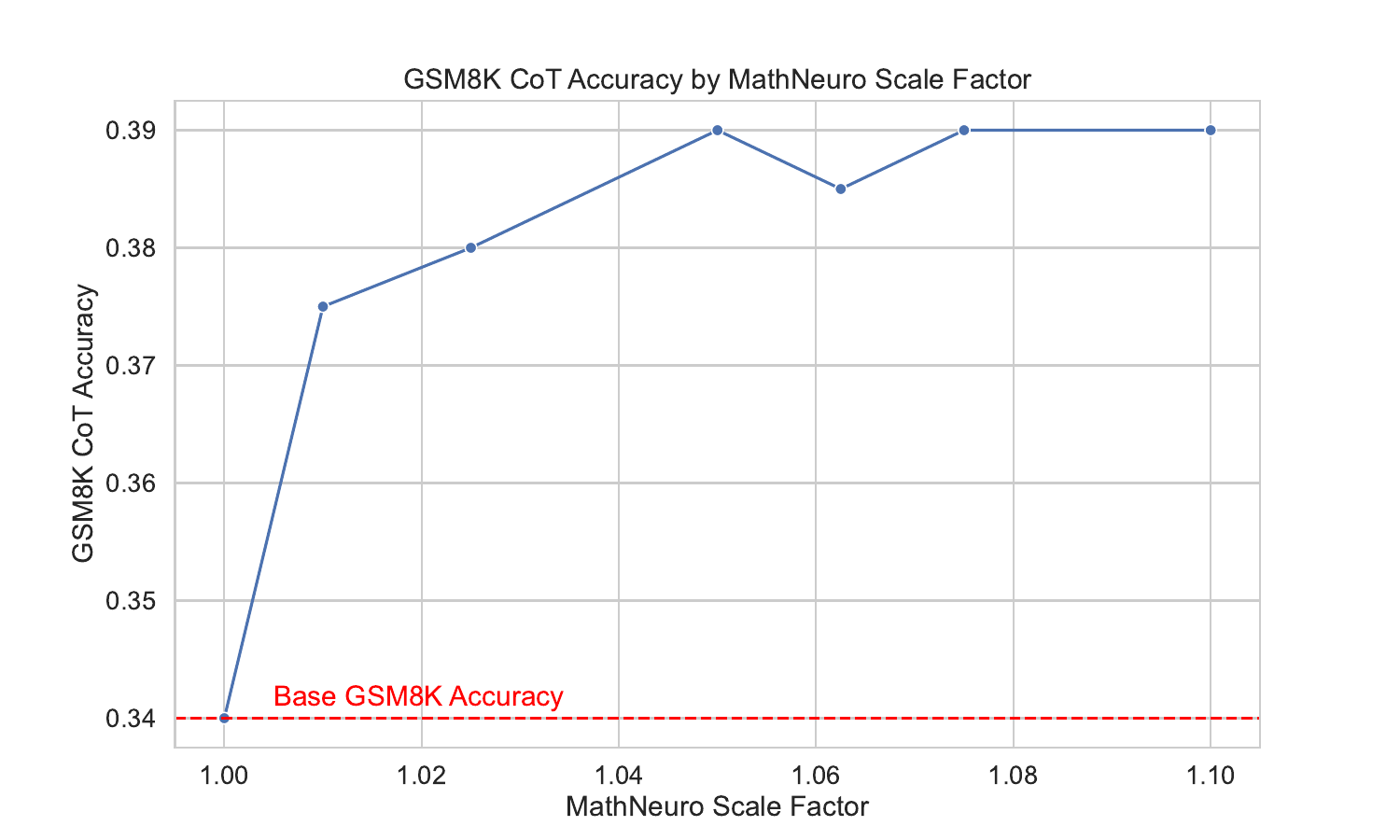}
    \caption{Impact of MathNeuro scale factor on GSM8K performance for Llama 3.2 1B IT.}
    \label{fig:scale_factor_grid}
\end{figure*}

For MATH, for small models, 1.01 or 1.05 worked best, while for Llama 3.1 8B IT, 1.01 worked best. For the smaller models, we next tried a scale factor between 1.01 and 1.05 (1.025), finding that 1.025 worked best for Llama 3.2 1B IT, Phi 1.5, and Llama 3.2 3B IT, and that 1.05 worked best for Gemma 2 2B IT. For MATH, scale factors beyond 1.05 either did not improve performance or harmed performance, suggesting a smaller scale factor is more optimal for this task. However, scale factors between 1.01 and 1.05 improved MATH performance across models. The results of this grid search for Llama 3.2 1B for GSM8K are displayed in Figure \ref{fig:scale_factor_grid}. 

%% file: sections/mmlu_param_location_number_figures.tex
\begin{figure}[t]
    \centering
    \includegraphics[width=1\linewidth]{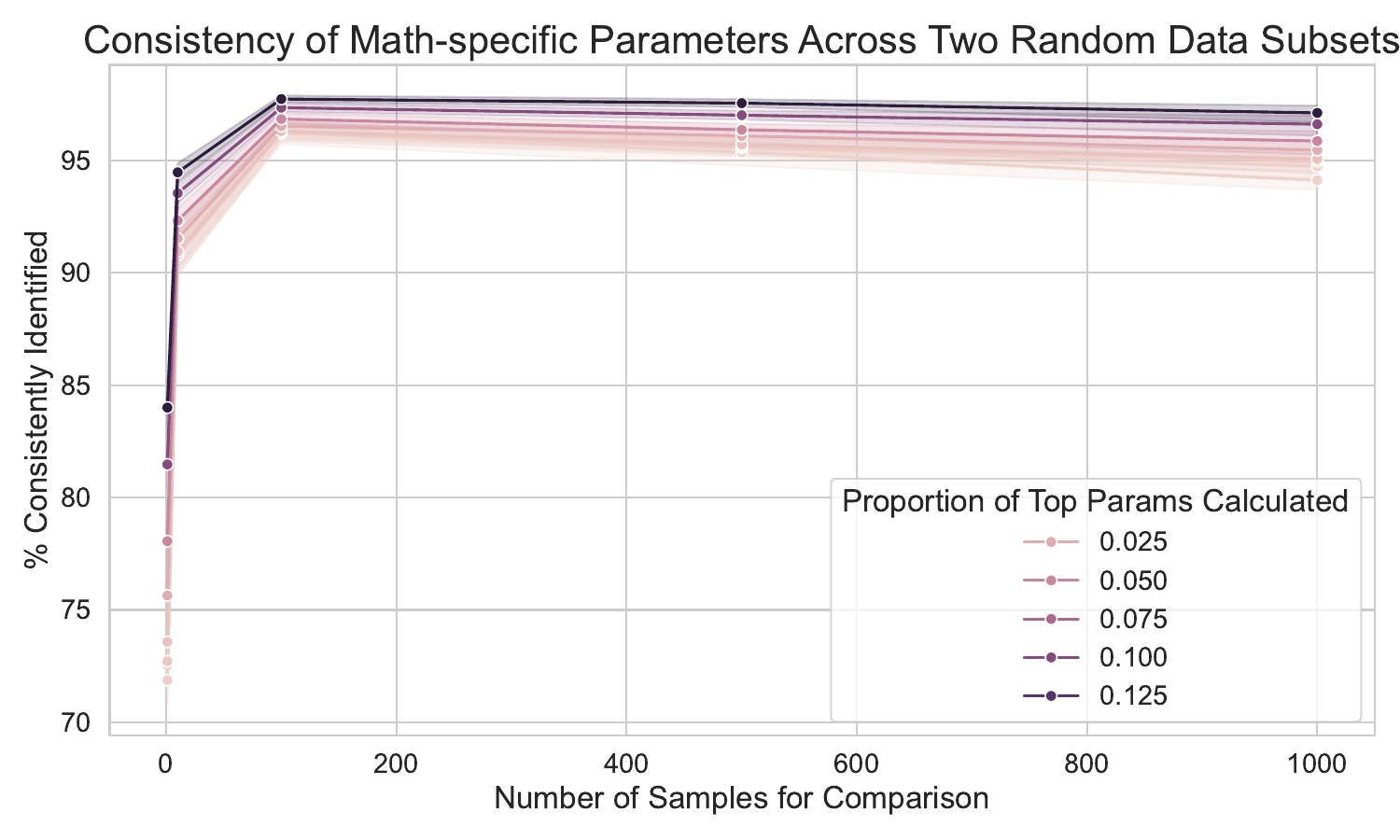}
    \caption{Consistency of math-specific parameters identified by MathNeuro for Llama 3.2 1B IT when identifying using GSM8K compared to MMLU.}
    \label{fig:mmlu_consistency}
\end{figure}
\begin{figure}[t]
    \centering
    \includegraphics[width=1\linewidth]{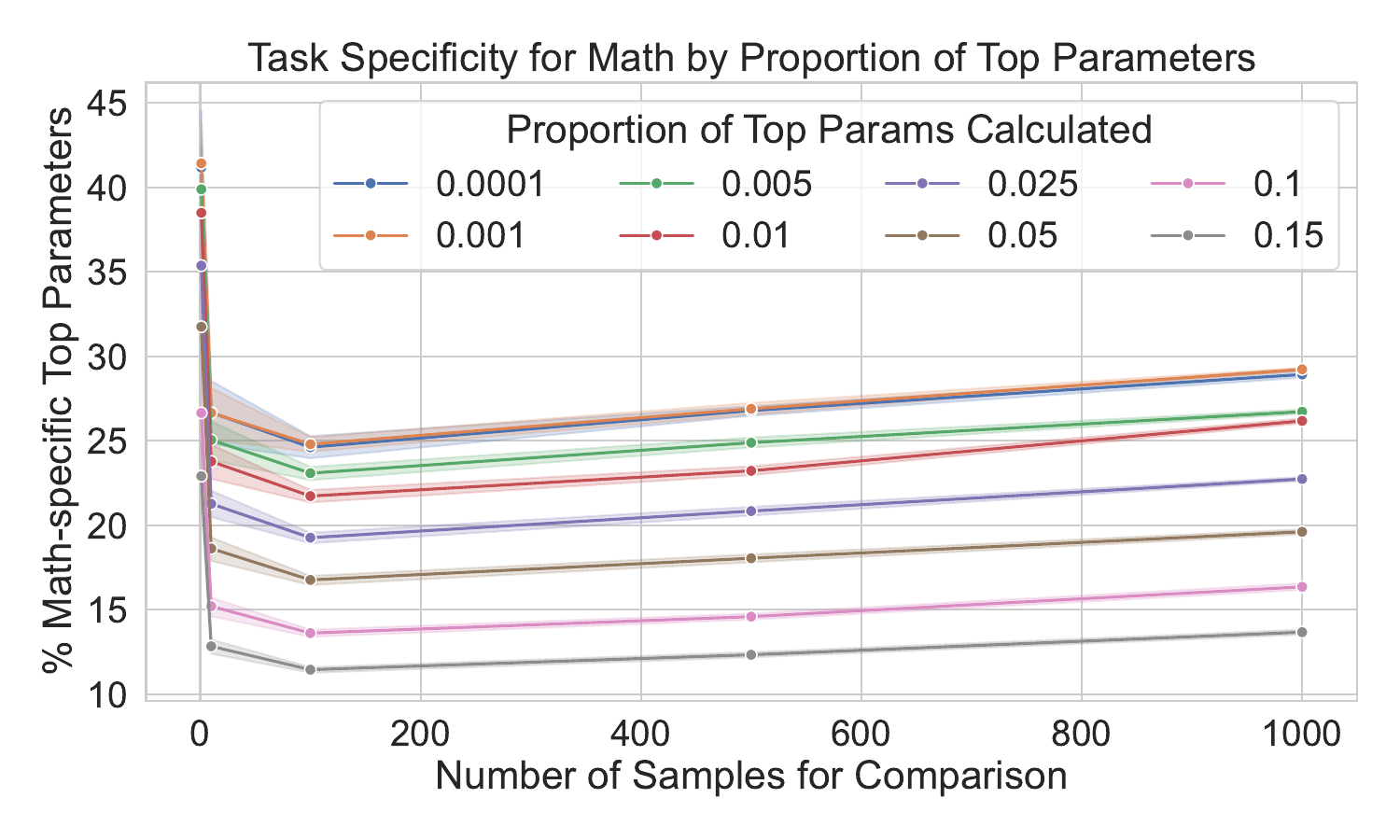}
    \caption{Percentage of top parameters that are math-specific as identified by MathNeuro for Llama 3.2 1B IT when identifying using GSM8K compared to MMLU.}
    \label{fig:mmlu_unique}
\end{figure}
\begin{figure}[t]
    \centering
    \includegraphics[width=1\linewidth]{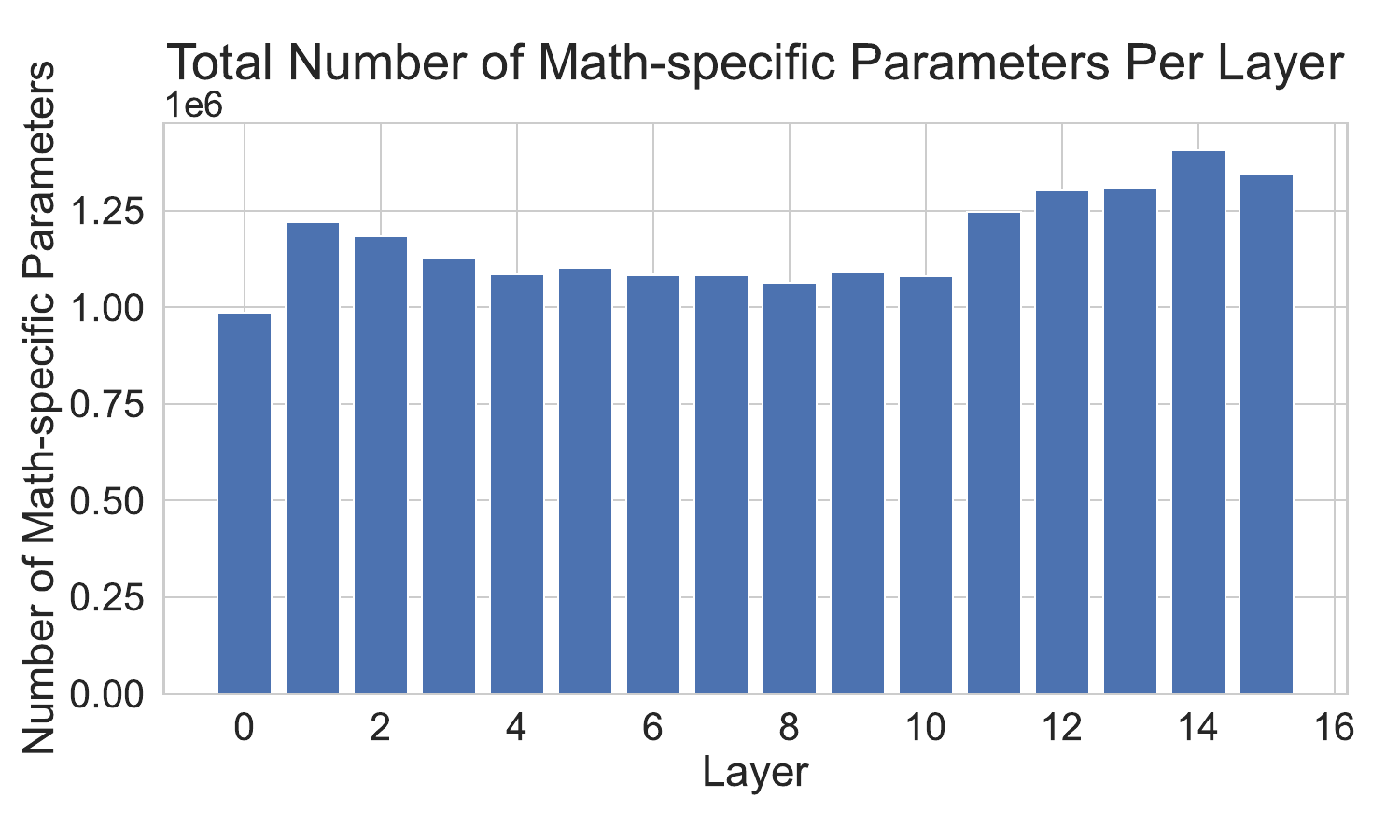}
    \caption{Distribution of math-specific parameters identified by MathNeuro for Llama 3.2 1B IT when identifying using GSM8K compared to MMLU.}
    \label{fig:mmlu_distribution}
\end{figure}

Figures \ref{fig:mmlu_consistency}, \ref{fig:mmlu_unique}, and \ref{fig:mmlu_distribution} show the consistency of math-specific parameters, percentage of top parameters that are math-specific, and distribution of math-specific parameters identified by MathNeuro using MMLU as the non-math dataset, respectively, based on the experiments described in Section \ref{param_location_and_number}.

